\documentclass{article}

    \PassOptionsToPackage{numbers, compress}{natbib}
\usepackage[main,final]{neurips_2026}

\usepackage[utf8]{inputenc} % allow utf-8 input
\usepackage[T1]{fontenc}    % use 8-bit T1 fonts
\usepackage{hyperref}       % hyperlinks
\usepackage{url}            % simple URL typesetting
\usepackage{booktabs}       % professional-quality tables
\usepackage{amsfonts}       % blackboard math symbols
\usepackage{nicefrac}       % compact symbols for 1/2, etc.
\usepackage{microtype}      % microtypography
\usepackage[table]{xcolor}         % colors
\usepackage{amsmath}
\usepackage{cleveref}
\usepackage{cancel}
\usepackage{tikz}
\usetikzlibrary{
    positioning,
    fit,
    shapes,
    shapes.geometric,
    arrows.meta,
    calc,
    backgrounds
}
\usepackage{multirow}
\usepackage{tabularx}
\usepackage{pifont}
\usepackage{wrapfig}
\usepackage{bm}
\usepackage{ulem} % in preamble
\usepackage{subcaption}
\usepackage{adjustbox}
\usepackage{placeins}
\usepackage{wrapfig}

\newcommand{\veps}{\boldsymbol{\epsilon}}
\newcommand{\x}{\mathbf{x}}
\newcommand{\z}{\mathbf{z}}

\newcommand{\logsnr}{\log \mathrm{SNR}}
\newcommand{\normal}{\mathcal{N}}

\newcommand{\identity}{\mathbf{I}}

\newcommand{\bmu}{\boldsymbol{\mu}}

\newcommand{\E}{\mathbb{E}}
\newcommand{\loss}{\mathcal{L}}
\newcommand{\inv}{\mathrm{-1}}
\newcommand{\IB}{\mathrm{IB}_{\kappa}}
\newcommand{\Feat}{\mathrm{Feat}_{\psi}}
\newcommand{\s}{\mathbf{s}}
\newcommand{\y}{\mathbf{y}}
\newcommand{\tokens}{\mathbf{S}}

\definecolor{steelblue}{HTML}{3F82CA}
\definecolor{cornflowerblue}{HTML}{6F9CEB}
\definecolor{babyblueice}{HTML}{98B9F2}
\definecolor{aliceblue}{HTML}{DFEAF6}
\definecolor{periwinkle}{HTML}{B6B5F7}
\definecolor{softperiwinkle}{HTML}{918EF4}
\definecolor{steelpurple}{HTML}{7B5FC8}
\definecolor{darkred}{RGB}{153, 0, 0}

\hypersetup{
  colorlinks,
  linkcolor=darkred,     % section/page links
  citecolor=steelblue,    % citations
  urlcolor=steelblue     % URLs
}

\title{Steering Training Trajectories in \\ Diffusion Representation Learning}

\author{
Rajat Rasal$^{\star}$, \quad Avinash Kori, \quad Tian Xia, \quad Ben Glocker \\
Imperial College London \\
\texttt{\{rrr2417,t.xia,agk21,b.glocker\}@imperial.ac.uk}
}

\begin{document}

\maketitle

\begin{abstract}
Across training runs of a diffusion autoencoder, we observe similar image reconstruction quality, but substantially different levels of disentanglement in the learned representations.
In this work, we study how training dynamics contribute to this behaviour.
By analysing \textit{training trajectories} in a controlled setting - curves of reconstruction quality against disentanglement during training - we identify two recurring trajectory patterns, which we call \textit{learning regimes}.
These regimes differ in how reconstruction and disentanglement evolve, despite similar reconstruction at the end of training.
In the \textit{reconstruction regime}, models improve image fidelity before disentanglement, while in the \textit{disentanglement regime}, models improve both reconstruction and disentanglement more gradually.
We hypothesise that modulating shortcut pathways in the diffusion U-Net and controlling early noise-level exposure can shape trajectories. 
To that end, we introduce \textsc{SteeringDRL}, combining gated residual U-Nets with a noise curriculum to steer models towards better disentanglement.
Across attribute disentanglement benchmarks, \textsc{SteeringDRL} has competitive disentanglement, with reduced seed sensitivity in many settings.
Our method extends to spatial disentanglement with object-centric learning, improving unsupervised segmentation on synthetic and real-world datasets.
Our code is publicly available at \texttt{\url{https://github.com/RajatRasal/SteeringDRL}}.
\end{abstract}

\section{Introduction}
A central hypothesis in representation learning is that decomposing observations into \textit{disentangled} components \citep{bengio2013representation,schmidhuber1992learning} enables compositional generalisation in a data-efficient manner \citep{piantadosi2016logical,collins2024building}.
Here, disentanglement refers to the separation of invariant factors in observational data \citep{bengio2013representation}.
As generative models are increasingly used for foundational tasks, the ability to learn structured representations becomes critical for controllability, transfer learning, and domain adaptation \citep{labs2025flux1kontextflowmatching,chambon2022roentgen,khanna2023diffusionsat,hudson2024soda}.
However, models can generate high-fidelity images whilst representations are poorly aligned with true causal factors \citep{leng2025repa,heek2026unified}.
In our work, we find that diffusion autoencoders, with identical objectives and architectures, can produce similar image quality with substantially different latent representations.

\citet{locatello2019challenging} prove that disentangled representation learning (DRL) requires appropriate inductive biases or appropriate distributional assumptions on both the model, such as supervision or regularisation, and the data, such as knowledge of the data-generating process.
Given the challenges with obtaining ground-truth labels, research efforts have focused on designing inductive biases for unsupervised attribute disentanglement \citep{higgins2017betavae} and spatial disentanglement with object-centric learning (OCL) \citep{locatello2020object}.
While effective, these methods often trade off between representation learning and generative fidelity \citep{burgess2018understanding,pmlr-v202-brady23a}.
Here, diffusion autoencoders provide a promising alternative \citep{preechakul2022diffusion,wang2023infodiffusion,zhang2022unsupervised}, potentially mitigating this trade-off \citep{yu2024representation,heek2026unified}.
Our findings suggest that diffusion autoencoders admit a spectrum of representational solutions, and which one is reached is influenced by training dynamics. % whether they emerge depends on the optimisation dynamics.
By analysing image reconstruction and latent disentanglement over training, we motivate inductive biases for steering diffusion autoencoders towards better disentangled latent representations.

% \newpage

Diffusion models exhibit a natural hierarchical structure: coarse semantic information is captured at high noise levels, while fine-grained details are recovered at lower noise levels \citep{wang2023diffusion,liu2024faster}.
Cross-attention provides a mechanism for querying this hierarchy, aligning conditioning signals with spatial representations across noise levels \citep{hertz2022prompt,tang2023daam}.
This inductive bias underpins both text-to-image and object-centric diffusion models \citep{rombach2022high}, where text tokens or object-slots are aligned with regions in pixel space \citep{wu2023slotdiffusion,jiang2023object}.
In particular, \citet{yang2024diffusion} demonstrate that cross-attention induces disentanglement without additional regularisation.
Although this simple setup is compatible with large generative models, progress in disentanglement often relied on complex regularisation \citep{chi2026disentangled,wang2025can}, computationally expensive compositional inductive biases \citep{Wu_Zheng_2024,jung2025disentangled,jung2024learning}, or migration to newer generative model classes.
We instead revisit diffusion autoencoders with cross-attention, and ask how their existing inductive biases can be better exploited for representation learning.

% In this work, 
We study the optimisation dynamics of unsupervised representation learning with diffusion autoencoders via \textit{training trajectories}: curves of reconstruction against representation quality during training.
In a controlled setting, trajectories exhibit two recurring patterns, which we call \textit{regimes}.
In the \textit{disentanglement regime}, models progressively improve disentanglement, while in the \textit{reconstruction regime}, models prioritise reconstruction early in training.
We hypothesise that early noise-level exposure and architectural shortcuts in the diffusion U-Net can influence which regime a run follows.
Therefore, we introduce \textsc{SteeringDRL}, combining gated residual U-Nets with a noise curriculum to restrict reconstruction shortcuts and control early training dynamics.
For attribute disentanglement, \textsc{SteeringDRL} improves or remains competitive with matched baselines, while reducing seed sensitivity in many settings.
For spatial disentanglement with OCL, it improves unsupervised segmentation over matched baselines on synthetic and real-world datasets.
% Relative to matched EncDiff baselines, \textsc{SteeringDRL} improves disentanglement on Cars3D and Shapes3D, remains competitive on MPI3D-toy, and reduces seed sensitivity.
% \textsc{SteeringDRL} improves or retains performance over matched baselines on attribute disentanglement benchmarks, while reducing seed sensitivity.
% For spatial disentanglement with object-centric learning (OCL), we improve unsupervised segmentation over matched baselines on synthetic and real-world datasets.
Our results show that steering training trajectories is a simple yet effective method for improving disentanglement in diffusion autoencoders.

We make the following contributions:
\begin{enumerate}
    \item \textbf{Regimes in Training Trajectories.} 
    We show that diffusion autoencoders have training trajectories which organise around two regimes early in training under specific conditions.
    We propose hypotheses for why this happens (\Cref{sec:analysis}).

    \item \textbf{Steering Trajectories.} 
    Our hypotheses motivate architectural inductive biases and training strategies to steer trajectories towards better disentanglement: \textbf{\textsc{SteeringDRL}} (\Cref{sec:steering}).

    \item \textbf{Scaling \textsc{SteeringDRL}.} 
    We show that \textsc{SteeringDRL} remains effective as model capacity increases, extending from attribute disentanglement to spatial disentanglement with OCL on synthetic and real-world datasets, while improving training efficiency (\Cref{sec:experiments}).
\end{enumerate}

\section{Related Work}

\textbf{Unsupervised Representation Learning.}
Diffusion autoencoders \citep{preechakul2022diffusion,wang2023infodiffusion,zhang2022unsupervised} have been used to modernise unsupervised representation learning, extending earlier VAE and GAN-based approaches for attribute disentanglement \citep{higgins2017betavae,kim2018disentangling,chen2018isolating,chen2016infogan} and object-centric learning \citep{burgess2019monet,locatello2020object,brady2024interaction}.
Recent diffusion methods introduce regularisations and compositional inductive biases to encourage disentanglement \citep{yang2023disdiff,Wu_Zheng_2024,jung2024learning,jung2025disentangled,chi2026disentangled,wang2025can}. 
In contrast, EncDiff \citep{yang2024diffusion} highlights cross-attention alone as a simple and scalable component for disentanglement in diffusion, which SlotDiffusion \citep{wu2023slotdiffusion} also adopts for object-centric learning.
This mechanism underpins control in foundational generative models \citep{chefer2023attendandexcite,hertz2022prompt,tang2023daam,wu2023slotdiffusion,jiang2023object}.
As such, our work builds on EncDiff and SlotDiffusion, demonstrating how their simple and scalable inductive biases can be better exploited by analysing training trajectories.

\textbf{Inductive Biases in Diffusion Models.}
A large body of work studies the design and optimisation of diffusion models, including noise schedules \citep{chen2023importance,nichol2021improved}, timestep sampling strategies \citep{hang2023efficient,choi2022perception}, and prediction parameterisations \citep{li2025back,salimans2022progressive,karras2022elucidating,lipman2022flow}. 
These components interact in subtle and often opaque ways, and are typically analysed in terms of generative quality, stability, or sampling efficiency.
To that end, we adopt the VDM++ formulation \citep{kingma2023understanding,hoogeboom2023simple,hoogeboom2024simpler,ribeiro2025demystifying}, which unifies a wide class of generative objectives as variance-preserving processes in $\logsnr$ space ($\lambda$), differing primarily by a loss-weighting function $w(\lambda)$.
As such, our framework naturally extends to other diffusion-style generative models.
Architectural choices, such as residual, gated or removed skip connections \citep{peebles2023scalable,hoogeboom2024simpler,touvron2021going,chen2026attnres}, have been shown to improve representational alignment \citep{yu2025repa}. 
Similarly, some works explore timestep sampling curricula or reweighting strategies during training \citep{kim2024adaptive,xu2024towards,kim2025denoising,yue2024exploring,wang2025closer,li2025understanding}, which control exposure to different noise regimes.
However, their interaction with training dynamics and the emergence of representations \citep{okawa2023compositional} has not been systematically studied.

\newpage

\section{Setup: Latent Variational Diffusion Autoencoder}
\label{sec:setup}
This section outlines the experimental setup that we build on throughout this paper.
Following prior work on unsupervised diffusion representation learning \citep{wu2023slotdiffusion,yang2024diffusion}, we model images using a diffusion autoencoder \citep{preechakul2022diffusion}.
This consists of a semantic encoder, which generates semantic tokens ($\tokens$) from images ($\mathbf{y}$), and the diffusion decoder, operating in the latent space ($\x_0$) of a pretrained VQ-VAE \citep{esser2021taming,rombach2022high} to reconstruct images from $\tokens$.
We train the diffusion model in $\logsnr$-space ($\lambda$) following the VDM++ framework \citep{kingma2023understanding,ribeiro2025demystifying}, which unifies a wide class of generative objectives as variance-preserving processes in $\logsnr$ space ($\lambda$), differing by a loss-weighting function $w(\lambda)$.

\subsection{Latent Diffusion Decoder}
Let $\x_0$ denote the projection of an image $\y$ into the latent space of a pretrained VQ-VAE \citep{rombach2022high,esser2021taming}.
We learn a diffusion decoder in continuous-time \citep{kingma2021variational} on $\x_0$, keeping the VQ-VAE parameters frozen. 
Operating in this perceptually compressed latent space reduces computational cost and encourages the diffusion model to focus on high-level semantics rather than high-frequency pixel-level details.
Here, $\x_0$ is mapped to a noisy intermediate $\x_t$ at time $t \in [0, 1]$ via the forward process,
\begin{align}
    q(\x_t \mid \x_0) = \normal(\x_t; \alpha_t \x_0, \sigma^2_t \identity), \qquad \x_t = \alpha_t \x_0 + \sigma_t \veps,
 \end{align}
where $\veps \sim \normal(0, \identity)$, $\alpha_t = \sqrt{\mathrm{sigmoid}(\lambda_t)}$ and $\sigma_t = \sqrt{\mathrm{sigmoid}(-\lambda_t)}$ satisfy the variance-preserving (VP) constraint $\alpha_t^2 + \sigma_t^2 = 1$, $\lambda_t = \logsnr(t) = \log(\alpha_t^2 / \sigma_t^2)$ is the log signal-to-noise ratio (logSNR).
The conditional generative transitions $p_\theta(\x_s|\x_t, \tokens)$ are defined 
% via the forward posterior $q(\x_s | \x_t, \x_0)$ 
with two timesteps $s < t$,
\begin{align}
    & p_\theta(\x_s \mid \x_t, \tokens) 
    % = q(\x_s \mid \x_t, \hat{\x}_\theta(\x_t; t, S))
    = \mathcal{N}(\x_s;\, \bmu_\theta(\x_t; s, t, \tokens),\, \sigma^2(s,t)\mathbf{I}), \\
    \text{with} \quad \sigma^2(s,t) = & \frac{\sigma_{t|s}^2 \sigma_s^2}{\sigma_t^2} \quad \text{and} \quad \bmu_\theta(\x_t;s,t,S) = \x_t \frac{\alpha_{t|s}\,\sigma_s^2}{ \sigma_t^2} + \hat{\x}_\theta(\x_t;t,S) \frac{\alpha_s\,\sigma_{t|s}^2}{\sigma_t^2},
\end{align}
where $\tokens$ are tokens generated by the semantic encoder (\Cref{sec:semantic_encoder}), $p_\theta(\x_s | \x_t, \tokens) = q(\x_s | \x_t, \x_0 = \hat{\x}_\theta(\x_t;t,\tokens))$, and a U-Net \citep{ronneberger2015u} $\hat{\veps}_\theta(\cdot)$ is used to implement $\hat{\x}_\theta(\cdot)$ via the identity
$\hat{\x}_\theta(\x_t;t,\tokens) = (\x_t - \sigma_t \hat{\veps}_\theta(\x_t; t, \tokens)) / \alpha_t$
with $\alpha_{t|s} = \alpha_t/\alpha_s$ and $\sigma_{t|s}^2 = \sigma_t^2 - \alpha_{t|s}^2\sigma_s^2$.

\subsection{Semantic Encoder}
\label{sec:semantic_encoder}
The semantic tokens $\tokens = \{\s_i\}^N_{i = 1}$ are generated from the image $\y$ by a semantic encoder $\mathrm{Enc}_{\phi}(\cdot)$,
\begin{equation}
    \tokens = \mathrm{Enc}_\phi(\y)
      = \big(\IB \circ \Feat)(\y), \qquad \z = \Feat(\y)
\end{equation}
where $\Feat(\cdot)$ is a feature extractor and $\IB(\cdot)$ is an inductive bias that maps features $\z$ into semantic tokens $\tokens$, which condition the decoder via cross-attention.
For disentanglement, \citep{yang2024diffusion,jun2025disentangling} implements $\IB(\cdot)$ using a split-MLP architecture, where $N$ parallel MLPs generate factor-wise tokens $\tokens$, while disentanglement is measured among scalars in $\z$.
For spatial disentanglement, \citep{wu2023slotdiffusion,jiang2023object} use slot attention for $\IB(\cdot)$, producing a set of object representations (\textit{slots}) $\tokens$ and object masks $\mathcal{M}$ \citep{locatello2020object}.

\subsection{Optimisation}
\label{sec:optim}
\citet{kingma2023understanding} show that the $\veps$-prediction loss in diffusion models \citep{ho2020denoising} is a special case of the estimator
\begin{equation}
    \loss(\x, \tokens)
    =
    \frac{1}{2}
    \E_{\veps\sim\normal(0,\identity),\,\lambda\sim p(\lambda)}
    \left[
    \frac{w(\lambda)}{p(\lambda)}
    \big\| \veps-\hat{\veps}_\theta(\x_\lambda;\lambda,\tokens) \big\|_2^2
    \right],
    \label{eq:weighted_elbo_importance_sampling}
\end{equation}
where $w(\lambda) = 1$ recovers the vanilla ELBO.
This form follows from a change of variables from time ($t$) to $\logsnr$ ($\lambda$) using a bijective noise schedule $f_\lambda: t \mapsto \lambda$.
This induces the distribution $p(\lambda) = -\frac{d}{d\lambda} f_{\lambda}^{-1}(\lambda)$,  truncated to $[\lambda_{\min}, \lambda_{\max}]$, when sampling $t \sim \mathcal{U}(0,1)$.
In this work, we use the cosine noise schedule (\Cref{app:definitions}).
It follows that \Cref{eq:weighted_elbo_importance_sampling} can be reformulated as
\begin{equation}
    \loss(\x, \tokens)
    =
    \frac{1}{2}
    \E_{\veps\sim\normal(0,\identity),\,t \sim \mathcal{U}(0, 1)} %p(\lambda)}
    \left[
    w(\lambda) \cdot \;
    -\frac{d\lambda}{dt} \;
    \big\| \veps-\hat{\veps}_\theta(\x_\lambda;\lambda,\tokens) \big\|_2^2
    \right],
    \label{eq:weighted_elbo_deriv}
\end{equation}
which may be more compatible with existing codebases, which commonly train in diffusion time ($t$).
Note that monotonic loss-weightings $w(\lambda)$ maximise an ELBO under Gaussian-noise data augmentation \citep{child2019generating}.
In particular, setting $w(\lambda)= \mathrm{sigmoid}(-\lambda)$ is shown to improve image fidelity \citep{kingma2023understanding,hoogeboom2024simpler}.
For $\x_0$-prediction, we use the weighting $w_{\x_0}(\lambda) = \mathrm{sigmoid}(\lambda)$, which induces the same effective $\veps$-prediction weighting $w_{\veps}(\lambda) = \mathrm{sigmoid}(-\lambda)$ (\Cref{app:xpred_weighting}).

\newpage

\section{Early Emergence of Regimes in Training Trajectories} 
\label{sec:analysis}

\begin{figure*}
    \centering
    \includegraphics[width=\textwidth]{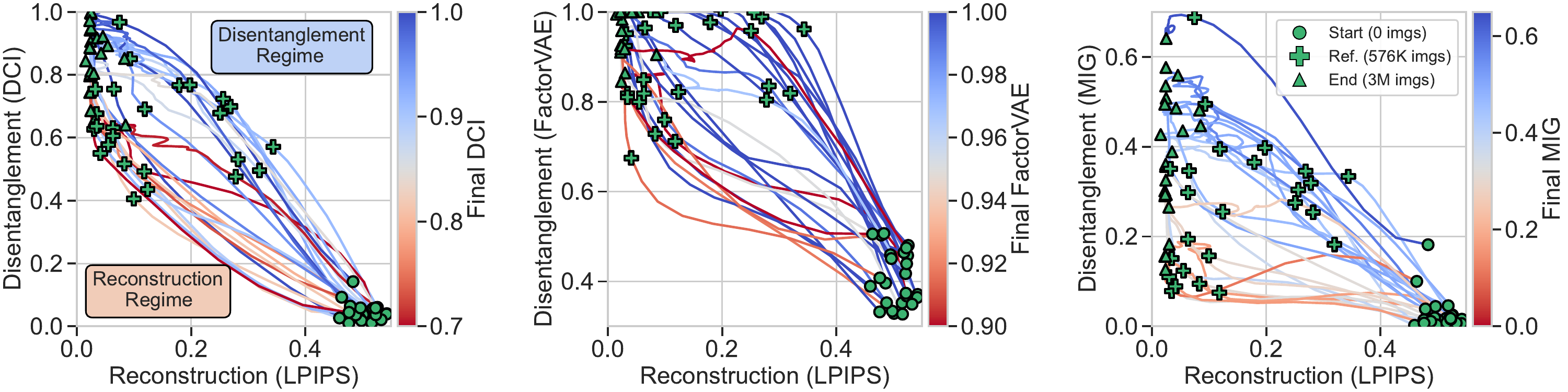}
\caption{
\textbf{Trajectories can organise around two regimes.}
% Reconstruction (LPIPS) vs. disentanglement (DCI left, FactorVAE centre, MIG right) trajectories for a low-capacity diffusion autoencoder on Shapes3D (25 runs).
Training trajectories for a low-capacity diffusion autoencoder on Shapes3D (25 seeds).
For interpretability, curves are coloured by final disentanglement, trajectories are smoothed and colour bars are centred at median final disentanglement.
}
\label{fig:shapes3d_regimes}
\end{figure*}

% We study diffusion autoencoders for attribute disentanglement in a controlled setting.
% Here, we observe that trajectories form two distinct regimes.
% \Cref{app:analysis} provides further examples and analysis, and outlines settings where this pattern may not be as clear.
% Our hypotheses for this pattern motivate our method (\Cref{sec:steering}), which we later show can steer training trajectories in larger models towards competitive disentanglement while reducing seed sensitivity (\Cref{sec:experiments}).

We study diffusion autoencoders for attribute disentanglement in a controlled setting.
Here, we observe that trajectories organise around two regimes.
These observations motivate hypotheses (\textbf{H1} / \textbf{H2}) about the underlying training dynamics and, in turn, our method (\S\ref{sec:steering}).
In larger models, where the baseline regime separation may not be as clear, our method is still able to steer trajectories to achieve competitive disentanglement and/or reduce seed sensitivity (\S\ref{sec:experiments}).
% We use our method to steer trajectories in larger models with more mature VQ-VAE checkpoints, matched in our baselines, to achieve competitive disentanglement and/or reduce seed sensitivity (\S\ref{sec:experiments}).
\Cref{app:analysis} provides further analyses and outlines when this pattern is less clear, including sensitivity to VQ-VAE maturity.
% We observe two recurring trajectory patterns that form clearly separated regimes.
% We use this particularly clear example to formulate hypotheses about the underlying training dynamics.
% \Cref{app:analysis} provides quantitative support and examines how distinctly these patterns separate across settings.
% These hypotheses guide the design of our method (\Cref{sec:steering}), which we show can steer training trajectories towards improved disentanglement in larger models (\Cref{sec:experiments}).

% We study diffusion autoencoders for attribute disentanglement in a controlled setting and observe two recurring trajectory patterns. % that training trajectories organise around two regimes.
% Here, we see two clearly separated regimes.
% \Cref{app:analysis} presents further analysis and discusses settings where clear regimes may not appear.
% Our hypotheses for regime separation, nevertheless, motivate our method to improve disentanglement (\Cref{sec:steering}), which successfully generalises to larger models (\Cref{sec:experiments}).
% % Our hypotheses for this behaviour motivate mechanisms to improve disentanglement (\Cref{sec:steering}), which generalise to larger models (\Cref{sec:experiments}).

% The resulting hypotheses motivate our mechanisms for improving disentanglement in \Cref{sec}, and in \Cref{sec}, we show that these mechanisms successfully generalise to larger models.

\noindent\textbf{Setup.} 
We construct a U-Net with 3M parameters, fixed up and down-sampling convolutions, using the split-MLP encoder for $\IB(\cdot)$ with $N = 10$ tokens.
We train using the EDM loss-weighting \citep{karras2022elucidating} with $\lambda$ shifted by $-3$ via the loss in \Cref{eq:weighted_elbo_importance_sampling}, with the band $[\lambda_{\min}, \lambda_{\max}] = [-5, 12]$, on Shapes3D \citep{kim2018disentangling}.
At regular intervals, we evaluate reconstruction (LPIPS \citep{zhang2018unreasonable}) against disentanglement in $\z$, measuring DCI disentanglement \citep{eastwood2018a}, FactorVAE \citep{kim2018disentangling} and MIG \citep{chen2018isolating}.
Details are in \Cref{app:analysis_setup}.

\noindent\textbf{Why do we see regimes?}
In \Cref{fig:shapes3d_regimes}, we see that training trajectories separate into two \textit{regimes} centred around the median final disentanglement.
Similar regimes are seen with the other loss-weightings (\Cref{app:traj_details}), and these trajectories cluster into two groups (\Cref{app:clustering}).
Both regimes achieve similar final reconstruction, but the \textit{disentanglement regime} yields higher final disentanglement, while the \textit{reconstruction regime} prioritises reconstruction before disentanglement.
% More mature VQ-VAEs and larger U-Nets collapse trajectories into the reconstruction regime (\Cref{fig:ablations_traj}).
Further, regimes correspond to different internal representations; \Cref{app:cross_attn_details} shows that the disentanglement regime produces cross-attention maps aligned with ground-truth factors, whereas those in the reconstruction regime appear diffuse.
This points to a mechanism which bypasses cross-attention; prior work shows that skip connections provide a high-bandwidth bypass around the U-Net trunk \citep{jun2025disentangling,ma2024surprising}.
We test this by zeroing out skip connections and using \textit{SkipDropout} \citep{jun2025disentangling} (\Cref{app:zero_skip}).
% We observe that while this improves disentanglement, it significantly worsens reconstruction.
This improves disentanglement but substantially worsens reconstruction.

\textit{\textbf{Hypothesis 1 (H1):} Since disentanglement via \Cref{eq:weighted_elbo_importance_sampling} is unsupervised, the U-Net may exploit shortcut pathways via skip-connections to optimise reconstruction without organising factors in $\z$.}

\noindent\textbf{Why do models commit early to a regime?}
We observe that trajectories diverge into their regimes early in training.
In \Cref{fig:shapes3d_regimes}, we highlight this with a reference point at $576$K / $3$M images seen, where reconstruction-regime models are already near-optimal in reconstruction but substantially worse in disentanglement, whereas models in the disentanglement regime improve gradually along both axes.
Prior work shows that diffusion training imposes an information bottleneck over noise levels \citep{yang2024diffusion}; high-noise levels make reconstruction difficult, increasing reliance on clean conditioning signals, while low-noise levels recover fine-grained details as training progresses \citep{tishby2000information,shwartz2017opening,wang2023diffusion,higgins2017betavae} -- widely observed in learning curves \cite{karras2022elucidating,peebles2023scalable,ma2024sit}.
In early training, specific noise-levels may shape the relative pace of learning reconstruction vs disentanglement.
We test this by retraining the current setup, modifying $[\lambda_{\min}, \lambda_{\max}]$ over 10 seeds (\Cref{app:early_noise_level_exposure}).
By $320$K images seen, the original band $[-5,12]$ yields fast reconstruction with competitive disentanglement.
Removing high-noise levels with $[0,12]$ delays reconstruction substantially, but does not collapse disentanglement.
Conversely, only high-noise $[-5,0]$ shows relatively strong early disentanglement but poor reconstruction.

\textit{\textbf{Hypothesis 2 (H2):} Early exposure to specific noise levels can accelerate regime selection by changing how quickly reconstruction improves relative to disentanglement.}

\newpage

\section{\textsc{SteeringDRL}: Steering Unsupervised Diffusion Representation Learning}
\label{sec:steering}

\begin{figure}[t]
    \centering

    \begin{subfigure}[t]{0.56\textwidth}
        \centering
        \includegraphics[
            width=\linewidth
        ]{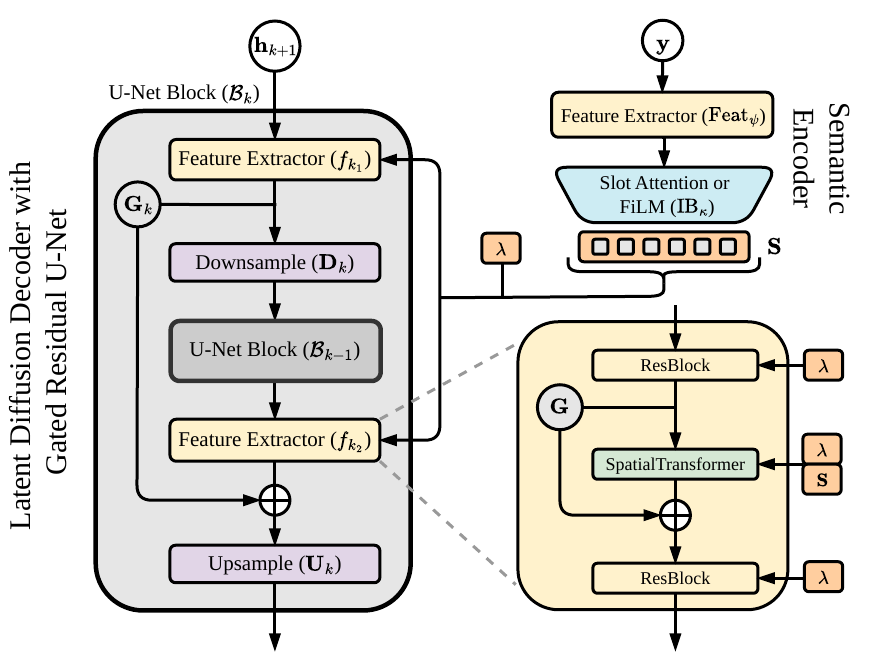}
        \caption{Diffusion Autoencoder w/ Gated Residual U-Net.}
        \label{fig:architecture_unet}
    \end{subfigure}
    \hfill
    \begin{subfigure}[t]{0.42\textwidth}
        \centering
        \includegraphics[
            width=\linewidth
        ]{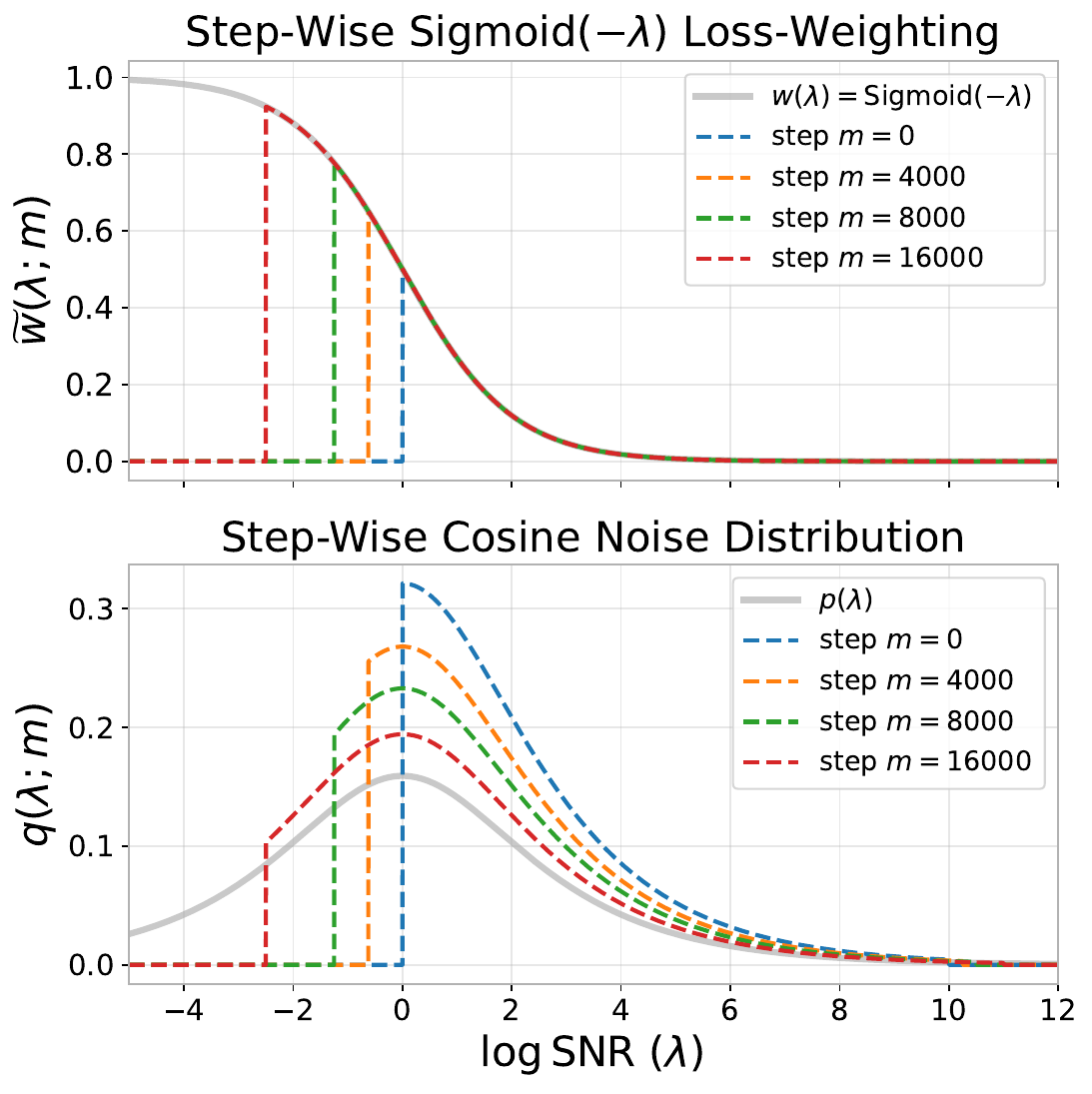}
        \caption{$\logsnr$ curriculum.}
        \label{fig:attr_curriculum}
    \end{subfigure}

    \caption{
    \textbf{Inductive biases in \textsc{SteeringDRL}.}
    \textbf{(a)} We define our U-Net recursively. We use gated residuals instead of skip connections, and gates in the SpatialTransformer.
    $\IB(\cdot)$ uses an amortised semantic encoder with FiLM \citep{perez2018film} for attribute disentanglement (c.f. \Cref{app:film_encoder}), and SlotAttention \citep{locatello2020object} for object-centric learning.
    \textbf{(b)} The $\logsnr$ curriculum linearly widens the sampled $\logsnr$ band from $[0, 10]$ to $[-5, 12]$ over the first $32$K steps in training.
    }
    \label{fig:architecture}
\end{figure}

\Cref{sec:analysis} shows that diffusion autoencoders can follow distinct training trajectories.
We hypothesise that shortcut pathways in the U-Net (\textbf{H1}) and noise-level exposure in early training (\textbf{H2}) contribute to this behaviour.
We introduce two mechanisms that target these factors to improve disentanglement.

\subsection{Gated Residual U-Net}
\label{sec:res_unet}
U-Nets pass encoder features to the decoder via skip connections, giving the decoder access to high-frequency details and reducing reliance on the semantic tokens \citep{li2025back,chen2025towards}.
To target \textbf{H1}, we replace concatenative skips with gated residuals \citep{hoogeboom2024simpler}, defining our U-Net block $\mathcal{B}(\cdot)$ recursively at level $k$ as
\begin{equation}
    \mathcal{B}_k(\mathbf{h}_{k + 1}) =
    \mathbf{U}_k 
    \Big(
    (f_{2,k} \circ \mathcal{B}_{k-1} \circ \mathbf{D}_k)(\mathbf{h}_k) \textcolor{blue}{\; \oplus \; \mathbf{G}_k(\mathbf{h}_k)} 
    \Big),
    \qquad
    \mathbf{h}_k = f_{1,k}(\mathbf{h}_{k + 1}),
\end{equation}
where $\mathbf{D}(\cdot)$ and $\mathbf{U}(\cdot)$ are down and upsampling blocks, $f_{1}(\cdot)$ and $f_{2}(\cdot)$ are feature extractors conditioned on $\mathbf{S}$ via cross-attention and on $\lambda$ via Fourier features, and $\mathbf{G}(\cdot)$ is a learnable gate \citep{bachlechner2021rezero,zhang2019fixup,saharia2022image},
\begin{equation}
    \mathbf{G}(\mathbf{h}) = \mathrm{Softplus}(\xi)\odot \mathbf{h}.
\end{equation}
We initialise $\xi$ such that $\mathrm{Softplus}(\xi) \approx 0$ to suppress shortcuts early in training, when we observe regime commitment to occur. 
This introduces an information bottleneck on the skip pathways \citep{tishby2000information,shwartz2017opening}; skip information enters the decoder as a gated residual rather than as a separate stream via concatenation.
This encourages greater reliance on the U-Net trunk and the conditioning signals in $\tokens$.
Similarly, we gate residual connections inside spatial transformer blocks in $f_{1}(\cdot)$ and $f_{2}(\cdot)$ \citep{touvron2021going,chen2026attnres},
\begin{equation}
\mathrm{SpatialTransformer}(\mathbf{h}, \tokens; \lambda) = F(\mathbf{h}, \tokens; \lambda) \textcolor{blue}{ \; \oplus \; \mathbf{G}(\mathbf{h})},
\end{equation}
 where $F(\cdot)$ is the transformer trunk.
This conditions on $\tokens$ via cross-attention to encourage disentanglement \citep{yang2024diffusion,wu2023slotdiffusion,jun2025disentangling,jiang2023object}, where queries are derived from $\mathbf{h}$ and keys/values from $\tokens$.
To stabilise representations, we normalise cross-attention queries using AdaLN-Zero modulated by $\lambda$ \citep{peebles2023scalable}.

\begin{table}[t]
    \centering
    % \caption{
    % \textbf{In \textsc{SteeringDRL}, the architecture and optimisation curriculum together improve disentanglement and stability.}
    % We progressively add components to the EncDiff U-Net \citep{yang2024diffusion} ($N{=}10$) on Shapes3D (10 seeds).
    % \textsc{SteeringDRL} matches the best standard U-Net variant while reducing variance.
    \caption{
    \textbf{Components in \textsc{SteeringDRL} interact to improve stability and retain competitive disentanglement.}
    Progressive ablations on Shapes3D using the EncDiff U-Net \citep{yang2024diffusion} ($N=10$; 10 seeds).
    Macro Mean and Macro SD average the means and stds. across metrics, respectively.
    The curriculum we use here is shown in \Cref{fig:attr_curriculum}.
    }
    % \scriptsize
    \resizebox{\linewidth}{!}{
    \begin{tabular}{lccccc}
    \toprule
    \textbf{Model Variant} & \textbf{DCI} $\uparrow$ & \textbf{FactorVAE} $\uparrow$ & \textbf{MIG} $\uparrow$ & \textbf{Macro Mean} $\uparrow$ & \textbf{Macro SD} $\downarrow$ \\
    \midrule
    U-Net + Split-MLP (EncDiff \citep{yang2024diffusion}) 
      & $0.828 \pm 0.099$ & $0.921 \pm 0.093$ & $0.218 \pm 0.087$ & $0.656$ & $0.093$ \\
    \quad + $\mathrm{Sigmoid}(-\lambda)$ Loss-Weighting
      & $\mathbf{0.932 \pm 0.077}$ & $\mathbf{0.971 \pm 0.056}$ & $\mathbf{0.399 \pm 0.102}$ & $\mathbf{0.767}$ & $0.078$ \\
    \quad + $\logsnr$ Curriculum 
      & $0.853 \pm 0.060$ & $0.917 \pm 0.063$ & $0.299 \pm 0.066$ & $0.690$ & $\mathbf{0.063}$ \\
    \midrule
    U-Net + Amortised Encoder (\textit{Baseline})
      & $0.832 \pm 0.087$ & $0.904 \pm 0.060$ & $0.352 \pm 0.098$ & $0.696$ & $0.082$ \\
    \quad + Gated Residual U-Net
      & $0.826 \pm 0.082$ & $0.885 \pm 0.050$ & $0.353 \pm 0.093$ & $0.688$ & $0.075$ \\
    \quad + $\mathrm{Sigmoid}(-\lambda)$ Loss-Weighting 
      & $0.901 \pm 0.065$ & $0.977 \pm 0.034$ & $\mathbf{0.386 \pm 0.092}$ & $0.755$ & $0.064$ \\
    \quad + $\logsnr$ Curriculum  (\textbf{\textsc{SteeringDRL}})
      & $\mathbf{0.926 \pm 0.043}$ & $\mathbf{0.991 \pm 0.017}$ & $0.353 \pm 0.056$ & $\mathbf{0.757}$ & $\mathbf{0.039}$ \\
    \bottomrule
    \end{tabular}
    }
    \label{tab:ablations_shapes3d}
\end{table}

\begin{figure}[t]
    \centering
    % \frame{
    \includegraphics[width=\linewidth, trim={0mm 0mm 0mm 0mm}, clip]{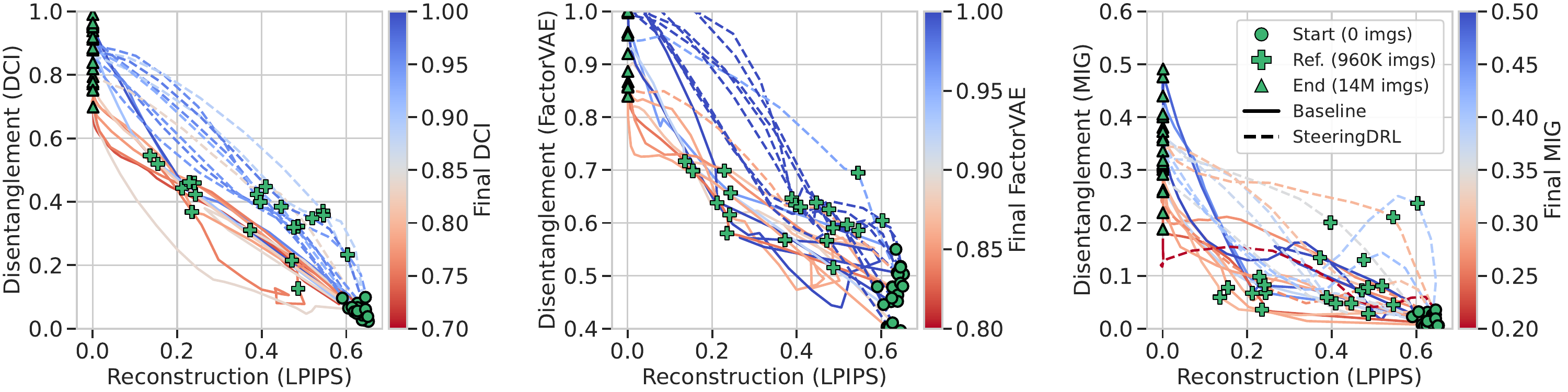}
    % }
    \caption{
    \textbf{\textsc{SteeringDRL} steers training trajectories.}
    Trajectories for models in \Cref{tab:ablations_shapes3d} show that \textsc{SteeringDRL} (dashed lines) can shift the \textit{baseline} (solid lines) towards the disentanglement regime; reconstruction is slower (\textsc{SteeringDRL} ref. points lag behind \textit{baseline} for reconstruction), trajectories shift upwards, final disentanglement is higher (dashed lines are bluer than solid lines).
    }
    \label{fig:ablations_traj}
\end{figure}

\begin{figure}[!htbp]
    \centering
    % \frame{
    \includegraphics[width=\linewidth, trim={0mm 0mm 0mm 0mm}, clip]{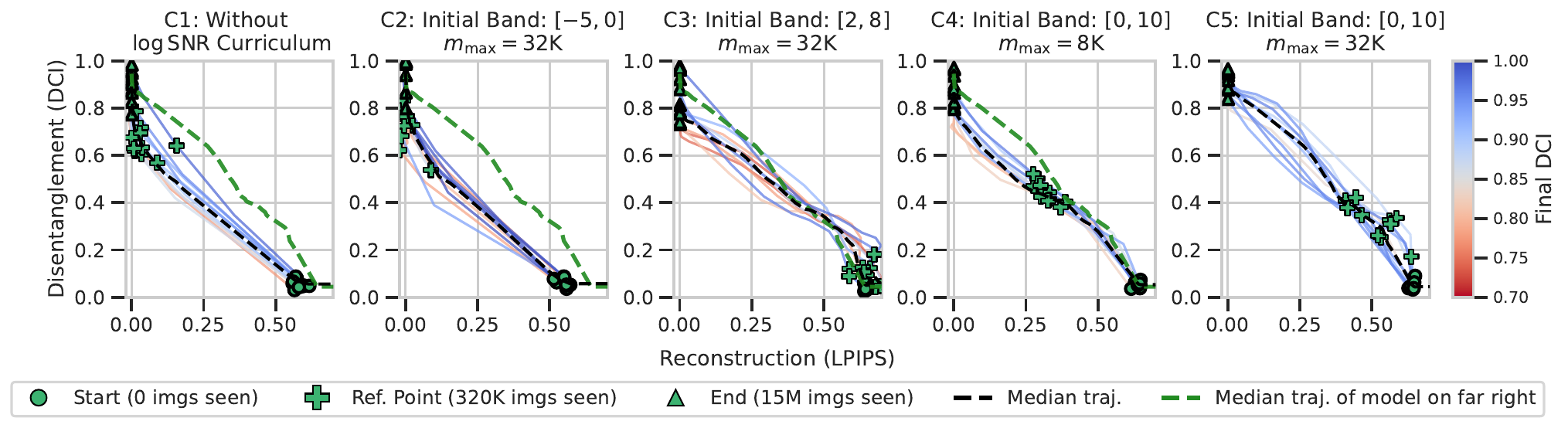}
    
    % }
    \caption{
    \textbf{$\logsnr$ curriculum selects the regime.}
    With architecture fixed (C1: Gated Residual U-Net + $\mathrm{sigmoid}(-\lambda)$ loss-weighting, \Cref{tab:ablations_shapes3d}), we vary the curriculum (C2 to C5).
    Starting from high-noise (C2) biases trajectories toward the reconstruction regime, whereas starting from low-noise (C5) steers them into the disentanglement regime.
    This shows that regime selection occurs early in training, and that curricula can be designed to follow different regimes whilst attaining similar disentanglement.
    C5 beats C2 in disentanglement and reduces variance across runs (\Cref{app:curriculum_ablations}).
    }    
    \label{fig:logsnr_ablations_traj}
\end{figure}

% \newpage

\subsection{$\logsnr$ Curriculum} 
\label{sec:lambda_curriculum}
To address \textbf{H2}, we control early noise-level exposure by training on a bounded $\logsnr$ band and gradually widening it to the full objective.
At step \(m\), we sample the active band \([\lambda_L(m),\lambda_U(m)]\) that widens linearly to \([\lambda_{\min},\lambda_{\max}]\) while \(m \le m_{\max}\):
\begin{equation}
\lambda_L(m) = \lambda_L(0) + \frac{m}{m_{\max}}\big(\lambda_{\min}-\lambda_L(0)\big), \qquad
\lambda_U(m) = \lambda_U(0) + \frac{m}{m_{\max}}\big(\lambda_{\max}-\lambda_U(0)\big),
\end{equation} 
with initial band $[\lambda_L(0), \lambda_U(0)]$ at $m = 0$.
This defines a step-wise sampling density
\begin{equation}
    q(\lambda; m) = \frac{p(\lambda)}{{Z(m)}} \cdot \mathbf{1}\{\lambda \in [\lambda_L(m), \lambda_U(m)]\}, \qquad Z(m) = t_L(m) - t_U(m)
\end{equation}
for noise schedules induced by uniform sampling \(t \sim \mathcal{U}(0,1)\), where $t_U(m) = f_\lambda^\inv(\lambda_U(m))$ and $t_L(m) = f_\lambda^\inv(\lambda_L(m))$.
Sampling \(\lambda\) from \(q(\lambda; m)\) rather than $p(\lambda)$, requires the importance correction \(w(\lambda) / q(\lambda; m) = Z(m)\, w(\lambda) / p(\lambda)\) to our objective in \Cref{eq:weighted_elbo_importance_sampling}:
\begin{equation}
\mathcal{L}(\x, \tokens; m)
=
\frac{1}{2}\,
\mathbb{E}_{\veps \sim \normal(0,\identity),\, \lambda \sim q(\lambda; m)}
\!\left[
Z(m)\,\frac{w(\lambda)}{p(\lambda)}\,
\big\|
\veps - \hat{\veps}_\theta(\x_\lambda, \tokens; \lambda)
\big\|_2^2
\right],
\label{eq:curriculum_importance_sampling}
\end{equation}
where the effective step-wise loss-weighting at step $m$ is,
\begin{equation}
    \widetilde{w}(\lambda; m) = w(\lambda) \cdot \mathbf{1}\{\lambda \in [\lambda_L(m), \lambda_U(m)]\}.
\end{equation}
This directly extends to the objectives in \Cref{eq:weighted_elbo_deriv} and $\x_0$-prediction (\Cref{app:curriculum_deriv}).

\section{Experiments}
\label{sec:experiments}
We evaluate \textsc{SteeringDRL} for disentanglement in two settings.
In \Cref{sec:attr_dis_res}, we study attribute disentanglement,
%on popular benchmarks
seeing how the trends in \Cref{sec:analysis} apply to larger U-Nets.
In \Cref{sec:ocl_res}, we apply our approach to spatial disentanglement on synthetic and real-world datasets, analysing generalisation when $\IB(\cdot)$ uses Slot Attention \citep{locatello2020object} and $\Feat(\cdot)$ is pretrained.
Setup details are in \Cref{app:exp_setup}.

\begin{table*}[t]
\centering
% \caption{
% \textbf{\textsc{SteeringDRL} improves attribute disentanglement across datasets.}
% We achieve performance on Cars3D and Shapes3D and remain competitive on MPI3D-toy.
% Metrics are computed over 10 seeds.
% We re-run the EncDiff baseline \citep{yang2024diffusion}, and mark baselines taken from literature in grey.
% }
\caption{
\textbf{\textsc{SteeringDRL} improves attribute disentanglement over matched EncDiff baselines on Cars3D and Shapes3D and remains competitive on MPI3D-toy.}
% FactorVAE and DCI scores on Cars3D, Shapes3D, and MPI3D-toy.
Methods marked in blue are run under matched conditions: EncDiff and \textsc{SteeringDRL} use matched U-Net capacity, images seen in training and evaluation protocols, over 10 seeds.
Baselines taken from literature are grey.
}
\label{tab:main_result}
\small
\resizebox{\linewidth}{!}{
\begin{tabular}{c l cc cc cc}
\toprule
& &
\multicolumn{2}{c}{{\textbf{Cars3D}}} &
\multicolumn{2}{c}{{\textbf{Shapes3D}}} &
\multicolumn{2}{c}{{\textbf{MPI3D-toy}}} \\
\cmidrule(lr){3-4}\cmidrule(lr){5-6}\cmidrule(lr){7-8}
\textbf{$N$} & \textbf{Method} & {\textbf{FactorVAE}} $\uparrow$ & {\textbf{DCI}} $\uparrow$ &
{\textbf{FactorVAE}} $\uparrow$ & {\textbf{DCI}} $\uparrow$ &
{\textbf{FactorVAE}} $\uparrow$ & {\textbf{DCI}} $\uparrow$ \\
\midrule

% \multirow{4}{*}{$-$}
\rowcolor{gray!30}
-
& DisDiff \citep{yang2023disdiff}        
& $0.976 \pm 0.018$ & $0.232 \pm 0.019 $ 
& $0.902 \pm 0.043$ & $0.723 \pm 0.013$ 
& $0.617 \pm 0.070$ & $0.337 \pm 0.057$ \\

\rowcolor{gray!30}
-
& FDAE \citep{Wu_Zheng_2024}               
& $0.918 \pm 0.027$ & $0.232 \pm 0.418$ 
& $0.987 \pm 0.023$ & $0.917 \pm 0.038$ 
& -- & -- \\

\rowcolor{gray!30}
-
& DyGA \citep{jun2025disentangling} 
& $0.941 \pm 0.002$ & $0.414 \pm 0.013$ 
& $1.000 \pm 0.000$ & $0.938 \pm 0.001$ 
& $0.930 \pm 0.004$ & $0.627 \pm 0.002$ \\

\rowcolor{gray!30}
-
& Comp. IB \citep{jung2025disentangled} 
& $0.877 \pm 0.089$ & $0.365 \pm 0.073$ 
& $0.975 \pm 0.059$ & $0.837 \pm 0.105$ 
& $0.708 \pm 0.060$ & $0.458 \pm 0.052$ \\

\midrule

\rowcolor{blue!10}
% \multirow{2}{*}{$20$}
20
& EncDiff \citep{yang2024diffusion} 
& $0.765 \pm 0.037$ & $0.258 \pm 0.024$ 
& $0.995 \pm 0.010$ & $0.916 \pm 0.041$ 
& $\mathbf{0.899 \pm 0.024}$ & $0.670 \pm 0.034$ \\

\rowcolor{blue!10}
20
& \textbf{\textsc{SteeringDRL}} 
& $\mathbf{1.000 \pm 0.001}$ & $\mathbf{0.624 \pm 0.071}$ 
& $\mathbf{1.000 \pm 0.000}$ & $\mathbf{0.954 \pm 0.030}$ 
& $0.870 \pm 0.021$ & $\mathbf{0.690 \pm 0.031}$ \\

\midrule

\rowcolor{blue!10}
% \multirow{2}{*}{10}
10
& EncDiff \citep{yang2024diffusion} 
& $0.714 \pm 0.035$ & $0.208 \pm 0.035$
& $0.921 \pm 0.093$ & $0.828 \pm 0.099$
& $\mathbf{0.805 \pm 0.061}$ & $\mathbf{0.627 \pm 0.050}$ \\

\rowcolor{blue!10}
10
& \textbf{\textsc{SteeringDRL}} 
& $\mathbf{0.987 \pm 0.031}$ & $\mathbf{0.527 \pm 0.112}$ 
& $\mathbf{0.991 \pm 0.017}$ & $\mathbf{0.926 \pm 0.043}$
& $0.800 \pm 0.043$ & $0.602 \pm 0.039$	\\

\bottomrule
\end{tabular}
}
\end{table*}
\begin{figure}[t]
    \begin{subfigure}[t]{0.445\textwidth}
        \includegraphics[width=\linewidth]{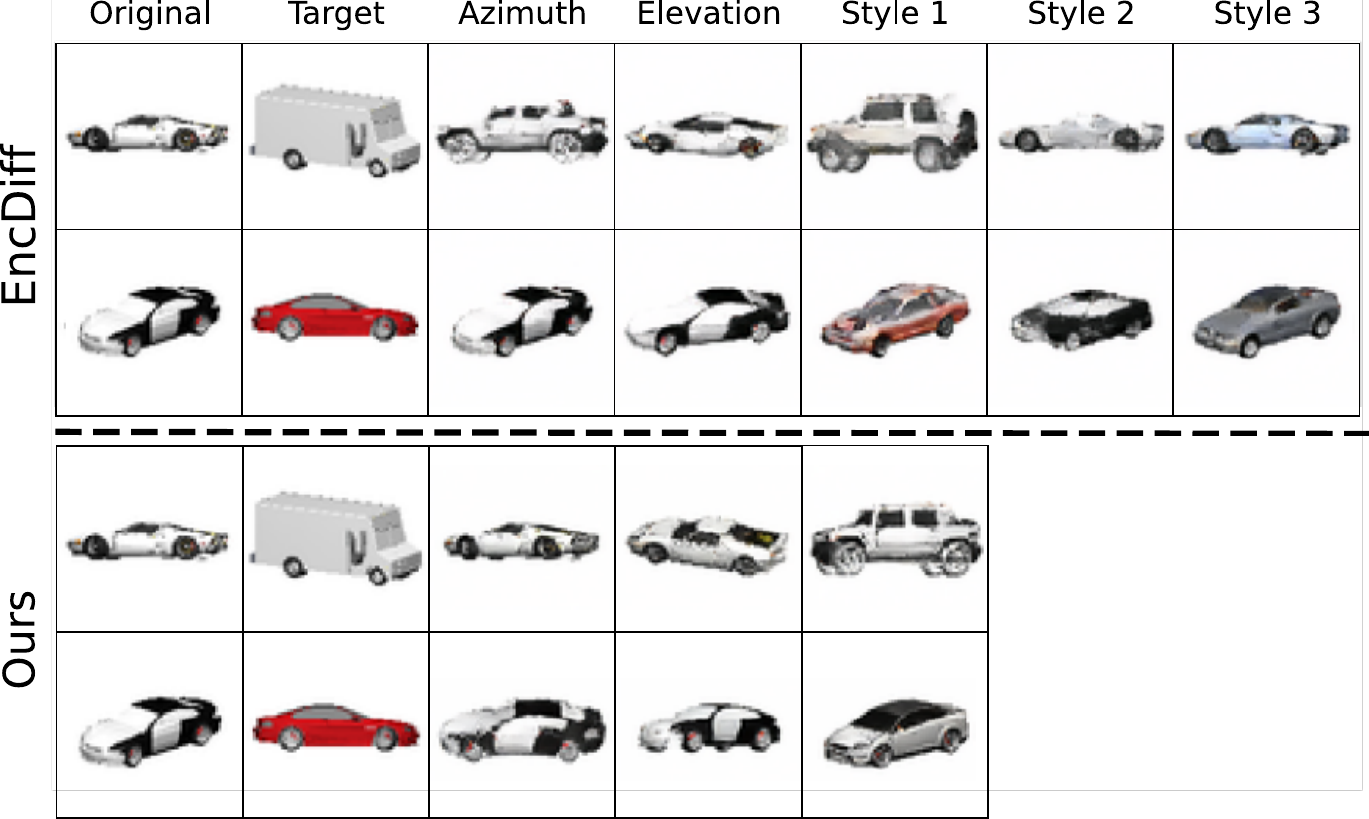}
    \end{subfigure}
    \hfill
    \begin{subfigure}[t]{0.5\textwidth}
        \includegraphics[width=\linewidth, trim={25mm 10mm 30mm 15mm}]{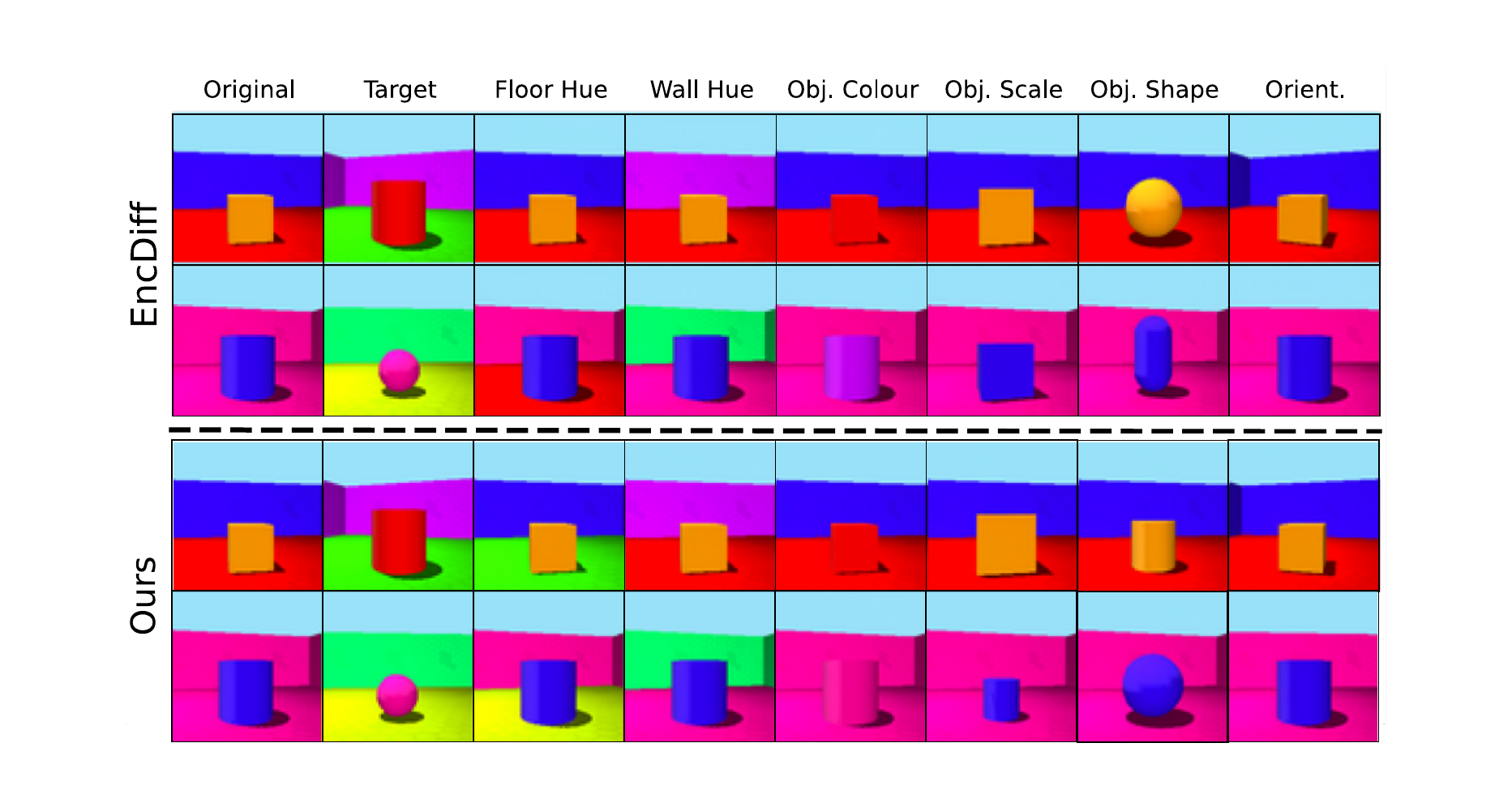}
    \end{subfigure}
    \caption{
    \textbf{Latent feature discovery and controlled image editing.}
    We manually select the latents for true attributes in the target image and swap them with the source.
    Compared to EncDiff, \textsc{SteeringDRL} better isolates Cars3D attributes and produces more consistent edits on Shapes3D.
    % the challenging \texttt{Style/Object Type} factor in a single variable, producing more faithful edits of \texttt{Azimuth} and \texttt{Elevation}.
    % On Shapes3D, \textsc{SteeringDRL} successfully performs all attribute edits, while EncDiff can often be inconsistent.
    % with \texttt{Floor Hue} and \texttt{Object Shape}.
    }
    \label{fig:attr_dis_main}
\end{figure}

\subsection{Attribute Disentanglement}
\label{sec:attr_dis_res}
We study unsupervised attribute disentanglement on \textit{Shapes3D} \citep{kim2018disentangling}, \textit{Cars3D} \citep{reed2015deep}, and \textit{MPI3D-Toy} \citep{gondal2019on}, where scalars in $\z$ should capture the true generative factors of the dataset.
We follow the evaluation protocol from \Cref{sec:analysis}, now building on top of a better parameterised U-Net and a more mature VQ-VAE checkpoint, with both matched in capacity and training to the EncDiff baseline \citep{yang2024diffusion}.
We use the loss in \Cref{eq:weighted_elbo_importance_sampling} and the $\logsnr$ curriculum as defined in \Cref{eq:curriculum_importance_sampling}.

\noindent\textbf{Inductive Biases in \textsc{SteeringDRL}.}
In \Cref{tab:ablations_shapes3d}, we progressively apply our inductive biases to EncDiff \citep{yang2024diffusion} on Shapes3D.
We replace the split-MLP with a parameter-efficient amortised encoder (\Cref{app:film_encoder}), matching EncDiff performance.
With EncDiff, the sigmoid loss-weighting has the highest Macro Mean;
% improves average disentanglement but increases variance; 
adding our curriculum does not consistently improve individual metrics.
With our gated residual U-Net, \textsc{SteeringDRL} has comparable Macro Means and substantially lower Macro SD, although MIG does not improve.
% these components maintain strong performance with substantially reduced variance.
\Cref{fig:ablations_traj} shows the corresponding training trajectories.
When using a larger U-Net and a more mature VQ-VAE checkpoint than \Cref{sec:analysis}, the marked baseline here no longer exhibits regime separation and instead follows the reconstruction regime.
% In this case, 
Despite this harder setting, adding \textsc{SteeringDRL} steers trajectories toward the disentanglement regime, improving disentanglement without sacrificing reconstruction.
In \Cref{fig:logsnr_ablations_traj}, we show that curricula can select different regimes, with the disentanglement-oriented trajectory showing better performance under the settings tested.
% isolate the role of the curriculum and show that it is primarily responsible for regime selection.
Varying the initial band changes the training trajectory, while $m_{\max}$ determines the convergence speed within a regime.
Overall, this shows that the curriculum influences regime selection, while our architectural bottlenecks and loss-weighting improve stability within it, as further supported by the skip-connection ablations in \Cref{app:zero_skip}.

\noindent\textbf{Benchmark Datasets.} 
In \Cref{tab:main_result}, we compare \textsc{SteeringDRL} from \Cref{tab:ablations_shapes3d} to a capacity-matched EncDiff.
On Cars3D, \textsc{SteeringDRL} improves disentanglement over EncDiff at $N \in \{10,20\}$, while  
for Shapes3D we additionally reduce variance across seeds.
We remain competitive on both metrics for MPI3D-toy while consistently reducing variance.
\Cref{fig:attr_dis_main} illustrates the differences in disentanglement at $N = 10$.
On Cars3D, \textsc{SteeringDRL} better isolates the challenging \texttt{Style/Object Type} factor, which EncDiff spreads over multiple latents.
On Shapes3D, \textsc{SteeringDRL} recovers all ground-truth factors, while EncDiff often confuses \texttt{Floor Hue}, \texttt{Wall Hue}, and \texttt{Object Shape}.

% % In \Cref{tab:main_result}, we compare \textsc{SteeringDRL} from \Cref{tab:ablations_shapes3d} to prior diffusion-based methods.
% In \Cref{tab:main_result}, we compare \textsc{SteeringDRL} from \Cref{tab:ablations_shapes3d} to a capacity-matched EncDiff baseline, with results from other works provided as context.
% For $N = 20$, our model outperforms prior work on Cars3D and Shapes3D in both FactorVAE and DCI, and shows lower variance across seeds.
% For $N = 10$, closer to the true number of generative factors, we see larger gains over EncDiff.
% On MPI3D-toy, we remain competitive on disentanglement whilst reducing seed sensitivity.
% % achieve the best DCI at $N = 20$ and remain competitive on other metrics.
% In \Cref{fig:attr_dis_main}, our improvements on Cars3D can be attributed to better isolation of the challenging \texttt{Style/Object Type} factor, which EncDiff spreads over multiple latents.
% On Shapes3D with $N=10$, our model recovers all ground-truth factors, while EncDiff often confuses \texttt{Floor Hue}, \texttt{Wall Hue}, and \texttt{Object Shape}.
% % \sout{Overall, our model improves disentanglement and stability, with latents better aligned to true generative factors.}

\subsection{Spatial Disentanglement with Object-Centric Learning}
\label{sec:ocl_res}

\begin{table}[t]
    \caption{
    \textbf{\textsc{SteeringDRL} improves segmentation and image fidelity with object-centric diffusion models.}
    Segmentation, reconstruction, and composition metrics on ClevrTex \citep{karazija2021clevrtex}.
    Methods marked in blue are run under the same conditions: Resnet34 for $\Feat(\cdot)$, SlotAttention for $\IB(\cdot)$, the same data splits and evaluation protocol over 5 seeds.
    Baselines taken from literature are marked in grey.
    }
    \resizebox{\linewidth}{!}{
    \begin{tabular}{lccccc}
        \toprule
        & \multicolumn{3}{c}{\textbf{Segmentation}}
        & \textbf{Recon.}
        & \textbf{Comp.} \\
        \cmidrule(lr){2-4} \cmidrule(lr){5-5} \cmidrule(lr){6-6} 

        \textbf{Method} 
        & \textbf{FG-ARI} $\uparrow$
        & \textbf{mIoU} $\uparrow$
        & \textbf{mBO} $\uparrow$
        & \textbf{LPIPS} $\downarrow$
        & \textbf{FID} $\downarrow$ \\
        \midrule

        \rowcolor{gray!30}
        SLATE$^{+}$ \citep{jiang2023object}
        & $70.71 \pm 3.35$
        & $52.96 \pm 1.40$
        & $54.90 \pm 1.15$
        & -- 
        & $69.23$ \\

        \rowcolor{gray!30}
        StableLSD (200 steps, DDPM) \citep{jiang2023object}
        & $64.41 \pm 8.53$
        & $62.52 \pm 1.88$
        & $63.93 \pm 1.93$
        & --
        & $29.53$ \\

        \rowcolor{gray!30}
        Modular IB \citep{jung2025disentangled}
        & $87.68$
        & $58.88$
        & $59.12$
        & --
        & -- \\

        \rowcolor{gray!30}
        MetaSlot \citep{liu2025metaslot}
        & $77.60 \pm 1.10$
        & $64.20 \pm 0.60$
        & $62.80 \pm 0.70$
        & --
        & -- \\

        \midrule

        \rowcolor{blue!10}
        SLATE \citep{singh2022illiterate}
        & $61.39 \pm 9.07$
        & $55.00 \pm 5.47$
        & $56.40 \pm 7.19$
        & $0.397 \pm 0.009$
        & $90.26 \pm 3.72$ \\

        \rowcolor{blue!10}
        SlotDiffusion (50 steps, DDIM) \citep{wu2023slotdiffusion}
        & $63.90 \pm 5.11$
        & $55.92 \pm 1.17$
        & $58.83 \pm 1.70$
        & $0.167 \pm 0.011$
        & $42.95 \pm 3.07$ \\

        \rowcolor{blue!10}
        \textbf{\textsc{SteeringDRL}} (50 steps, DDIM)
        & $\mathbf{74.39 \pm 4.12}$
        & $\mathbf{68.31 \pm 2.39}$
        & $\mathbf{69.69 \pm 2.07}$
        & $0.157 \pm 0.005$
        & $26.67 \pm 2.17$ \\

        \rowcolor{blue!10}
        \textbf{\textsc{SteeringDRL}} (200 steps, DDPM)
        & "
        & "
        & "
        & $\mathbf{0.147 \pm 0.005}$
        & $\mathbf{23.36 \pm 2.01}$ \\
        
        \bottomrule
    \end{tabular}
    }
    \label{tab:ocl_ablations_table}
\end{table}

\begin{figure}[t]
    \centering
    \begin{subfigure}{0.3\textwidth}
        \centering
        \includegraphics[width=\linewidth]{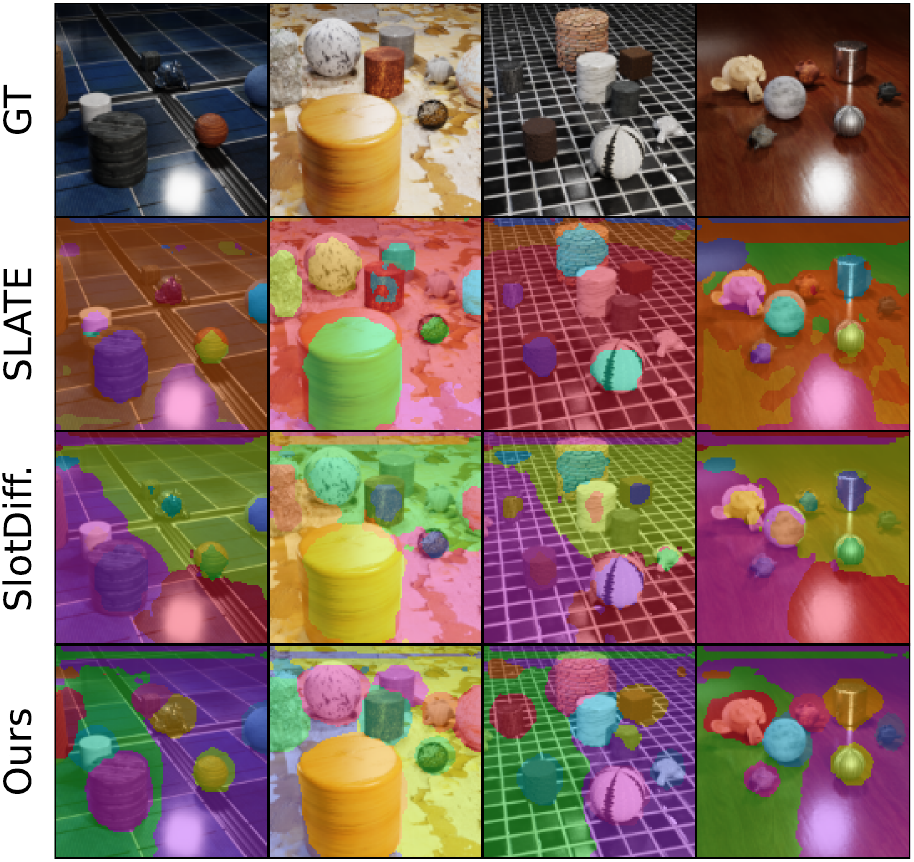}
        \caption{Segmentations.}
        \label{fig:ocl_clevrtex_segmentations}
    \end{subfigure}
    \hfill
    \begin{subfigure}{0.3\textwidth}
        \centering
        \includegraphics[width=\linewidth]{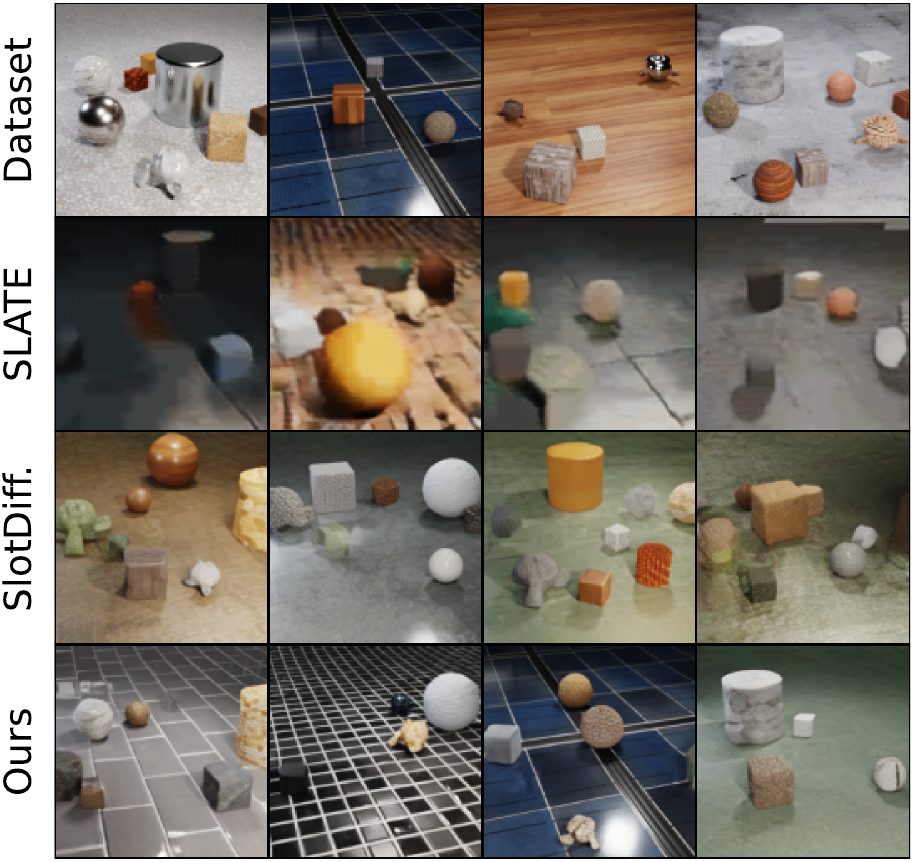}
        \caption{Random Compositions}
        \label{fig:ocl_clevrtex_compositions}
    \end{subfigure}
    \hfill
    \begin{subfigure}{0.365\textwidth}
        \centering
        % \frame{
        \includegraphics[width=\linewidth, trim={345mm 0mm 2mm 0mm}, clip]{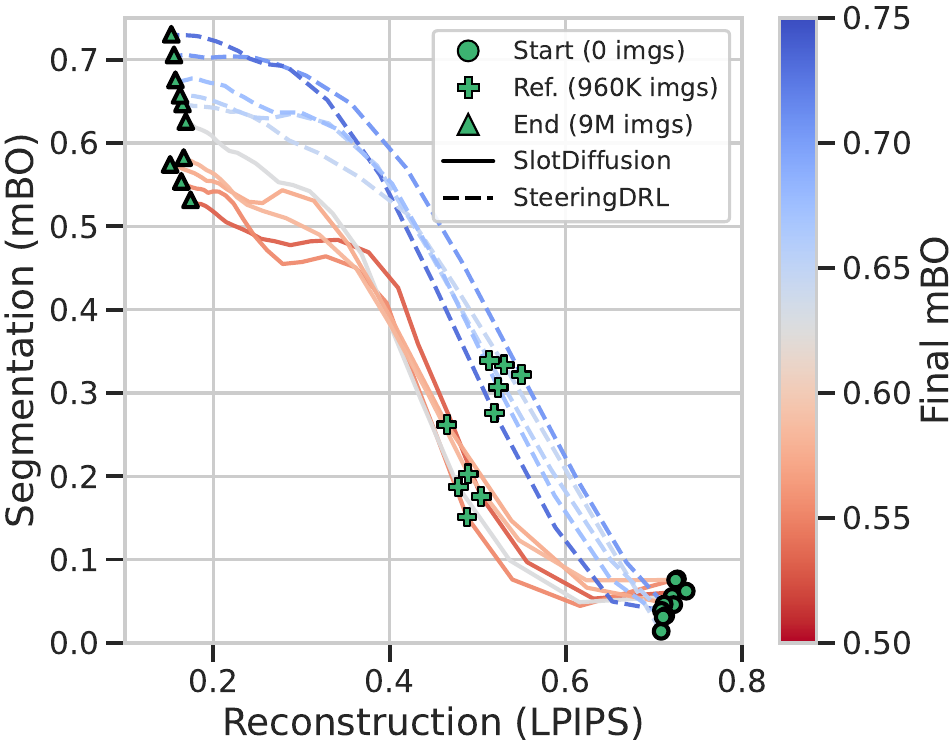}
        % }
        \caption{Training Trajectories.}
        \label{fig:ocl_clevrtex_trajectories}
    \end{subfigure}
    \caption{
    \textbf{Training trajectories support visual improvements.}
    (a) \textsc{SteeringDRL} discovers more objects with more complete segmentations than SlotDiffusion.
    (b) Compositions preserve object structure and background textures.
    (c) Training trajectories show trends analogous to \Cref{fig:ablations_traj}.
    % to attribute disentanglement (\Cref{fig:ablations_traj}), converging to higher segmentation scores than the SlotDiffusion baseline.
    }
\end{figure}

We study spatial disentanglement with OCL on \textit{ClevrTex} \citep{karazija2021clevrtex} and \textit{PascalVOC} \citep{Everingham10}, using segmentation metrics on $\mathcal{M}$ as a proxy for representation quality.
We build on SlotDiffusion \citep{wu2023slotdiffusion}, follow its evaluation protocol, and use the objective in \Cref{eq:weighted_elbo_deriv} with curriculum in \Cref{app:curriculum_deriv}.
We use an effective $\mathrm{sigmoid}(-\lambda)$ weighting, and a neutral initial band $[-2,2]$ with $m_{\max} = 200$k for the curriculum; ClevrTex uses $\x_0$-prediction (\Cref{app:xpred_weighting}) and PascalVOC uses $\veps$-prediction.

\noindent\textbf{Synthetic Dataset (ClevrTex).}
In \Cref{tab:ocl_ablations_table}, we compare \textsc{SteeringDRL} to matched baselines and see improvements on all metrics.
% outperforms prior models on mBO and mIoU, including those with additional object-centric inductive biases (ModularIB, MetaSlot).
For segmentation, improvements are driven by better identification of occluded objects and regions with low foreground-background contrast (\Cref{fig:ocl_clevrtex_segmentations}), where SlotDiffusion in particular often under-segments or misses objects.
Improved composition and reconstruction are linked to better preservation of background textures (\Cref{fig:ocl_clevrtex_compositions}).
Notably, \Cref{fig:ocl_clevrtex_trajectories} shows that our model exhibits training dynamics similar to the attribute disentanglement setting (\Cref{fig:ablations_traj}), while learning semantics faster and converging to higher segmentation scores than SlotDiffusion.
This is consistent with \textbf{H1} and \textbf{H2}; controlling early noise-level exposure and introducing a skip bottleneck can also steer larger models toward better spatial disentanglement.

\begin{table}
    \centering
    \caption{
    \textbf{\textsc{SteeringDRL} improves spatial disentanglement via object-centric learning on Pascal VOC.}
    Instance and class-wise segmentation metrics on PascalVOC \citep{Everingham10}.
    Methods marked blue are trained under the same conditions: DINO ViT \citep{oquab2023dinov2} for $\Feat(\cdot)$, SlotAttention for $\IB(\cdot)$, the same data splits and evaluation protocol over 3 seeds.
    Other baselines from literature are marked in grey.
    % Other results are taken from the cited papers.
    }
    \scriptsize
    % \resizebox{\linewidth}{!}{
    \begin{tabular}{lccccc}
    \toprule
    \textbf{Model} 
    & FG-ARI & mIoU$_i$ & mIoU$_c$ & mBO$_i$ & mBO$_c$ \\
    \midrule

    \rowcolor{gray!30}
    DINOSAUR MLP \cite{seitzer2023bridging} 
    & 24.6 & 39.1 & 41.0 & 39.7 & 41.2 \\

    \rowcolor{gray!30}
    DINOSAUR Trans. \cite{seitzer2023bridging} 
    & 23.1 & 42.0 & 47.5 & 43.2 & 47.8 \\

    \rowcolor{gray!30}
    StableLSD \cite{jiang2023object} 
    & 8.7 & 31.5 & 35.4 & 32.1 & 35.4 \\

    \midrule

    \rowcolor{blue!10}
    SA + DINO ViT \cite{wu2023slotdiffusion} 
    & 12.3 & -- & -- & 24.6 & 24.9 \\

    \rowcolor{blue!10}
    SLATE + DINO ViT \cite{singh2022illiterate,wu2023slotdiffusion} 
    & 15.6 & -- & -- & 35.9 & 41.5 \\

    \rowcolor{blue!10}
    SlotDiffusion \cite{wu2023slotdiffusion} %$\dagger$
    & 15.9 % $\pm$ 0.34
    & 43.2 % $\pm$ 0.33
    & 47.7 % $\pm$ 0.31
    & 47.2 % $\pm$ 0.37
    & 52.2 \\ % $\pm$ 0.46 \\

    \rowcolor{blue!10}
    \textbf{\textsc{SteeringDRL}} %$\dagger$
    & \textbf{17.6} % 9 $\pm$ 0.68}
    & \textbf{44.5} % $\pm$ 0.17}
    & \textbf{48.6} % $\pm$ 0.53}
    & \textbf{49.1} % $\pm$ 0.96}
    & \textbf{53.6} \\ % $\pm$ 1.54} \\
    
    \bottomrule
    \end{tabular}
    % }
    \label{tab:voc_results}
\end{table}

\begin{figure}[t]
    \centering
    \begin{subfigure}{0.68\textwidth}
        \centering
        \includegraphics[width=\linewidth, trim={0mm 0mm 0mm 0mm}, clip]{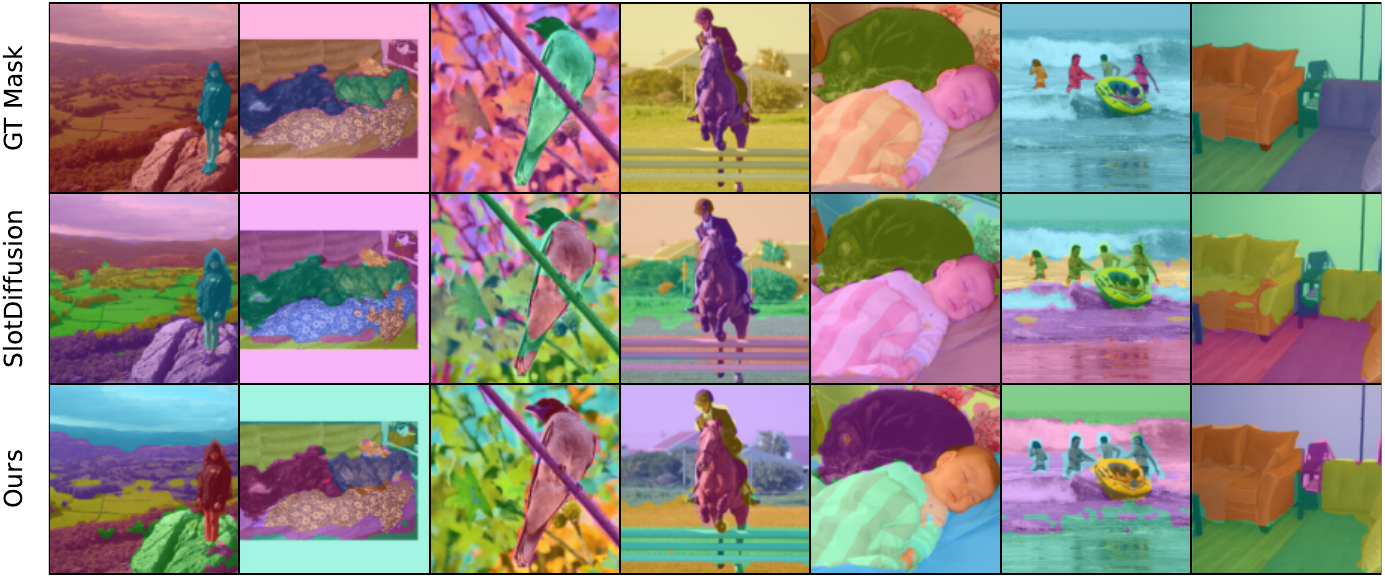}
        \caption{Segmentations.}
        \label{fig:ocl_voc_segmentations}
    \end{subfigure}
    \hfill
    \begin{subfigure}{0.31\textwidth}
        \centering
        % \frame{
        \includegraphics[width=\linewidth, trim={247mm 0mm 0mm 0mm}, clip]{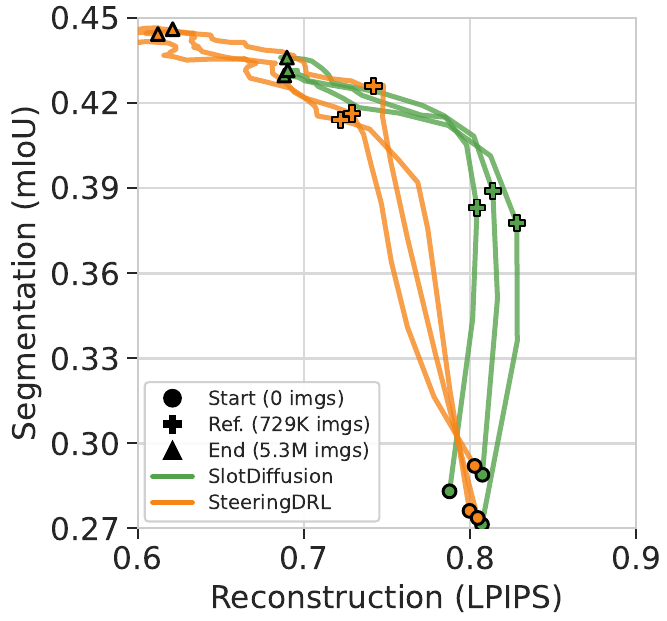}
        % }
        \caption{Training Trajectories.}
        \label{fig:ocl_voc_trajectories}
    \end{subfigure}
    \caption{
        \textbf{\textsc{SteeringDRL} accelerates object-centric learning with pretrained features.}
        \textbf{(a)} \textsc{SteeringDRL} better segments dominant objects while SlotDiffusion often undersegments.
        \textbf{(b)} \textsc{SteeringDRL} reaches higher segmentation and reconstruction earlier in training than SlotDiffusion.
    }
\end{figure}

\noindent\textbf{Training Efficiency with Pretrained Features (PascalVOC).}
We scale \textsc{SteeringDRL} to a much larger U-Net ($\sim138$M params), using a pretrained DINO ViT for $\Feat(\cdot)$.
In \Cref{tab:voc_results}, we improve over matched baselines on all metrics. % and remain competitive with prior methods on FG-ARI.
% These results are competitive with baselines using higher-capacity generative models (StableLSD) and non-generative approaches (DINOSAUR).
Our improvements stem from better segmentation of dominant objects that SlotDiffusion often merges (\Cref{fig:ocl_voc_segmentations}).
We also find meaningful unannotated regions.
\Cref{fig:ocl_voc_trajectories} shows different trajectories to ClevrTex despite using a similar curriculum.
On ClevrTex, where $\Feat(\cdot)$ is learned from scratch, representations and reconstructions are learned together progressively.
With pretrained features, semantics are already available, so learning shifts from feature discovery to slot binding, allowing \textsc{SteeringDRL} to converge faster than SlotDiffusion with better segmentations and reconstructions (see \Cref{app:pascal} for more results).
We see a similar pattern on the COCO dataset in the early stages of training (c.f. \Cref{app:coco}).

\section{Conclusion}
\label{sec:conclusion}
We studied a key challenge in diffusion autoencoders for unsupervised disentanglement: models with similar reconstruction quality can learn representations with different levels of disentanglement.
In a controlled setting, we observe training trajectories organising around two regimes that differ in how reconstruction and disentanglement develop.
% \textcolor{red}{Although regime separation is clearest in this setting, our hypotheses about these dynamics motivate \textsc{SteeringDRL}, which shapes trajectories in larger models.}
Our hypotheses about these dynamics motivate \textsc{SteeringDRL}, which steers models even when the baseline trajectories show no clear regime separation.
Against matched baselines, \textsc{SteeringDRL} is highly competitive on attribute disentanglement, with reduced seed sensitivity in many settings, and improves spatial disentanglement with OCL.
These results support \textbf{H1} and \textbf{H2} as useful design principles for disentanglement in diffusion autoencoders: restrict reconstruction shortcuts and control when models see specific noise levels.

\noindent\textbf{Limitations and Future Work.}
We use the standard VQ-VAE architecture from EncDiff and SlotDiffusion; although we control for its capacity in matched comparisons and analyse its sensitivity in \Cref{app:analysis}, testing alternative autoencoders \citep{zheng2026diffusion}, pixel-space generative models \citep{li2026back} and different conditioning mechanisms \citep{peebles2023scalable} remains future work.
We fix the curriculum's starting band and widening duration across datasets in each disentanglement setting, but new settings may require tuning; a robust unsupervised proxy may help with this \citep{kim2025denoising,kim2024adaptive}.
Our work supports \textbf{H1} and \textbf{H2} as factors affecting disentanglement, but there may be confounders that our analysis did not cover.

\begin{ack}
We thank Pavithra Manoj, Fabio De Sousa Ribeiro, Xiaodan Xing, Vishal Jain, and Raghav Mehta for their detailed and insightful feedback on early versions of this manuscript. We also thank Nairouz Shehata for her support with the tooling used to produce the figures. Finally, we thank Bernhard Kainz for his support and encouragement during this project.

R.R. is supported by EPSRC through a Doctoral Training Partnerships PhD Scholarship. A.K. is supported by the EPSRC Doctoral Prize. T.X. is supported through the Imperial College London UKRI Impact Acceleration Account EP/X52556X/1. B.G. was supported by the Royal Academy of Engineering as part of the Kheiron/RAEng Research Chair. The research was supported by the NIHR Imperial Biomedical Research Centre.

\end{ack}

\newpage

\bibliographystyle{plainnat}
\bibliography{references}

@article{kingma2021variational,
  title={Variational diffusion models},
  author={Kingma, Diederik and Salimans, Tim and Poole, Ben and Ho, Jonathan},
  journal={Advances in neural information processing systems},
  volume={34},
  pages={21696--21707},
  year={2021}
}

@article{karras2022elucidating,
  title={Elucidating the design space of diffusion-based generative models},
  author={Karras, Tero and Aittala, Miika and Aila, Timo and Laine, Samuli},
  journal={Advances in neural information processing systems},
  volume={35},
  pages={26565--26577},
  year={2022}
}

@article{ribeiro2025demystifying,
  title={Demystifying variational diffusion models},
  author={Ribeiro, Fabio De Sousa and Glocker, Ben and others},
  journal={Foundations and Trends{\textregistered} in Computer Graphics and Vision},
  volume={17},
  number={2},
  pages={76--170},
  year={2025},
  publisher={Now Publishers, Inc.}
}

@article{bengio2013representation,
  title={Representation learning: A review and new perspectives},
  author={Bengio, Yoshua and Courville, Aaron and Vincent, Pascal},
  journal={IEEE transactions on pattern analysis and machine intelligence},
  volume={35},
  number={8},
  pages={1798--1828},
  year={2013},
  publisher={IEEE}
}

@inproceedings{locatello2019challenging,
  title={Challenging common assumptions in the unsupervised learning of disentangled representations},
  author={Locatello, Francesco and Bauer, Stefan and Lucic, Mario and Raetsch, Gunnar and Gelly, Sylvain and Sch{\"o}lkopf, Bernhard and Bachem, Olivier},
  booktitle={international conference on machine learning},
  pages={4114--4124},
  year={2019},
  organization={PMLR}
}

@inproceedings{preechakul2022diffusion,
  title={Diffusion autoencoders: Toward a meaningful and decodable representation},
  author={Preechakul, Konpat and Chatthee, Nattanat and Wizadwongsa, Suttisak and Suwajanakorn, Supasorn},
  booktitle={Proceedings of the IEEE/CVF conference on computer vision and pattern recognition},
  pages={10619--10629},
  year={2022}
}

@article{zhang2022unsupervised,
  title={Unsupervised representation learning from pre-trained diffusion probabilistic models},
  author={Zhang, Zijian and Zhao, Zhou and Lin, Zhijie},
  journal={Advances in neural information processing systems},
  volume={35},
  pages={22117--22130},
  year={2022}
}

@inproceedings{wang2023infodiffusion,
  title={Infodiffusion: Representation learning using information maximizing diffusion models},
  author={Wang, Yingheng and Schiff, Yair and Gokaslan, Aaron and Pan, Weishen and Wang, Fei and De Sa, Christopher and Kuleshov, Volodymyr},
  booktitle={International conference on machine learning},
  pages={36336--36354},
  year={2023},
  organization={PMLR}
}

@article{yang2024diffusion,
  title={Diffusion model with cross attention as an inductive bias for disentanglement},
  author={Yang, Tao and Lan, Cuiling and Lu, Yan and Zheng, Nanning},
  journal={Advances in Neural Information Processing Systems},
  volume={37},
  pages={82465--82492},
  year={2024}
}

@inproceedings{jun2025disentangling,
  title={Disentangling Disentangled Representations: Towards Improved Latent Units via Diffusion Models},
  author={Jun, Youngjun and Park, Jiwoo and Choo, Kyobin and Choi, Tae Eun and Hwang, Seong Jae},
  booktitle={2025 IEEE/CVF Winter Conference on Applications of Computer Vision (WACV)},
  pages={3559--3569},
  year={2025},
  organization={IEEE}
}

@inproceedings{higgins2017betavae,
title={beta-{VAE}: Learning Basic Visual Concepts with a Constrained Variational Framework},
author={Irina Higgins and Loic Matthey and Arka Pal and Christopher Burgess and Xavier Glorot and Matthew Botvinick and Shakir Mohamed and Alexander Lerchner},
booktitle={International Conference on Learning Representations},
year={2017},
url={https://openreview.net/forum?id=Sy2fzU9gl}
}

@inproceedings{kim2018disentangling,
  title={Disentangling by factorising},
  author={Kim, Hyunjik and Mnih, Andriy},
  booktitle={International conference on machine learning},
  pages={2649--2658},
  year={2018},
  organization={PMLR}
}

@article{chen2016infogan,
  title={Infogan: Interpretable representation learning by information maximizing generative adversarial nets},
  author={Chen, Xi and Duan, Yan and Houthooft, Rein and Schulman, John and Sutskever, Ilya and Abbeel, Pieter},
  journal={Advances in neural information processing systems},
  volume={29},
  year={2016}
}

@article{yang2023disdiff,
  title={Disdiff: Unsupervised disentanglement of diffusion probabilistic models},
  author={Yang, Tao and Wang, Yuwang and Lv, Yan and Zheng, Nanning},
  journal={arXiv preprint arXiv:2301.13721},
  year={2023}
}

@article{jung2025disentangled,
  title={Disentangled Representation Learning via Modular Compositional Bias},
  author={Jung, Whie and Lee, Dong Hoon and Hong, Seunghoon},
  journal={arXiv preprint arXiv:2510.21402},
  year={2025}
}

@article{locatello2020object,
  title={Object-centric learning with slot attention},
  author={Locatello, Francesco and Weissenborn, Dirk and Unterthiner, Thomas and Mahendran, Aravindh and Heigold, Georg and Uszkoreit, Jakob and Dosovitskiy, Alexey and Kipf, Thomas},
  journal={Advances in neural information processing systems},
  volume={33},
  pages={11525--11538},
  year={2020}
}

@article{wu2023slotdiffusion,
  title={Slotdiffusion: Object-centric generative modeling with diffusion models},
  author={Wu, Ziyi and Hu, Jingyu and Lu, Wuyue and Gilitschenski, Igor and Garg, Animesh},
  journal={Advances in Neural Information Processing Systems},
  volume={36},
  pages={50932--50958},
  year={2023}
}

@article{wang2023diffusion,
  title={Diffusion models generate images like painters: an analytical theory of outline first, details later},
  author={Wang, Binxu and Vastola, John J},
  journal={arXiv preprint arXiv:2303.02490},
  year={2023}
}

@inproceedings{tang2023daam,
  title={What the daam: Interpreting stable diffusion using cross attention},
  author={Tang, Raphael and Liu, Linqing and Pandey, Akshat and Jiang, Zhiying and Yang, Gefei and Kumar, Karun and Stenetorp, Pontus and Lin, Jimmy and T{\"u}re, Ferhan},
  booktitle={Proceedings of the 61st Annual Meeting of the Association for Computational Linguistics (Volume 1: Long Papers)},
  pages={5644--5659},
  year={2023}
}

@inproceedings{rombach2022high,
  title={High-resolution image synthesis with latent diffusion models},
  author={Rombach, Robin and Blattmann, Andreas and Lorenz, Dominik and Esser, Patrick and Ommer, Bj{\"o}rn},
  booktitle={Proceedings of the IEEE/CVF conference on computer vision and pattern recognition},
  pages={10684--10695},
  year={2022}
}

@inproceedings{
kim2025denoising,
title={Denoising Task Difficulty-based Curriculum for Training Diffusion Models},
author={Jin-Young Kim and Hyojun Go and Soonwoo Kwon and Hyun-Gyoon Kim},
booktitle={The Thirteenth International Conference on Learning Representations},
year={2025},
url={https://openreview.net/forum?id=96GMFXsbJE}
}

@article{burgess2018understanding,
  title={Understanding disentangling in $\beta$-VAE},
  author={Burgess, Christopher P and Higgins, Irina and Pal, Arka and Matthey, Loic and Watters, Nick and Desjardins, Guillaume and Lerchner, Alexander},
  journal={arXiv preprint arXiv:1804.03599},
  year={2018}
}

@inproceedings{li2026back,
  title={Back to basics: Let denoising generative models denoise},
  author={Li, Tianhong and He, Kaiming},
  booktitle={Proceedings of the IEEE/CVF Conference on Computer Vision and Pattern Recognition},
  pages={36115--36125},
  year={2026}
}

@article{chen2018isolating,
  title={Isolating sources of disentanglement in variational autoencoders},
  author={Chen, Ricky TQ and Li, Xuechen and Grosse, Roger B and Duvenaud, David K},
  journal={Advances in neural information processing systems},
  volume={31},
  year={2018}
}

@article{brady2024interaction,
  title={Interaction asymmetry: A general principle for learning composable abstractions},
  author={Brady, Jack and von K{\"u}gelgen, Julius and Lachapelle, S{\'e}bastien and Buchholz, Simon and Kipf, Thomas and Brendel, Wieland},
  journal={arXiv preprint arXiv:2411.07784},
  year={2024}
}

@article{burgess2019monet,
  title={Monet: Unsupervised scene decomposition and representation},
  author={Burgess, Christopher P and Matthey, Loic and Watters, Nicholas and Kabra, Rishabh and Higgins, Irina and Botvinick, Matt and Lerchner, Alexander},
  journal={arXiv preprint arXiv:1901.11390},
  year={2019}
}

@article{piantadosi2016logical,
  title={The logical primitives of thought: Empirical foundations for compositional cognitive models.},
  author={Piantadosi, Steven T and Tenenbaum, Joshua B and Goodman, Noah D},
  journal={Psychological review},
  volume={123},
  number={4},
  pages={392},
  year={2016},
  publisher={American Psychological Association}
}

@article{collins2024building,
  title={Building machines that learn and think with people},
  author={Collins, Katherine M and Sucholutsky, Ilia and Bhatt, Umang and Chandra, Kartik and Wong, Lionel and Lee, Mina and Zhang, Cedegao E and Zhi-Xuan, Tan and Ho, Mark and Mansinghka, Vikash and others},
  journal={Nature human behaviour},
  volume={8},
  number={10},
  pages={1851--1863},
  year={2024},
  publisher={Nature Publishing Group UK London}
}

@article{schmidhuber1992learning,
  title={Learning factorial codes by predictability minimization},
  author={Schmidhuber, J{\"u}rgen},
  journal={Neural computation},
  volume={4},
  number={6},
  pages={863--879},
  year={1992},
  publisher={MIT Press}
}

@article{kingma2023understanding,
  title={Understanding diffusion objectives as the elbo with simple data augmentation},
  author={Kingma, Diederik and Gao, Ruiqi},
  journal={Advances in Neural Information Processing Systems},
  volume={36},
  pages={65484--65516},
  year={2023}
}

@article{hoogeboom2024simpler,
  title={Simpler diffusion (sid2): 1.5 fid on imagenet512 with pixel-space diffusion},
  author={Hoogeboom, Emiel and Mensink, Thomas and Heek, Jonathan and Lamerigts, Kay and Gao, Ruiqi and Salimans, Tim},
  journal={arXiv preprint arXiv:2410.19324},
  year={2024}
}

@article{jiang2023object,
  title={Object-centric slot diffusion},
  author={Jiang, Jindong and Deng, Fei and Singh, Gautam and Ahn, Sungjin},
  journal={arXiv preprint arXiv:2303.10834},
  year={2023}
}

@article{hertz2022prompt,
  title={Prompt-to-prompt image editing with cross attention control},
  author={Hertz, Amir and Mokady, Ron and Tenenbaum, Jay and Aberman, Kfir and Pritch, Yael and Cohen-Or, Daniel},
  journal={arXiv preprint arXiv:2208.01626},
  year={2022}
}

@article{okawa2023compositional,
  title={Compositional abilities emerge multiplicatively: Exploring diffusion models on a synthetic task},
  author={Okawa, Maya and Lubana, Ekdeep S and Dick, Robert and Tanaka, Hidenori},
  journal={Advances in Neural Information Processing Systems},
  volume={36},
  pages={50173--50195},
  year={2023}
}

@inproceedings{hang2023efficient,
  title={Efficient diffusion training via min-snr weighting strategy},
  author={Hang, Tiankai and Gu, Shuyang and Li, Chen and Bao, Jianmin and Chen, Dong and Hu, Han and Geng, Xin and Guo, Baining},
  booktitle={Proceedings of the IEEE/CVF international conference on computer vision},
  pages={7441--7451},
  year={2023}
}

@inproceedings{choi2022perception,
  title={Perception prioritized training of diffusion models},
  author={Choi, Jooyoung and Lee, Jungbeom and Shin, Chaehun and Kim, Sungwon and Kim, Hyunwoo and Yoon, Sungroh},
  booktitle={Proceedings of the IEEE/CVF conference on computer vision and pattern recognition},
  pages={11472--11481},
  year={2022}
}

@article{li2025back,
  title={Back to basics: Let denoising generative models denoise},
  author={Li, Tianhong and He, Kaiming},
  journal={arXiv preprint arXiv:2511.13720},
  year={2025}
}

@article{salimans2022progressive,
  title={Progressive distillation for fast sampling of diffusion models},
  author={Salimans, Tim and Ho, Jonathan},
  journal={arXiv preprint arXiv:2202.00512},
  year={2022}
}

@article{chen2023importance,
  title={On the importance of noise scheduling for diffusion models},
  author={Chen, Ting},
  journal={arXiv preprint arXiv:2301.10972},
  year={2023}
}

@inproceedings{leng2025repa,
  title={Repa-e: Unlocking vae for end-to-end tuning of latent diffusion transformers},
  author={Leng, Xingjian and Singh, Jaskirat and Hou, Yunzhong and Xing, Zhenchang and Xie, Saining and Zheng, Liang},
  booktitle={Proceedings of the IEEE/CVF International Conference on Computer Vision},
  pages={18262--18272},
  year={2025}
}

@article{yu2024representation,
  title={Representation alignment for generation: Training diffusion transformers is easier than you think},
  author={Yu, Sihyun and Kwak, Sangkyung and Jang, Huiwon and Jeong, Jongheon and Huang, Jonathan and Shin, Jinwoo and Xie, Saining},
  journal={arXiv preprint arXiv:2410.06940},
  year={2024}
}

@article{heek2026unified,
  title={Unified Latents (UL): How to train your latents},
  author={Heek, Jonathan and Hoogeboom, Emiel and Mensink, Thomas and Salimans, Tim},
  journal={arXiv preprint arXiv:2602.17270},
  year={2026}
}

@inproceedings{wang2025closer,
  title={A closer look at time steps is worthy of triple speed-up for diffusion model training},
  author={Wang, Kai and Shi, Mingjia and Zhou, Yukun and Li, Zekai and Yuan, Zhihang and Shang, Yuzhang and Peng, Xiaojiang and Zhang, Hanwang and You, Yang},
  booktitle={Proceedings of the Computer Vision and Pattern Recognition Conference},
  pages={12934--12944},
  year={2025}
}

@inproceedings{singh2022illiterate,
    title={Illiterate {DALL}-E Learns to Compose},
    author={Gautam Singh and Fei Deng and Sungjin Ahn},
    booktitle={International Conference on Learning Representations},
    year={2022},
    url={https://openreview.net/forum?id=h0OYV0We3oh}
}

@inproceedings{reed2015deep,
 author = {Reed, Scott E and Zhang, Yi and Zhang, Yuting and Lee, Honglak},
 booktitle = {Advances in Neural Information Processing Systems},
 editor = {C. Cortes and N. Lawrence and D. Lee and M. Sugiyama and R. Garnett},
 pages = {},
 publisher = {Curran Associates, Inc.},
 title = {Deep Visual Analogy-Making},
 url = {https://proceedings.neurips.cc/paper_files/paper/2015/file/e07413354875be01a996dc560274708e-Paper.pdf},
 volume = {28},
 year = {2015}
}

@inproceedings{gondal2019on,
 author = {Gondal, Muhammad Waleed and Wuthrich, Manuel and Miladinovic, Djordje and Locatello, Francesco and Breidt, Martin and Volchkov, Valentin and Akpo, Joel and Bachem, Olivier and Sch\"{o}lkopf, Bernhard and Bauer, Stefan},
 booktitle = {Advances in Neural Information Processing Systems},
 editor = {H. Wallach and H. Larochelle and A. Beygelzimer and F. d\textquotesingle Alch\'{e}-Buc and E. Fox and R. Garnett},
 pages = {},
 publisher = {Curran Associates, Inc.},
 title = {On the Transfer of Inductive Bias from Simulation to the Real World: a New Disentanglement Dataset},
 url = {https://proceedings.neurips.cc/paper/2019/file/d97d404b6119214e4a7018391195240a-Paper.pdf},
 volume = {32},
 year = {2019}
}

@article{tishby2000information,
  title={The information bottleneck method},
  author={Tishby, Naftali and Pereira, Fernando C and Bialek, William},
  journal={arXiv preprint physics/0004057},
  year={2000}
}

@article{shwartz2017opening,
  title={Opening the black box of deep neural networks via information},
  author={Shwartz-Ziv, Ravid and Tishby, Naftali},
  journal={arXiv preprint arXiv:1703.00810},
  year={2017}
}

@article{Everingham10,
  author    = {Mark Everingham and
               Luc Gool and
               Christopher K. I. Williams and
               John Winn and
               Andrew Zisserman},
  title     = {The Pascal Visual Object Classes (VOC) Challenge},
  journal   = {International Journal of Computer Vision},
  volume    = {88},
  number    = {2},
  year      = {2010},
  pages     = {303-338},
}

@inproceedings{karazija2021clevrtex,
  author =        {Laurynas Karazija and Iro Laina and
                   Christian Rupprecht},
  booktitle =     {Thirty-fifth Conference on Neural Information
                   Processing Systems Datasets and Benchmarks Track},
  title =         {{C}levr{T}ex: {A} {T}exture-{R}ich {B}enchmark for {U}nsupervised
                   {M}ulti-{O}bject {S}egmentation},
  year =          {2021},
}

@article{liu2025metaslot,
  title={MetaSlot: Break Through the Fixed Number of Slots in Object-Centric Learning},
  author={Liu, Hongjia and Zhao, Rongzhen and Chen, Haohan and Pajarinen, Joni},
  journal={arXiv preprint arXiv:2505.20772},
  year={2025}
}

@inproceedings{ren2022DisCo,
  title   = {Learning Disentangled Representation by Exploiting Pretrained Generative Models: A Contrastive Learning View},
  author  = {Ren, Xuanchi and Yang, Tao and Wang, Yuwang and Zeng, Wenjun},
  booktitle = {ICLR},
  year    = {2022}
}

@inproceedings{singh2025glass,
  title={Glass: Guided latent slot diffusion for object-centric learning},
  author={Singh, Krishnakant and Schaub-Meyer, Simone and Roth, Stefan},
  booktitle={Proceedings of the Computer Vision and Pattern Recognition Conference},
  pages={28673--28683},
  year={2025}
}

@inproceedings{seitzer2023bridging,
    title={Bridging the Gap to Real-World Object-Centric Learning},
    author={Maximilian Seitzer and Max Horn and Andrii Zadaianchuk and Dominik Zietlow and Tianjun Xiao and Carl-Johann Simon-Gabriel and Tong He and Zheng Zhang and Bernhard Sch{\"o}lkopf and Thomas Brox and Francesco Locatello},
    booktitle={The Eleventh International Conference on Learning Representations},
    year={2023},
    url={https://openreview.net/forum?id=b9tUk-f_aG}
}

@article{li2025understanding,
  title={Understanding representation dynamics of diffusion models via low-dimensional modeling},
  author={Li, Xiao and Zhang, Zekai and Li, Xiang and Chen, Siyi and Zhu, Zhihui and Wang, Peng and Qu, Qing},
  journal={arXiv preprint arXiv:2502.05743},
  year={2025}
}

@inproceedings{touvron2021going,
  title={Going deeper with image transformers},
  author={Touvron, Hugo and Cord, Matthieu and Sablayrolles, Alexandre and Synnaeve, Gabriel and J{\'e}gou, Herv{\'e}},
  booktitle={Proceedings of the IEEE/CVF international conference on computer vision},
  pages={32--42},
  year={2021}
}

@article{xu2024towards,
  title={Towards faster training of diffusion models: An inspiration of a consistency phenomenon},
  author={Xu, Tianshuo and Mi, Peng and Wang, Ruilin and Chen, Yingcong},
  journal={arXiv preprint arXiv:2404.07946},
  year={2024}
}

@inproceedings{yu2025repa,
  title={Representation Alignment for Generation: Training Diffusion Transformers Is Easier Than You Think},
  author={Sihyun Yu and Sangkyung Kwak and Huiwon Jang and Jongheon Jeong and Jonathan Huang and Jinwoo Shin and Saining Xie},
  year={2025},
  booktitle={International Conference on Learning Representations},
}

@inproceedings{wang2025can,
title={Can Diffusion Models Disentangle? A Theoretical Perspective},
author={Liming Wang and Muhammad Jehanzeb Mirza and Yishu Gong and Yuan Gong and Jiaqi Zhang and Brian H. Tracey and Katerina Placek and Marco Vilela and James R. Glass},
booktitle={The Thirty-ninth Annual Conference on Neural Information Processing Systems},
year={2025},
url={https://openreview.net/forum?id=u1HclHIsLQ}
}

@article{liu2024faster,
  title={Faster diffusion via temporal attention decomposition},
  author={Liu, Haozhe and Zhang, Wentian and Xie, Jinheng and Faccio, Francesco and Xu, Mengmeng and Xiang, Tao and Shou, Mike Zheng and Perez-Rua, Juan-Manuel and Schmidhuber, J{\"u}rgen},
  journal={arXiv preprint arXiv:2404.02747},
  year={2024}
}

@inproceedings{hoogeboom2023simple,
  title={simple diffusion: End-to-end diffusion for high resolution images},
  author={Hoogeboom, Emiel and Heek, Jonathan and Salimans, Tim},
  booktitle={International Conference on Machine Learning},
  pages={13213--13232},
  year={2023},
  organization={PMLR}
}

@article{ho2020denoising,
  title={Denoising diffusion probabilistic models},
  author={Ho, Jonathan and Jain, Ajay and Abbeel, Pieter},
  journal={Advances in neural information processing systems},
  volume={33},
  pages={6840--6851},
  year={2020}
}

@InProceedings{pmlr-v202-brady23a,
  title = 	 {Provably Learning Object-Centric Representations},
  author =       {Brady, Jack and Zimmermann, Roland S. and Sharma, Yash and Sch\"{o}lkopf, Bernhard and Von K\"{u}gelgen, Julius and Brendel, Wieland},
  booktitle = 	 {Proceedings of the 40th International Conference on Machine Learning},
  pages = 	 {3038--3062},
  year = 	 {2023},
  editor = 	 {Krause, Andreas and Brunskill, Emma and Cho, Kyunghyun and Engelhardt, Barbara and Sabato, Sivan and Scarlett, Jonathan},
  volume = 	 {202},
  series = 	 {Proceedings of Machine Learning Research},
  month = 	 {23--29 Jul},
  publisher =    {PMLR},
  url = 	 {https://proceedings.mlr.press/v202/brady23a.html}
}

@misc{chen2026attnres,
  title         = {Attention Residuals},
  author        = {Kimi Team  and Chen, Guangyu  and Zhang, Yu  and Su, Jianlin  and Xu, Weixin  and Pan, Siyuan  and Wang, Yaoyu  and Wang, Yucheng  and Chen, Guanduo  and Yin, Bohong  and Chen, Yutian  and Yan, Junjie  and Wei, Ming  and Zhang, Y.  and Meng, Fanqing  and Hong, Chao  and Xie, Xiaotong  and Liu, Shaowei  and Lu, Enzhe  and Tai, Yunpeng  and Chen, Yanru  and Men, Xin  and Guo, Haiqing  and Charles, Y.  and Lu, Haoyu  and Sui, Lin  and Zhu, Jinguo  and Zhou, Zaida  and He, Weiran  and Huang, Weixiao  and Xu, Xinran  and Wang, Yuzhi  and Lai, Guokun  and Du, Yulun  and Wu, Yuxin  and Yang, Zhilin  and Zhou, Xinyu},
  year          = {2026},
  archiveprefix = {arXiv},
  eprint        = {2603.15031},
  primaryclass  = {cs.CL}
}

@article{chambon2022roentgen,
  title={Roentgen: vision-language foundation model for chest x-ray generation},
  author={Chambon, Pierre and Bluethgen, Christian and Delbrouck, Jean-Benoit and Van der Sluijs, Rogier and Po{\l}acin, Ma{\l}gorzata and Chaves, Juan Manuel Zambrano and Abraham, Tanishq Mathew and Purohit, Shivanshu and Langlotz, Curtis P and Chaudhari, Akshay},
  journal={arXiv preprint arXiv:2211.12737},
  year={2022}
}

@inproceedings{hudson2024soda,
  title={Soda: Bottleneck diffusion models for representation learning},
  author={Hudson, Drew A and Zoran, Daniel and Malinowski, Mateusz and Lampinen, Andrew K and Jaegle, Andrew and McClelland, James L and Matthey, Loic and Hill, Felix and Lerchner, Alexander},
  booktitle={Proceedings of the IEEE/CVF Conference on Computer Vision and Pattern Recognition},
  pages={23115--23127},
  year={2024}
}

@inproceedings{
yue2024exploring,
title={Exploring Diffusion Time-steps for Unsupervised Representation Learning},
author={Zhongqi Yue and Jiankun Wang and Qianru Sun and Lei Ji and Eric I-Chao Chang and Hanwang Zhang},
booktitle={The Twelfth International Conference on Learning Representations},
year={2024},
url={https://openreview.net/forum?id=bWzxhtl1HP}
}

@misc{chefer2023attendandexcite,
      title={Attend-and-Excite: Attention-Based Semantic Guidance for Text-to-Image Diffusion Models}, 
      author={Hila Chefer and Yuval Alaluf and Yael Vinker and Lior Wolf and Daniel Cohen-Or},
      year={2023},
      eprint={2301.13826},
      archivePrefix={arXiv},
      primaryClass={cs.CV}
}

@article{khanna2023diffusionsat,
  title={Diffusionsat: A generative foundation model for satellite imagery},
  author={Khanna, Samar and Liu, Patrick and Zhou, Linqi and Meng, Chenlin and Rombach, Robin and Burke, Marshall and Lobell, David and Ermon, Stefano},
  journal={arXiv preprint arXiv:2312.03606},
  year={2023}
}

@misc{labs2025flux1kontextflowmatching,
      title={FLUX.1 Kontext: Flow Matching for In-Context Image Generation and Editing in Latent Space},
      author={Black Forest Labs and Stephen Batifol and Andreas Blattmann and Frederic Boesel and Saksham Consul and Cyril Diagne and Tim Dockhorn and Jack English and Zion English and Patrick Esser and Sumith Kulal and Kyle Lacey and Yam Levi and Cheng Li and Dominik Lorenz and Jonas Müller and Dustin Podell and Robin Rombach and Harry Saini and Axel Sauer and Luke Smith},
      year={2025},
      eprint={2506.15742},
      archivePrefix={arXiv},
      primaryClass={cs.GR},
      url={https://arxiv.org/abs/2506.15742},
}

@article{jung2024learning,
  title={Learning to compose: Improving object centric learning by injecting compositionality},
  author={Jung, Whie and Yoo, Jaehoon and Ahn, Sungjin and Hong, Seunghoon},
  journal={arXiv preprint arXiv:2405.00646},
  year={2024}
}

@inproceedings{esser2021taming,
  title={Taming transformers for high-resolution image synthesis},
  author={Esser, Patrick and Rombach, Robin and Ommer, Bjorn},
  booktitle={Proceedings of the IEEE/CVF conference on computer vision and pattern recognition},
  pages={12873--12883},
  year={2021}
}

@inproceedings{zhang2018unreasonable,
  title={The unreasonable effectiveness of deep features as a perceptual metric},
  author={Zhang, Richard and Isola, Phillip and Efros, Alexei A and Shechtman, Eli and Wang, Oliver},
  booktitle={Proceedings of the IEEE conference on computer vision and pattern recognition},
  pages={586--595},
  year={2018}
}

@inproceedings{nichol2021improved,
  title={Improved denoising diffusion probabilistic models},
  author={Nichol, Alexander Quinn and Dhariwal, Prafulla},
  booktitle={International conference on machine learning},
  pages={8162--8171},
  year={2021},
  organization={PMLR}
}

@article{Wu_Zheng_2024, title={Factorized Diffusion Autoencoder for Unsupervised Disentangled Representation Learning}, volume={38}, url={https://ojs.aaai.org/index.php/AAAI/article/view/28407}, DOI={10.1609/aaai.v38i6.28407}, number={6}, journal={Proceedings of the AAAI Conference on Artificial Intelligence}, author={Wu, Ancong and Zheng, Wei-Shi}, year={2024}, month={Mar.}, pages={5930-5939} }

@article{kim2024adaptive,
  title={Adaptive non-uniform timestep sampling for diffusion model training},
  author={Kim, Myunsoo and Ki, Donghyeon and Shim, Seong-Woong and Lee, Byung-Jun},
  journal={arXiv preprint arXiv:2411.09998},
  year={2024}
}

@inproceedings{ronneberger2015u,
  title={U-net: Convolutional networks for biomedical image segmentation},
  author={Ronneberger, Olaf and Fischer, Philipp and Brox, Thomas},
  booktitle={International Conference on Medical image computing and computer-assisted intervention},
  pages={234--241},
  year={2015},
  organization={Springer}
}

@inproceedings{
eastwood2018a,
title={A framework for the quantitative evaluation of disentangled representations},
author={Cian Eastwood and Christopher K. I. Williams},
booktitle={International Conference on Learning Representations},
year={2018},
url={https://openreview.net/forum?id=By-7dz-AZ},
}

@article{child2019generating,
  title={Generating long sequences with sparse transformers},
  author={Child, Rewon and Gray, Scott and Radford, Alec and Sutskever, Ilya},
  journal={arXiv preprint arXiv:1904.10509},
  year={2019}
}

@inproceedings{chen2025towards,
  title={Towards stabilized and efficient diffusion transformers through long-skip-connections with spectral constraints},
  author={Chen, Guanjie and Zhao, Xinyu and Zhou, Yucheng and Qu, Xiaoye and Chen, Tianlong and Cheng, Yu},
  booktitle={Proceedings of the IEEE/CVF International Conference on Computer Vision},
  pages={17708--17718},
  year={2025}
}

@inproceedings{bachlechner2021rezero,
  title={Rezero is all you need: Fast convergence at large depth},
  author={Bachlechner, Thomas and Majumder, Bodhisattwa Prasad and Mao, Henry and Cottrell, Gary and McAuley, Julian},
  booktitle={Uncertainty in artificial intelligence},
  pages={1352--1361},
  year={2021},
  organization={PMLR}
}

@article{zhang2019fixup,
  title={Fixup initialization: Residual learning without normalization},
  author={Zhang, Hongyi and Dauphin, Yann N and Ma, Tengyu},
  journal={arXiv preprint arXiv:1901.09321},
  year={2019}
}

@article{saharia2022image,
  title={Image super-resolution via iterative refinement},
  author={Saharia, Chitwan and Ho, Jonathan and Chan, William and Salimans, Tim and Fleet, David J and Norouzi, Mohammad},
  journal={IEEE transactions on pattern analysis and machine intelligence},
  volume={45},
  number={4},
  pages={4713--4726},
  year={2022},
  publisher={IEEE}
}

@article{oquab2023dinov2,
  title={Dinov2: Learning robust visual features without supervision},
  author={Oquab, Maxime and Darcet, Timoth{\'e}e and Moutakanni, Th{\'e}o and Vo, Huy and Szafraniec, Marc and Khalidov, Vasil and Fernandez, Pierre and Haziza, Daniel and Massa, Francisco and El-Nouby, Alaaeldin and others},
  journal={arXiv preprint arXiv:2304.07193},
  year={2023}
}

@article{engelcke2019genesis,
  title={Genesis: Generative scene inference and sampling with object-centric latent representations},
  author={Engelcke, Martin and Kosiorek, Adam R and Jones, Oiwi Parker and Posner, Ingmar},
  journal={arXiv preprint arXiv:1907.13052},
  year={2019}
}

@article{lipman2022flow,
  title={Flow matching for generative modeling},
  author={Lipman, Yaron and Chen, Ricky TQ and Ben-Hamu, Heli and Nickel, Maximilian and Le, Matt},
  journal={arXiv preprint arXiv:2210.02747},
  year={2022}
}

@inproceedings{perez2018film,
  title={Film: Visual reasoning with a general conditioning layer},
  author={Perez, Ethan and Strub, Florian and De Vries, Harm and Dumoulin, Vincent and Courville, Aaron},
  booktitle={Proceedings of the AAAI conference on artificial intelligence},
  volume={32},
  number={1},
  year={2018}
}

@article{hubert1985comparing,
  title={Comparing partitions},
  author={Hubert, Lawrence and Arabie, Phipps},
  journal={Journal of Classification},
  volume={2},
  number={1},
  pages={193--218},
  year={1985},
  publisher={Springer},
  doi={10.1007/BF01908075}
}

@article{Rand1971Objective,
  Author = {W.M. Rand},
  Journal = {Journal of the American Statistical Association},
  Number = {336},
  Pages = {846--850},
  Title = {Objective criteria for the evaluation of clustering methods},
  Volume = {66},
  Year = {1971}}

@inproceedings{greff2019multi,
  title={Multi-object representation learning with iterative variational inference},
  author={Greff, Klaus and Kaufman, Rapha{\"e}l Lopez and Kabra, Rishabh and Watters, Nick and Burgess, Christopher and Zoran, Daniel and Matthey, Loic and Botvinick, Matthew and Lerchner, Alexander},
  booktitle={International conference on machine learning},
  pages={2424--2433},
  year={2019},
  organization={PMLR}
}

@article{pedregosa2011scikit,
  title={Scikit-learn: Machine learning in Python},
  author={Pedregosa, Fabian and Varoquaux, Ga{\"e}l and Gramfort, Alexandre and Michel, Vincent and Thirion, Bertrand and Grisel, Olivier and Blondel, Mathieu and Prettenhofer, Peter and Weiss, Ron and Dubourg, Vincent and others},
  journal={the Journal of machine Learning research},
  volume={12},
  pages={2825--2830},
  year={2011},
  publisher={JMLR. org}
}

@inproceedings{zheng2026diffusion,
  title={Diffusion transformers with representation autoencoders},
  author={Zheng, Boyang and Ma, Nanye and Tong, Shengbang and Xie, Saining},
  booktitle={International Conference on Learning Representations},
  volume={2026},
  pages={35791--35820},
  year={2026}
}

@article{chi2026disentangled,
  title={Disentangled Representation Learning via Flow Matching},
  author={Chi, Jinjin and Liu, Taoping and Yin, Mengtao and Li, Ximing and Jing, Yongcheng and Tao, Dacheng},
  journal={arXiv preprint arXiv:2602.05214},
  year={2026}
}

@article{ma2024surprising,
  title={The surprising effectiveness of skip-tuning in diffusion sampling},
  author={Ma, Jiajun and Xue, Shuchen and Hu, Tianyang and Wang, Wenjia and Liu, Zhaoqiang and Li, Zhenguo and Ma, Zhi-Ming and Kawaguchi, Kenji},
  journal={arXiv preprint arXiv:2402.15170},
  year={2024}
}

@inproceedings{peebles2023scalable,
  title={Scalable diffusion models with transformers},
  author={Peebles, William and Xie, Saining},
  booktitle={Proceedings of the IEEE/CVF international conference on computer vision},
  pages={4195--4205},
  year={2023}
}

@inproceedings{ma2024sit,
  title={Sit: Exploring flow and diffusion-based generative models with scalable interpolant transformers},
  author={Ma, Nanye and Goldstein, Mark and Albergo, Michael S and Boffi, Nicholas M and Vanden-Eijnden, Eric and Xie, Saining},
  booktitle={European Conference on Computer Vision},
  pages={23--40},
  year={2024},
  organization={Springer}
}

\newpage

\appendix

\section{Background}
\label{app:definitions}

\subsection{Cosine Noise Schedule}

We use the cosine noise schedule in all of our experiments. It is defined as 
\begin{equation}
    \lambda = f_\lambda(t) = -2 \log\bigg(\tan\bigg(\frac{\pi t}{2}\bigg)\bigg),
    \qquad
    t = f_\lambda^{-1}(\lambda) = \frac{2}{\pi} \arctan(e^{-\lambda/2}),
\end{equation}
with corresponding density
\begin{equation}
    p(\lambda)
    =
    -\frac{d}{d\lambda} f^{-1}_\lambda(\lambda)
    =
    \frac{\mathrm{sech}(\lambda / 2)}{2\pi}.
\end{equation}

% Also describe truncation and other schedules used in this paper.
% \subsection{Equivalent forms for $\veps$-Prediction
% \subsection{Equivalent Forms of the Objective}

% \paragraph{From $\lambda$-space to time-space.}
% Starting from \Cref{eq:weighted_elbo_importance_sampling}, we write
% \begin{align}
%     \mathcal{L}(\x,\tokens)
%     &=
%     \frac{1}{2}
%     \mathbb{E}_{\veps}
%     \left[
%     \int_{\lambda_{\min}}^{\lambda_{\max}}
%     p(\lambda)
%     \frac{w(\lambda)}{p(\lambda)}
%     \ell(\lambda)
%     d\lambda
%     \right] \\
%     &=
%     \frac{1}{2}
%     \mathbb{E}_{\veps}
%     \left[
%     \int_{\lambda_{\min}}^{\lambda_{\max}}
%     w(\lambda)\ell(\lambda)d\lambda
%     \right],
% \end{align}
% where
% \begin{equation}
%     \ell(\lambda)=
%     \big\|\veps-\hat{\veps}_\theta(\x_\lambda;\lambda,\tokens)\big\|_2^2.
% \end{equation}
% Using $\lambda=f_\lambda(t)$ with $t\sim\mathcal{U}(0,1)$ and $f_\lambda$ decreasing, we have
% \begin{equation}
%     d\lambda = \frac{d\lambda}{dt}dt,
%     \qquad
%     \int_{\lambda_{\min}}^{\lambda_{\max}}(\cdot)\,d\lambda
%     =
%     \int_0^1(\cdot)\left(-\frac{d\lambda}{dt}\right)dt.
% \end{equation}
% Thus,
% \begin{equation}
%     \mathcal{L}(\x,\tokens)
%     =
%     \frac{1}{2}
%     \mathbb{E}_{\veps\sim\normal(0,\identity),\,t\sim\mathcal{U}(0,1)}
%     \left[
%     w(\lambda)
%     \left(-\frac{d\lambda}{dt}\right)
%     \ell(\lambda)
%     \right],
%     \qquad \lambda=f_\lambda(t),
% \end{equation}
% which is the derivative-weighted time-space form used in \Cref{eq:weighted_elbo_deriv}.

\subsection{Objective for $\x_0$-Prediction}
\label{app:xpred_weighting}

Our main objective is written in $\veps$-prediction form, where the loss is weighted by $w_{\veps}(\lambda)$.
For some experiments, however, we use $\x_0$-prediction.
Here, we derive the equivalent weighting required for $\x_0$-prediction to induce the same effective $\veps$-prediction objective.
For variance-preserving diffusion, the forward process is
\begin{equation}
    \x_\lambda = \alpha_\lambda \x + \sigma_\lambda \veps,
    \qquad
    \alpha_\lambda^2 = \mathrm{sigmoid}(\lambda),
    \qquad
    \sigma_\lambda^2 = \mathrm{sigmoid}(-\lambda),
\end{equation}
so that
\begin{equation}
    \frac{\alpha_\lambda^2}{\sigma_\lambda^2}
    =
    \exp(\lambda).
\end{equation}
Given an $\x_0$-prediction model $\hat{\x}_\theta(\x_\lambda;\lambda,\tokens)$, the corresponding noise prediction is
\begin{equation}
    \hat{\veps}_\theta(\x_\lambda;\lambda,\tokens)
    =
    \frac{\x_\lambda - \alpha_\lambda \hat{\x}_\theta(\x_\lambda;\lambda,\tokens)}
    {\sigma_\lambda}.
\end{equation}
Substituting $\x_\lambda = \alpha_\lambda \x + \sigma_\lambda \veps$ gives
\begin{equation}
    \veps - \hat{\veps}_\theta
    =
    \frac{\alpha_\lambda}{\sigma_\lambda}
    \left(
        \x - \hat{\x}_\theta
    \right).
\end{equation}
Therefore,
\begin{equation}
    \big\|\veps - \hat{\veps}_\theta\big\|_2^2
    =
    \frac{\alpha_\lambda^2}{\sigma_\lambda^2}
    \big\|\x - \hat{\x}_\theta\big\|_2^2
    =
    \exp(\lambda)
    \big\|\x - \hat{\x}_\theta\big\|_2^2.
    \label{eq:xpred_eps_equiv}
\end{equation}

It follows that an $\veps$-prediction objective with weighting $w_{\veps}(\lambda)$,
\begin{equation}
    w_{\veps}(\lambda)
    \big\|\veps - \hat{\veps}_\theta\big\|_2^2,
\end{equation}
is equivalent to an $\x_0$-prediction objective with weighting
\begin{equation}
    w_\x(\lambda)
    =
    \exp(\lambda) w_{\veps}(\lambda).
    \label{eq:xpred_weight_conversion}
\end{equation}
In our main experiments for attribute disentanglement, we use
\begin{equation}
    w_{\veps}(\lambda)
    =
    \mathrm{sigmoid}(-\lambda).
\end{equation}
Using \Cref{eq:xpred_weight_conversion}, the corresponding $\x_0$-prediction weighting is
\begin{equation}
    w_\x(\lambda)
    =
    \exp(\lambda) \; \mathrm{sigmoid}(-\lambda)
    =
    \mathrm{sigmoid}(\lambda).
\end{equation}
Thus, when training $\x_0$-prediction models for spatial disentanglement with ClevrTex, we use $w_\x(\lambda)=\mathrm{sigmoid}(\lambda)$ to preserve the same effective $\veps$-space weighting as $w_{\veps}(\lambda)=\mathrm{sigmoid}(-\lambda)$.

\section{Metrics}

\subsection{Attribute Disentanglement}
\label{app:attr_dis_metrics}
For attribute disentanglement metrics (DCI, FactorVAE and MIG), we use code from \citet{ren2022DisCo}.

\noindent\textbf{DCI Disentanglement \citep{eastwood2018a}.} 
We report the disentanglement component of the DCI framework \citep{eastwood2018a}.
This measures whether the predictive information in each latent dimension is concentrated on a single ground-truth factor.
We train a gradient-boosted tree classifier from Scikit-Learn \citep{pedregosa2011scikit} to predict each of the $M$ ground-truth factors from the learned representation vector $\mathbf{z} \in \mathbb{R}^D$ produced by $\Feat(\cdot)$, using $10{,}000$ training and $5{,}000$ test images.
We use the absolute values of the weights to form a feature importance matrix $\mathbf{R} \in \mathbb{R}^{D \times M}$, where the rows correspond to latents and the columns to ground truth factors.
For each latent dimension $i$, we form a distribution over true factors,
\begin{equation}
    p_{ij}
    =
    \frac{R_{ij}}{\sum_{j'=1}^{M} R_{ij'}}, \qquad \mathbf{p}_i = (p_{i1}, ..., p_{iM}).
\end{equation}
The per-latent disentanglement score is then defined as
\begin{equation}
    D_i
    =
    1 - H_M(\mathbf{p}_i),
    \qquad
    H_M(\mathbf{p}_i)
    =
    -\sum_{j=1}^{M} p_{ij}\log_M p_{ij}.
\end{equation}
If latent dimension $i$ is important for only one ground-truth factor, then $H_M(\mathbf{p}_i)=0$ and $D_i=1$.
If it is equally important for all $M$ factors, then $H_M(\mathbf{p}_i)=1$ and $D_i=0$.
The per-latent relative importance is defined by normalising each row w.r.t every cell in $\mathbf{R}$,
\begin{equation}
    \rho_i
    =
    \frac{\sum_{j=1}^{M} R_{ij}}
         {\sum_{i'=1}^{D}\sum_{j=1}^{M} R_{i'j}} .
\end{equation}
The final score is the importance-weighted average over latent dimensions,
\begin{equation}
    \mathrm{DCI\text{-}D}
    =
    \sum_{i=1}^{D} \rho_i D_i .
\end{equation}
High values indicate that individual latent dimensions are predictive of single ground-truth factors.
We report this value on the test samples.

\noindent\textbf{FactorVAE \citep{kim2018disentangling}.}
This tests whether a fixed generative factor can be identified by finding which latent dimension changes the least.
We first estimate the global variance of each latent coordinate in $\mathbf{z} \in \mathbb{R}^D$ using $10{,}000$ training images, and prune inactive dimensions whose standard deviation is below $0.05$.
Then, we repeatedly sample mini-batches of size $64$, with the value of one ground-truth factor $v_j$ fixed across the batch, and sampling all other factors randomly.
For each mini-batch, we compute the variance of each active latent coordinate and normalise it by its global variance.
The active latent dimension with the smallest normalised variance casts a vote for the fixed factor $v_j$.
Aggregating these votes over $10{,}000$ images gives a majority-vote classifier from latent dimensions to factors.
We report the accuracy of this classifier on $5{,}000$ test samples.
High values indicate that fixing a ground-truth factor reliably makes one latent dimension remain stable while the others vary.

\noindent\textbf{Mutual Information Gap (MIG) \citep{chen2018isolating}.}
This measures whether each ground-truth factor is captured by one clearly dominant latent dimension rather than being spread across several latents.
We sample $10{,}000$ images and compute their learned representations $\mathbf{z} \in \mathbb{R}^D$.
Each latent dimension $z_i$ is discretised into $20$ bins, after which we estimate the discrete mutual information between every discretised latent dimension $\widetilde{z}_i \in \{1, ..., 20\}$ and every ground-truth factor $v_j$.
This gives a mutual information matrix $\mathbf{M} \in \mathbb{R}^{D \times M}$, where $M_{ij} = I(\widetilde{z}_i, v_j)$.
For each factor $v_j$, we sort the mutual information values across latent dimensions and compute the gap between the largest and second-largest values, normalised by the entropy $H(v_j)$:
\begin{equation}
    \mathrm{MIG}
    =
    \frac{1}{M}
    \sum_{j=1}^{M}
    \frac{
        M_{(1)j} - M_{(2)j}
    }{
        H(v_j)
    },
\end{equation}
where $M_{(1)j}$ and $M_{(2)j}$ denote the largest and second-largest entries in column $j$ of $\mathbf{M}$.
High values indicate that each ground-truth factor has a single dominant latent dimension that contains substantially more information about it than any other latent dimension.

\subsection{Spatial Disentanglement}

For the segmentation metrics for spatial disentanglement (FG-ARI, mBO, mIoU), we use the code from \citet{wu2023slotdiffusion}.
All of these metrics are converted to percentages for our tables.

\noindent\textbf{Foreground Adjusted Rand Index (FG-ARI) \citep{rand1971objective,hubert1985comparing,greff2019multi}.}
This measures whether foreground pixels that belong to the same ground-truth object are assigned to the same predicted segment.
The adjusted Rand Index (ARI) measures clustering similarity between pixels in a ground truth and predicted mask, where a score of $1$ corresponds to perfect agreement, while a score near $0$ corresponds to chance-level agreement.
Following common practice in object-centric learning \citep{locatello2020object}, we exclude pixels labelled as background in the ground-truth mask before computing ARI.
Higher values indicate that foreground object instances are better recovered.
Since FG-ARI ignores ground-truth background pixels, it does not penalise errors that only affect background segmentation \citep{engelcke2019genesis,karazija2021clevrtex}.

\noindent\textbf{Mean Intersection-over-Union (mIoU).}
This measures the spatial overlap between predicted masks and ground-truth masks.
For a predicted mask $P$ and ground-truth mask $G$, the IoU is
\begin{equation}
    \mathrm{IoU}(P,G)
    =
    \frac{|P \cap G|}{|P \cup G|}.
\end{equation}
Since predicted slots are unordered, we first compute all pairwise IoUs between predicted masks and the relevant ground-truth masks, and then use Hungarian matching to find an assignment that maximises total IoU.
The matched IoUs are averaged to obtain mIoU.
We report both instance-wise and class-wise variants.
For $\mathrm{mIoU}_i$, the ground-truth masks correspond to object instances.
For $\mathrm{mIoU}_c$, the ground-truth masks correspond to semantic classes.
Thus, $\mathrm{mIoU}_i$ measures recovery of individual object instances, while $\mathrm{mIoU}_c$ measures semantic class-level overlap.
Higher values indicate better spatial agreement between predicted and ground-truth masks.

\noindent\textbf{Mean Best Overlap (mBO).}
Mean best overlap also measures mask overlap, but unlike mIoU it does not enforce a one-to-one matching between predicted and ground-truth masks.
For each ground-truth mask $G_k$, we compute its best overlap with any predicted mask,
\begin{equation}
    \mathrm{BO}(G_k)
    =
    \max_l
    \frac{|G_k \cap P_l|}{|G_k \cup P_l|},
\end{equation}
where $\{P_l\}$ are the predicted masks.
The mean best overlap is then
\begin{equation}
    \mathrm{mBO}
    =
    \frac{1}{K}
    \sum_{k=1}^{K}
    \mathrm{BO}(G_k).
\end{equation}
We report both instance-wise and class-wise variants.
For $\mathrm{mBO}_i$, the ground-truth masks $\{G_k\}_{k=1}^{K}$ correspond to object instances.
For $\mathrm{mBO}_c$, they correspond to semantic classes.
Thus, $\mathrm{mBO}_i$ measures whether each object instance is covered by at least one predicted mask, while $\mathrm{mBO}_c$ measures whether each semantic class region is covered by at least one predicted mask.
Higher values indicate better best-case overlap between ground-truth masks and predicted masks.

\newpage

\section{Method}

\subsection{Amortised Semantic Encoder with FiLM Modulation}
\label{app:film_encoder}
To improve the parameter efficiency of the semantic encoder (\Cref{sec:semantic_encoder}), we design $\IB(\cdot)$ attribute disentanglement using a shared backbone with FiLM modulation \citep{perez2018film} in place of the split-MLP \citep{yang2024diffusion},
\begin{equation}
\textbf{Split-MLP:} \quad \s_i = \mathrm{IB}_{\xi_i}(z_i), 
\qquad
\textbf{Ours:} \quad \s_i = \mathrm{IB}_{\xi}\big(z_i; \{ \gamma_i^{(l)}, \beta_i^{(l)} \}_{l=1}^{L}\big),
\end{equation}
where the split-MLP uses separate parameters $\xi_i$ for each latent scalar $z_i$, while our approach shares weights $\xi$ across all latents. The FiLM parameters $\{\gamma_i^{(l)}, \beta_i^{(l)}\}_{l=1}^{L}$ provide latent-specific feature-wise modulation across $L$ layers of the shared network via scaling and shifting of intermediate activations. This design preserves factor-wise independence (each token $\s_i$ depends only on the scalar $z_i$) while reducing parameters by approximately $\mathcal{O}(N)$.
In this work, we set $L = 2$.
We found this to provide similar disentanglement as the split-MLP, seen in \Cref{tab:ablations_shapes3d}, when paired with a deeper feature extractor, which we refer to as Residual ConvNet in \Cref{tab:dataset_settings}.

\begin{center}
    \includegraphics[width=0.26\linewidth]{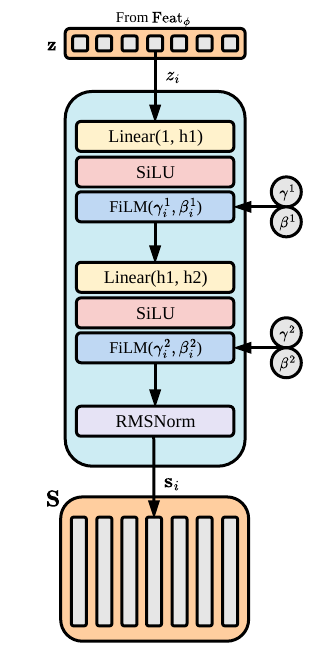}
\captionof{figure}{Architecture for Amortised Semantic Encoder}
\end{center}

% \subsection{Gate Design}
% More details to follow.

\subsection{Deriving the $\logsnr$ Curriculum Loss}
\label{app:curriculum_deriv}

Our curriculum restricts training at step $m$ to an active $\logsnr$ band
$[\lambda_L(m), \lambda_U(m)]$.
This defines the truncated density
\begin{equation}
    q(\lambda;m)
    =
    \frac{p(\lambda)}{Z(m)}
    \mathbf{1}\{\lambda \in [\lambda_L(m), \lambda_U(m)]\},
    \qquad
    Z(m)
    =
    \int_{\lambda_L(m)}^{\lambda_U(m)} p(\lambda)\,d\lambda .
\end{equation}
Sampling from $q(\lambda;m)$ therefore requires the importance correction
\begin{equation}
    \frac{w(\lambda)}{q(\lambda;m)}
    =
    Z(m)\frac{w(\lambda)}{p(\lambda)} .
\end{equation}
Thus, the curriculum objective in $\lambda$-space is
\begin{equation}
\mathcal{L}(\x,\tokens;m)
=
\frac{1}{2}
\E_{\veps\sim\normal(0,\identity),\,\lambda\sim q(\lambda;m)}
\left[
Z(m)\frac{w(\lambda)}{p(\lambda)}
\big\|
\veps-\hat{\veps}_\theta(\x_\lambda;\lambda,\tokens)
\big\|_2^2
\right].
\label{eq:curriculum_importance_appendix}
\end{equation}

Equivalently, when implementing the objective by sampling time, we sample
$t$ uniformly over the corresponding active time interval
$[t_U(m), t_L(m)]$, where
$t_U(m)=f_\lambda^{-1}(\lambda_U(m))$ and
$t_L(m)=f_\lambda^{-1}(\lambda_L(m))$.
For decreasing noise schedules, this interval has mass
\begin{equation}
    Z(m) = t_L(m)-t_U(m).
\end{equation}
Using $\lambda=f_\lambda(t)$, the derivative-form curriculum objective is then
\begin{equation}
\mathcal{L}(\x,\tokens;m)
=
\frac{1}{2}
\E_{\veps\sim\normal(0,\identity),\,t\sim\mathcal{U}(t_U(m),t_L(m))}
\left[
Z(m)\,
w(\lambda)
\left(-\frac{d\lambda}{dt}\right)
\big\|
\veps-\hat{\veps}_\theta(\x_\lambda;\lambda,\tokens)
\big\|_2^2
\right].
\label{eq:curriculum_deriv_appendix}
\end{equation}

\newpage

\section{Experimental Setup}
\label{app:exp_setup}

\subsection{Analysis}
\label{app:analysis_setup}
For the trajectory analysis in \Cref{sec:analysis}, we modify the lightweight \texttt{imagev3} U-Net implementation from \texttt{k-diffusion}\footnote{\url{https://github.com/crowsonkb/k-diffusion}} by adding EncDiff SpatialTransformer blocks for conditioning on the output of $\mathrm{Enc}_\phi(\cdot)$, together with a middle bottleneck layer.
This U-Net uses fixed convolutions for up- and down-sampling.
We use the \texttt{Encoder4} encoder for $\Feat(\cdot)$ and the split-MLP for $\IB(\cdot)$ from EncDiff \citep{yang2024diffusion} to construct $\mathrm{Enc}_\phi(\cdot)$.
Models are trained until they have seen 3M images, which is sufficient for most runs to reach stable reconstruction and disentanglement scores.
For this analysis, we use the EncDiff VQ-VAE checkpoint trained for 30K steps.
We use 5 model initialisation seeds and 5 $\lambda$-sampling seeds, giving 25 runs in total.
This lets us test whether early disentanglement correlates with model initialisation or noise-level sampling; we did not find strong evidence for either.
As discussed in \Cref{app:noise_level_ablations}, this checkpoint makes regime separation more visible, whereas longer VQ-VAE training improves reconstruction and reduces the separation between regimes.

\subsection{Main Experiments}
\label{app:main_exp_setup}
Architecture and training details for all datasets are summarised in \Cref{tab:dataset_settings}.

For the main attribute disentanglement experiments, we follow EncDiff \citep{yang2024diffusion} unless otherwise specified.
We use the same datasets, evaluation protocol, and VQ-VAE architecture (now trained to 150K steps), while replacing the split-MLP encoder with our Scalar-to-Token FiLM encoder and the standard U-Net with our gated residual U-Net.
All attribute models use $\veps$-prediction with $\mathrm{sigmoid}(-\lambda)$ weighting and the $\logsnr$ curriculum described in \Cref{sec:lambda_curriculum}.
We train until the model has seen 15M images, which is sufficient for convergence.
This takes approximately 16 hours on an NVIDIA A40 or 10 hours on an NVIDIA RTX 4090.

For object-centric learning, we build on SlotDiffusion \citep{wu2023slotdiffusion}, instantiate $\IB(\cdot)$ with Slot Attention, and follow its evaluation protocol on ClevrTex and PascalVOC.
ClevrTex uses $\x_0$-prediction with the equivalent weighting in \Cref{app:xpred_weighting}, while PascalVOC uses $\veps$-prediction.
We use the VQ-VAE checkpoints provided by the SlotDiffusion codebase.
For fair comparison to SlotDiffusion, we train ClevrTex for 400 epochs and PascalVOC for 500 epochs.
Trajectory plots are computed on the validation set for ClevrTex.
For PascalVOC, the val and test set are the same in the SlotDiffusion evaluation protocol.
On H100 GPUs, both object-centric models can be trained within one day.

We obtain the full $\logsnr$ band by mapping the noise schedules in the matched baselines, indexed by time, into $\logsnr$-space, with minor adjustments for our formulation. 

\begin{table}[!t]
\centering
\caption{Variations in model architectures and training settings across datasets.}
\label{tab:dataset_settings}
\footnotesize
\resizebox{\linewidth}{!}{
\begin{tabular}{lccccc}
\toprule
\textbf{Dataset} & \textbf{Shapes3D} & \textbf{Cars3D} & \textbf{MPI3D} & \textbf{ClevrTex} & \textbf{PascalVOC} \\
\midrule
Feature Extractor ($\Feat(\cdot)$) & Residual ConvNet & Residual ConvNet & Residual ConvNet & ResNet34 & DINO ViT-S/8 \\
Disentanglement Inductive Bias ($\IB(\cdot)$) & Amortised FiLM & Amortised FiLM & Amortised FiLM & Slot Attention & Slot Attention \\
Number of Tokens / Slots ($N$) & 10/20 & 10/20 & 10/20 & 11 & 6 \\
Token / Slot Dimension & 16 & 16 & 16 & 192 & 192 \\
Slot Attention Iterations & -- & -- & -- & 3 & 3 \\
\midrule
Prediction Target & $\veps$ & $\veps$ & $\veps$ & $\x_0$ & $\veps$ \\
Loss Weighting & $\mathrm{Sigmoid}(-\lambda)$ & $\mathrm{Sigmoid}(-\lambda)$ & $\mathrm{Sigmoid}(-\lambda)$ & $\mathrm{Sigmoid}(\lambda)$ & $\mathrm{Sigmoid}(-\lambda)$ \\
Initial Curriculum Band ($[\lambda_L(0), \lambda_U(0)]$) & $[0, 10]$ & $[0, 10]$ & $[0, 10]$ & $[-2,2]$ & $[-2,2]$ \\
Full $\logsnr$ Band ($[\lambda_{\min}, \lambda_{\max}]$) & $[-5,12]$ & $[-5,12]$ & $[-5,12]$ & $[-8.5, 6.5]$ & $[-15, 15]$ \\
Widening Steps ($m_{\max}$) & $32$K & $32$K & $32$K & $200$K & $200$K \\
Learning Rate & $5{\times}10^{-4}$ & $5{\times}10^{-4}$ & $5{\times}10^{-4}$ & $10^{-4}$ & $10^{-4}$ \\
\midrule
LR Schedule & Constant & Constant & Constant & Cosine & Cosine \\
EMA & \multicolumn{3}{c}{U-Net + encoder EMA; inverse warmup, max $0.9999$} & -- & -- \\
Optimiser & AdamW & AdamW & AdamW & Adam & Adam \\
Batch Size & 128 & 128 & 128 & 64 & 64 \\
\midrule
Base U-Net Architecture & \multicolumn{3}{c}{EncDiff \citep{yang2024diffusion}} & \multicolumn{2}{c}{SlotDiffusion \citep{wu2023slotdiffusion}} \\
\midrule
Base VQ-VAE Architecture & \multicolumn{3}{c}{\citet{esser2021taming} with LPIPS Discriminator} & \multicolumn{2}{c}{Ckpts from SlotDiffusion Repo \citep{wu2023slotdiffusion}} \\
Latent Resolution $\x_{0}$ & $16{\times}16$ & $16{\times}16$ & $16{\times}16$ & $32{\times}32$ & $56{\times}56$ \\
\bottomrule
\end{tabular}
}
\label{tab:training_settings}
\end{table}

\newpage

\subsection{OCL Curricula}
% This is the curriculum we use for the PascalVOC dataset (\Cref{sec:ocl_res}). It is more neutral than the curriculum used for ClevrTex (\Cref{fig:attr_curriculum}), which steers towards the disentanglement regime.

For spatial disentanglement on ClevrTex and PascalVOC, we use the same initial log SNR band \([-2,2]\) and widening duration \(m_{\max}=200\text{K}\).
The final bands follow the corresponding baseline schedules: \([-8.5,6.5]\) for ClevrTex and \([-15,15]\) for PascalVOC (\Cref{tab:training_settings}).
\Cref{fig:shapes3d_regimes_sech_figure9} illustrates the PascalVOC curriculum.

\begin{center}
    \includegraphics[width=0.5\textwidth]{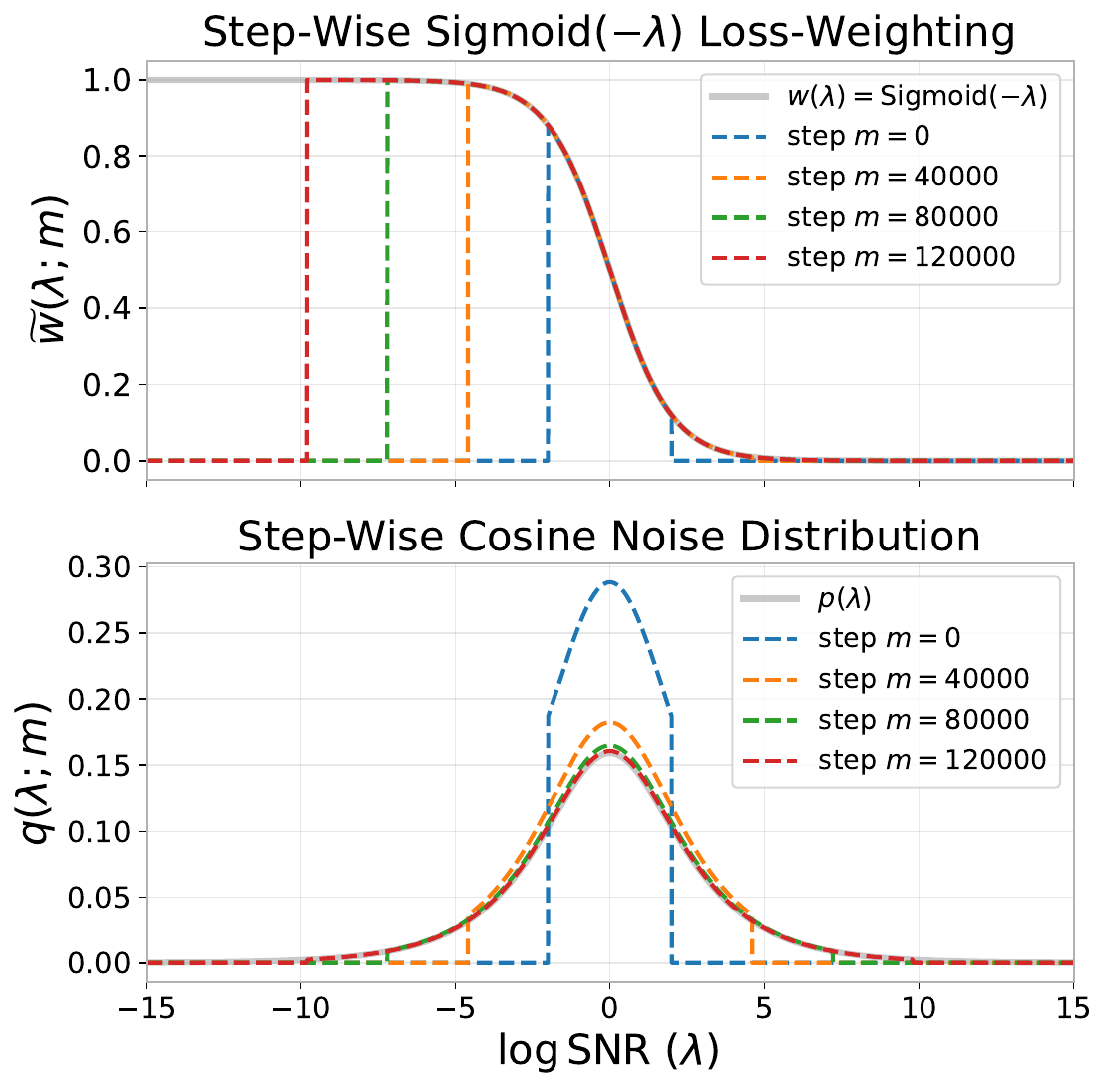}
\captionof{figure}{$\logsnr$ curriculum for object-centric learning on the Pascal VOC dataset.}
\label{fig:shapes3d_regimes_sech_figure9}
\end{center}

\newpage

\section{Additional Results}
\label{app:analysis}

\subsection{Training Trajectories}
\label{app:traj_details}
We repeat the trajectory analysis from \Cref{fig:shapes3d_regimes} under alternative loss-weighting functions.
In \Cref{fig:shapes3d_regimes_sech}, we use the non-monotonic weighting $w(\lambda)=\mathrm{sech}(\lambda)$, which places most weight near intermediate $\logsnr$ values, akin to the EDM with shift $-3$ we used in \Cref{sec:analysis}.
The same qualitative behaviour appears; runs separate into trajectories that prioritise reconstruction early and trajectories that maintain stronger disentanglement throughout training.
This suggests that the regime structure is not specific to a single loss-weighting choice, but reflects a broader optimisation property of diffusion autoencoders which becomes clear in this low-capacity setting.

\begin{center}
    \includegraphics[width=\textwidth]{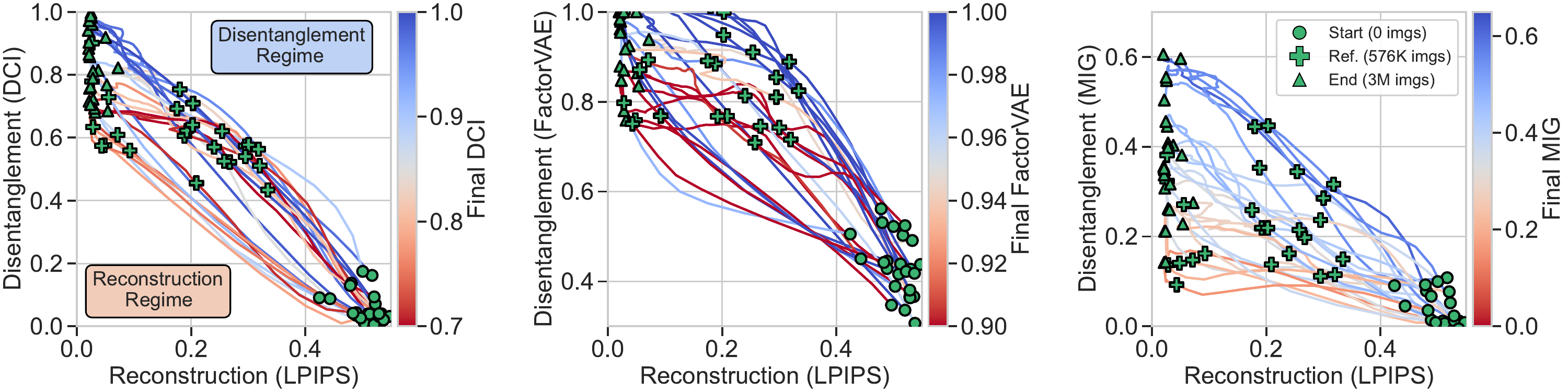}
\captionof{figure}{
\textbf{Training trajectories organise around two distinct regimes.}
This follows the same setup as \Cref{fig:ablations_traj}, now using a different non-monotonic loss-weighting function, $w(\lambda) = \mathrm{sech}(\lambda).$
}
\label{fig:shapes3d_regimes_sech}
\end{center}

\subsection{Analysing Regimes in Training Trajectories}
\label{app:clustering}
Here we present a post-hoc analysis of the same 25 trajectories shown in \Cref{fig:shapes3d_regimes,fig:shapes3d_regimes_sech}.
We assess the two-regime structure by constructing a summary statistic of each training trajectory.
For each run, at a step $s$, we define the normalised reconstruction progress 
\begin{equation}
R(s)
=
\frac{\mathrm{LPIPS}(s_1)-\mathrm{LPIPS}(s)}
     {\mathrm{LPIPS}(s_1)},
\end{equation}
and normalised disentanglement progress for the metrics $m \in \{\mathrm{DCI}, \mathrm{FactorVAE}, \mathrm{MIG}\}$ as
\begin{equation}
D_m(s)
=
\frac{m(s)-m(s_1)}
     {1-m(s_1)},
\end{equation}
where $s$ is the number of training images seen and $s_1$ is the first checkpoint included in this analysis.
The two quantities measure the relative progress of the reconstruction and disentanglement, and are therefore both equal zero at $s_1$.
We set $s_1 = 64K$, which is the first evaluated checkpoint after $s = 0$.
In \Cref{fig:shapes3d_regimes,fig:shapes3d_regimes_sech}, the starting point at $s = 0$ is marked by a circle and the reference point at $s_{\mathrm{ref}}=576\mathrm{K}$ is marked by a plus.
The relative progress between $s_1$ and $s_{\mathrm{ref}}$ is
\begin{equation}
J_m
=
\frac{1}{s_{\mathrm{ref}}-s_1}
\int_{s_1}^{s_{\mathrm{ref}}}
\left[D_m(s)-R(s)\right]\,\mathrm{d}s
=
\frac{1}{512\mathrm{K}}
\int_{64\mathrm{K}}^{576\mathrm{K}}
\left[D_m(s)-R(s)\right]\,\mathrm{d}s.
\end{equation}
Positive values indicate that disentanglement progresses more quickly than reconstruction over this interval, while negative values indicate the converse.
The integral is approximated using the trapezium rule over the nine measurements from $64\mathrm{K}$ to $576\mathrm{K}$.
We compute $J_m$ for each of the 25 runs denoted $J_m^r$, $r \in \{1,\ldots,25\}$.
For each metric $m$, we calculate the mean and standard deviation across runs:
\begin{equation}
\mu_m
=
\frac{1}{25}\sum_{r=1}^{25}J_m^r,
\qquad
\sigma_m
=
\sqrt{
\frac{1}{25}\sum_{r=1}^{25}
\left(J_m^r-\mu_m\right)^2
}.
\end{equation}
The $Z$-score $Z_m^r = (J_m^r-\mu_m) / \sigma_m$ is averaged over the three metrics to obtain a per-run statistic
\begin{equation}
J_c^r
=
\frac{1}{3}
\left(
Z_{\mathrm{DCI}}^r
+
Z_{\mathrm{FactorVAE}}^r
+
Z_{\mathrm{MIG}}^r
\right),
\end{equation}
where larger values of $J_c^r$ indicate greater early disentanglement progress relative to reconstruction.
All normalisation and standardisation operations are performed separately for each loss weighting.

For each loss weighting, we fit one-dimensional Gaussian mixture models (GMMs) to the $J_c^r$ values using \texttt{scikit-learn}.
We compare $K \in \{1,2,3,4,5\}$ mixture components and compare models using the Bayesian information criterion (BIC), for which lower values are preferred.
Relative to the two component model, we also define
\begin{equation}
\label{eq:bic_diff_metric}
\Delta\mathrm{BIC}_{K-2}
=
\mathrm{BIC}_K-\mathrm{BIC}_2,
\end{equation}
Positive values of $\Delta\mathrm{BIC}_{k-2}$ indicate that the two-component model is preferred over the corresponding $k$-component alternative.
For the selected $K=2$ model, we identify the component with the larger mean $J_c^r$ as the disentanglement regime and the other as the reconstruction regime.

% We select the number of clusters to be fit as $K \in \{1,2,3,4,5\}$
% Positive values for $\Delta \text{BIC}_{k - 2} = \text{BIC}_{k} - \text{BIC}_2$, with $k \in \{1,3,4,5\}$, measure whether clustering favours two regimes over $k$ regimes.
% We record how many of the 25 runs fall into each regime.

\paragraph{Evidence for Two Regimes.}
The two-regime structure visible in  \Cref{fig:shapes3d_regimes,fig:shapes3d_regimes_sech} is supported quantitatively.
For each loss weighting, we fit GMMs with $K\in \{1,2,3\}$ using \texttt{n\_init=100} and select the model with the lowest BIC.
For both EDM-3 and sech, BIC favours $K=2$ over $K=1$ and $K=3$.
In \Cref{tab:gmm_regime_selection}, both resulting regimes contain a substantial fraction of the 25 trajectories.

\begin{table}[!htbp]
    \centering
    \small
    \caption{
    BIC differences and regime assignments for the two loss weightings.
    Positive $\Delta\mathrm{BIC}_{k-2}$ values indicate that the two-component model is preferred over the $K \in \{1, 3\}$ component alternative.
    }
    \begin{tabular}{@{}lccc@{}}
        \toprule
        Weighting
        & $\Delta\mathrm{BIC}_{1-2}$
        & $\Delta\mathrm{BIC}_{3-2}$
        & Recon. / Dis.\ Regime \\
        \midrule
        EDM-3 & $5.548$ & $5.452$ & $13/12$ \\
        Sech  & $3.086$ & $2.992$ & $9/16$ \\
        \bottomrule
    \end{tabular}
    \label{tab:gmm_regime_selection}
\end{table}

\paragraph{Stability of Model Selection.}
To assess sensitivity to GMM initialisation, we repeat model selection 100 times.
In each trial, we fit one GMM for every $K \in \{1,2,3,4,5\}$ using \texttt{n\_init=1}, with \texttt{random\_state} set to each value from 0 to 99.
We then select the $K$ with the lowest BIC within each trial.
For both loss weightings, we see that $K = 2$ has the lowest $\text{BIC}$ in \Cref{tab:bic_with_different_k}.
Here we found that all 100 repeats select $K = 2$ via the metric in \Cref{eq:bic_diff_metric}.
% Equivalently, $\Delta\mathrm{BIC}_{k-2} > 0$ for every $k \in \{1,3,4,5\}$ in every trial.

\begin{table}[!htbp]
    \centering
    \caption{\textbf{Clustering favours two regimes.} We report the mean ± std BIC over 100 repeats of the $J_c^r$ clustering procedure over $K \in \{1,2,3,4,5\}$, and find that lowest values for $K = 2$.}
    \small
    \begin{tabular}{@{}lcccccc@{}}
        \toprule
        \textbf{Weighting}
        & \textbf{$K=1$}
        & \textbf{$K=2$}
        & \textbf{$K=3$}
        & \textbf{$K=4$}
        & \textbf{$K=5$} \\
        \midrule
        EDM-3
        & $76.05 \pm 0.00$
        & $70.50 \pm 0.00$
        & $76.61 \pm 1.63$
        & $83.25 \pm 1.49$
        & $87.21 \pm 2.95$ \\
        Sech
        & $75.51 \pm 0.00$
        & $72.42 \pm 0.00$
        & $79.31 \pm 1.66$
        & $85.59 \pm 2.19$
        & $88.01 \pm 1.70$ \\
        \bottomrule
    \end{tabular}
    \label{tab:bic_with_different_k}
\end{table}

\paragraph{Uncertainty in Individual Assignments.}
We next assess the ambiguity of the assignments made by the selected $K=2$ model.
For each loss weighting, we use the GMM fitted with \texttt{n\_init=100} to the 25 values of $J_c^r$.
For each trajectory, the GMM returns a posterior probability for each component, and we assign a run to the component with the larger probability.
\Cref{tab:gmm_assignment_probabilities} reports summary statistics of the maximum posterior probability across runs.
All sech weighted trajectories and 24 of the 25 EDM-3-weighted trajectories have maximum posterior probabilities above $0.90$, indicating that the fitted models make largely unambiguous assignments.

\begin{table}[!htbp]
    \centering
    \small
    \caption{
    \textbf{Regime assignments for each run have high confidence.}
    Mean, median, and minimum for the maximum posterior probability across the 25 trajectories.
Values closer to one indicate less ambiguous assignments.
    }
    \begin{tabular}{@{}lcccc@{}}
        \toprule
        \textbf{Weighting}
        & \textbf{Mean}
        & \textbf{Median}
        & \textbf{Minimum}
        & \textbf{Runs below $0.90$} \\
        \midrule
        EDM-3 & $0.995$ & $1.000$ & $0.898$ & $1/25$ \\
        Sech  & $0.992$ & $0.999$ & $0.951$ & $0/25$ \\
        \bottomrule
    \end{tabular}
    \label{tab:gmm_assignment_probabilities}
\end{table}

\paragraph{Stability of the Partitions.}
Finally, we test whether all 100 separately fitted GMMs with \texttt{n\_init = 1} and $K=2$ recover the same partition of the runs.
Since GMM component labels may be permuted across fits, we measure agreement using the adjusted Rand index (ARI), which is invariant to this ordering.
We compute the ARI between the assignments produced by every pair of initialisations, where an ARI of one indicates identical partitions up to label permutation.
In \Cref{tab:gmm_partition_stability}, every pair has an ARI of one, showing that all 100 fits recover the same partition of the 25 runs.

\begin{table}[!htbp]
    \centering
    \small
    \caption{
    \textbf{The two-regime partition is stable across initialisations.}
    Mean and minimum pairwise ARI across the 100 independently initialised $K=2$ GMMs.
    An ARI of one denotes identical partitions up to a label permutation.
    }
    \begin{tabular}{@{}lcc@{}}
        \toprule
        \textbf{Weighting}
        & \textbf{Mean Pairwise ARI}
        & \textbf{Minimum Pairwise ARI} \\
        \midrule
        EDM-3 & $1.000$ & $1.000$ \\
        Sech  & $1.000$ & $1.000$ \\
        \bottomrule
    \end{tabular}
    \label{tab:gmm_partition_stability}
\end{table}

\subsection{Analysing Early Regime Selection} 
% We perform further analysis to test our claim that early separation of trajectories into the reconstruction and disentanglement regimes is predictive of the final disentanglement outcomes.
For each loss weighting and disentanglement metric, we compute Pearson's $r$ across the 25 runs between the disentanglement at the reference point and its final value.
This tests whether a run's early disentanglement is informative of its final disentanglement.
As shown in \Cref{tab:early_final_correlations}, all correlations are positive and range from $0.520$ to $0.827$.
These scores suggest that early disentanglement is positively correlated with the final outcome under both weightings and all three metrics in \Cref{fig:shapes3d_regimes,fig:shapes3d_regimes_sech}.

\begin{table}[!htbp]
    \centering
    \small
    \caption{
    Pearson's $r$ correlation between disentanglement at the reference point of $576\mathrm{K}$ images seen, marked with a + in \Cref{fig:shapes3d_regimes,fig:shapes3d_regimes_sech}, and the final disentanglement for each metric.
    }
    \begin{tabular}{@{}lccc@{}}
        \toprule
        \textbf{Weighting}
        & \textbf{DCI}
        & \textbf{FactorVAE}
        & \textbf{MIG} \\
        \midrule
        EDM-3 & $0.662$ & $0.520$ & $0.827$ \\
        Sech  & $0.622$ & $0.781$ & $0.767$ \\
        \bottomrule
    \end{tabular}
    \label{tab:early_final_correlations}
\end{table}

\subsection{Sensitivity to VQ-VAE Training}
The analysis in \Cref{fig:shapes3d_regimes,fig:shapes3d_regimes_sech} uses a VQ-VAE trained for $30\mathrm{K}$ steps, for which the separation between regimes is particularly clear.
This setting motivates our hypotheses \textbf{H1} and \textbf{H2}, which we test through further experiments and use to design our method for steering training trajectories.
To assess how this pattern changes with VQ-VAE, we repeat the GMM and correlation analyses with the EDM-3 loss weighting and using the $150\mathrm{K}$ checkpoint employed in our main attribute disentanglement experiments (\Cref{sec:attr_dis_res}).

With this VQ-VAE, BIC now favours $K = 1$, aligned with a more continuous spectrum of trajectories rather than two clearly separated regimes.
However, final disentanglement continues to vary substantially across runs, and early disentanglement remains positively correlated with its final value (\Cref{tab:vqvae_maturity_correlations}).
The correlations range from $0.485$ to $0.776$, showing that early disentanglement remains informative of final disentanglement across all three metrics.
These results indicate that VQ-VAE training affects how distinctly the trajectory patterns separate, while the broader relationship between early and final representation quality persists.
Since our main attribute experiments use the $150\mathrm{K}$ checkpoint and still exhibit trajectory steering, \textsc{SteeringDRL} does not require the clean regime separation in its base model to be effective.
This supports \textbf{H1} and \textbf{H2} as design hypotheses that remain useful beyond the controlled, motivating diagnostic setting.

\begin{table}[!htbp]
    \centering
    \small
    \caption{
    \textbf{Early disentanglement remains informative with a mature VQ-VAE.}
    Pearson correlations between disentanglement at $576\mathrm{K}$ images seen and its final value across 25 runs using the $150\mathrm{K}$ VQ-VAE checkpoint.
    We also report the distribution of final scores.
    }
    \begin{tabular}{@{}lccc@{}}
        \toprule
        \textbf{Metric}
        & \textbf{Pearson's $r$}
        & \textbf{Final median [IQR]}
        & \textbf{Final range} \\
        \midrule
        DCI
        & $0.560$
        & $0.823\ [0.789,\,0.891]$
        & $0.634$--$0.951$ \\
        FactorVAE
        & $0.485$
        & $0.929\ [0.885,\,0.992]$
        & $0.680$--$1.000$ \\
        MIG
        & $0.776$
        & $0.363\ [0.274,\,0.396]$
        & $0.133$--$0.569$ \\
        \bottomrule
    \end{tabular}
    \label{tab:vqvae_maturity_correlations}
\end{table}

\newpage 

\subsection{Cross-Attention Maps in each Regime}
\label{app:cross_attn_details}
We visualise cross-attention maps from the low-capacity U-Net used in \Cref{sec:analysis}.
We use the second downsampling layer in the U-Net encoder, which operates at spatial resolution $8^2$, and visualise maps at the same layer and logSNR level for both models.
For readability, attention maps are upsampled and smoothed.
\Cref{fig:shapes3d_regimes_cross_attn} compares models from the reconstruction and disentanglement regimes.
Although both models achieve strong reconstructions, their internal attention patterns differ substantially.
In the disentanglement regime, several attention maps are spatially localised and align with all ground-truth factors.
In contrast, attention maps in the reconstruction regime are more diffuse and less consistently aligned with individual factors.
This supports the interpretation that the two regimes correspond not only to different metric values, but also to different internal representations.

\begin{center}
    \includegraphics[width=0.49\textwidth]{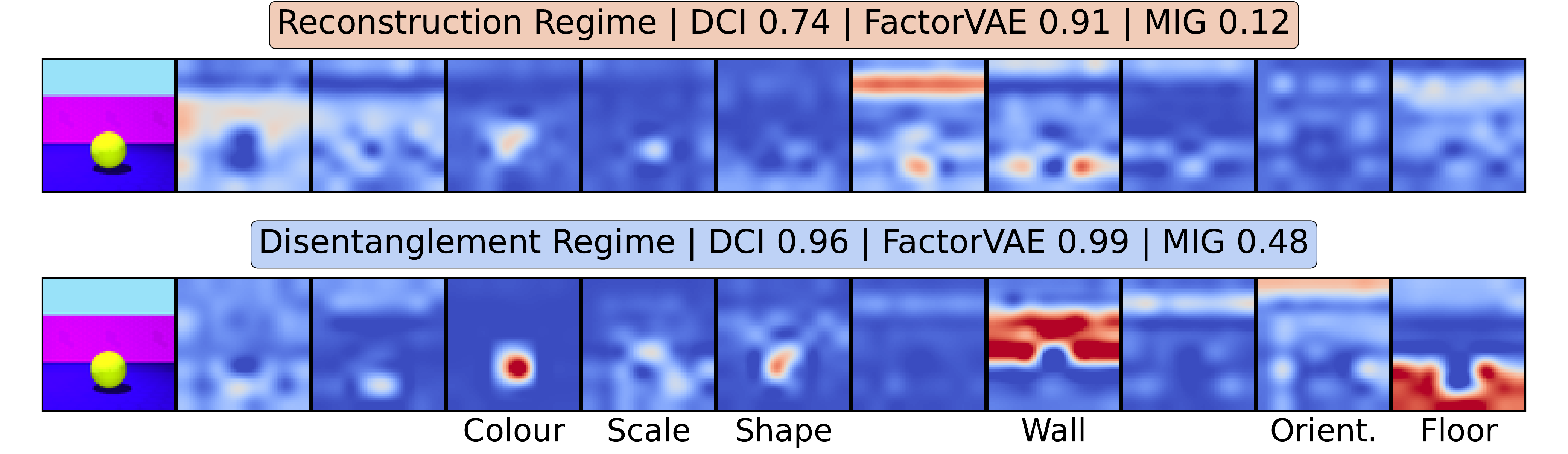}
    \includegraphics[width=0.49\textwidth]{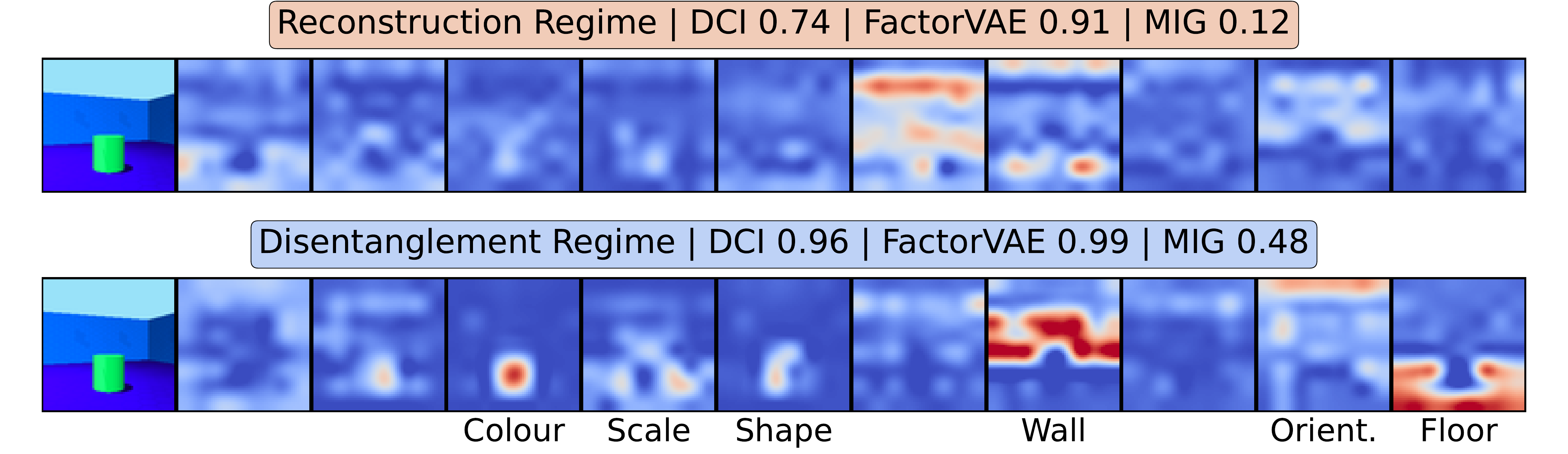} \\[1em]
    \includegraphics[width=0.49\textwidth]{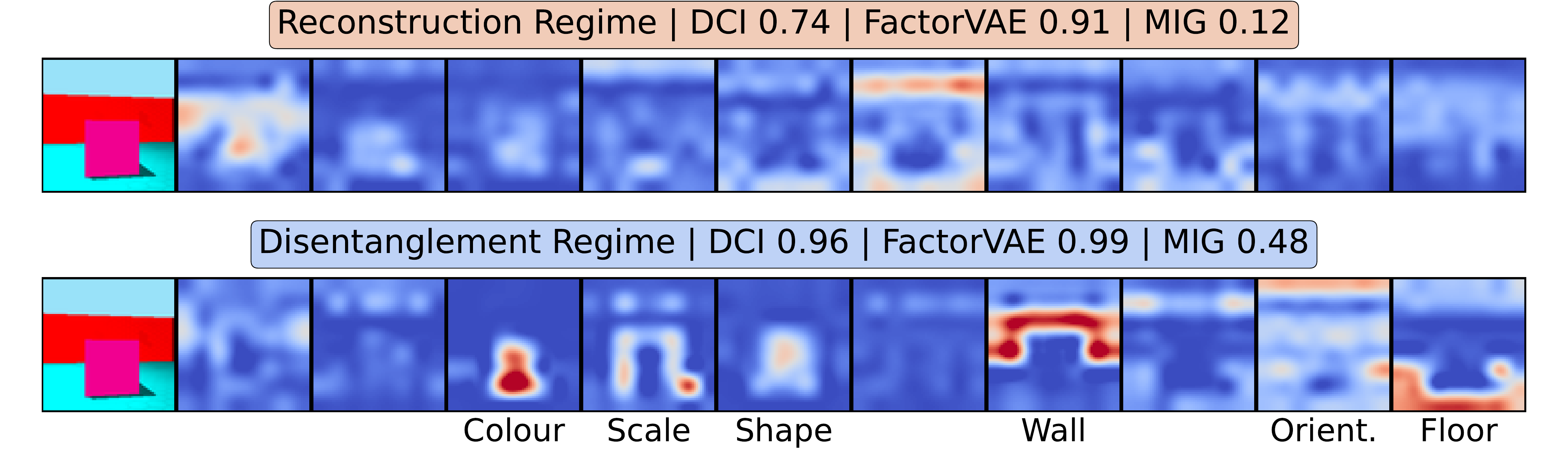}
    \includegraphics[width=0.49\textwidth]{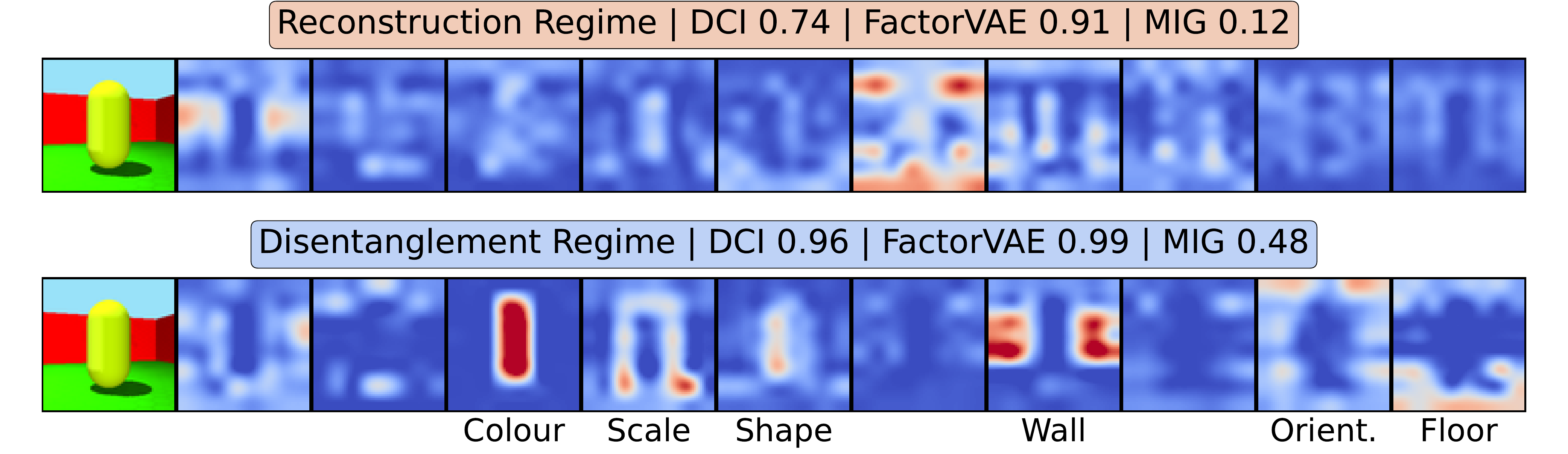}
    \captionof{figure}{
    \textbf{Cross-attention maps reveal semantic structure in the disentanglement regime.}
    These are taken from the models used in \Cref{fig:shapes3d_regimes}.
    At the same U-Net layer and logSNR level, attention maps in the disentanglement regime are localised and align with ground-truth factors, but those in the reconstruction regime are diffuse.
    }
    \label{fig:shapes3d_regimes_cross_attn}
\end{center}

\subsection{Skip Connections}
\label{app:zero_skip}

\begin{table}[!htbp]
\centering
\caption{
\textbf{Skip pathways trade reconstruction for disentanglement.}
We ablate skip connections in the EncDiff U-Net on Shapes3D over 10 seeds.
Zeroing skip features removes the encoder-decoder shortcut entirely, substantially improving disentanglement metrics but sharply degrading reconstruction.
Skip Dropout provides a softer intervention, preserving reconstruction while giving smaller disentanglement gains.
Combining zeroed skips with the $\logsnr$ curriculum further improves disentanglement, suggesting that restricting shortcut pathways can make the curriculum's trajectory-steering effect more pronounced (akin to \Cref{tab:ablations_shapes3d}).
}
\resizebox{\linewidth}{!}{
\scriptsize
\begin{tabular}{lcccc}
\toprule
\textbf{Model} & \textbf{LPIPS} $\downarrow$ & \textbf{DCI} $\uparrow$ & \textbf{FactorVAE} $\uparrow$ & \textbf{MIG} $\uparrow$ \\
\midrule
EncDiff U-Net + Split-MLP \citep{yang2024diffusion}
& $0.0025 \pm 0.0003$ 
& $0.828 \pm 0.099$ 
& $0.921 \pm 0.093$ 
& $0.218 \pm 0.087$ \\

\quad + $\mathrm{sigmoid(-\lambda)}$*
& $\mathbf{0.0020 \pm 0.0001}$
& $0.932 \pm 0.077$
& $0.971 \pm 0.056$
& $0.399 \pm 0.102$ \\

\quad + $\logsnr$ Curriculum & $0.0020 \pm 0.0003$ & $0.853 \pm 0.060$ & $0.917 \pm 0.063$ & $0.299 \pm 0.066$ \\

\midrule

Zeroed Skips + $\mathrm{sigmoid(-\lambda)}$*
& $0.1573 \pm 0.0221$ 
& $0.905 \pm 0.079$ 
& $0.940 \pm 0.062$ 
& $0.439 \pm 0.109$ \\

\quad + $\logsnr$ Curriculum 
& $0.2029 \pm 0.0049$ 
& $\mathbf{0.942 \pm 0.064}$ 
& $\mathbf{0.975 \pm 0.059}$ 
& $\mathbf{0.548 \pm 0.161}$ \\

\midrule

Skip Dropout \citep{jun2025disentangling} + $\mathrm{sigmoid(-\lambda)}$ 
& $0.0030 \pm 0.0003$ 
& $0.892 \pm 0.075$ 
& $0.954 \pm 0.052$ 
& $0.408 \pm 0.088$ \\

\quad + $\logsnr$ Curriculum 
& $0.0030 \pm 0.0002$ 
& $0.876 \pm 0.101$ 
& $0.935 \pm 0.061$ 
& $0.397 \pm 0.123$ \\

\bottomrule
\end{tabular}
}
\label{tab:zero_skip_results}
\end{table}

\begin{figure}[t]
    \centering
    \begin{subfigure}{\textwidth}
        \centering
        \includegraphics[width=\linewidth]{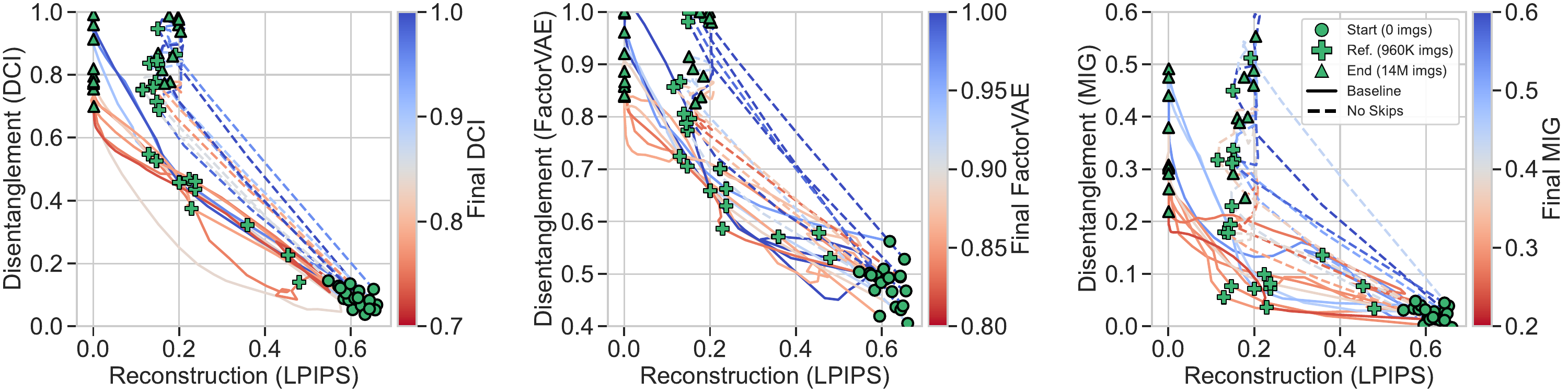}
        \caption{Zeroed skips.}
        \label{fig:shapes3d_regimes_zeroed}
    \end{subfigure}
    \hfill
    \begin{subfigure}{\textwidth}
        \centering
        \includegraphics[width=\linewidth]{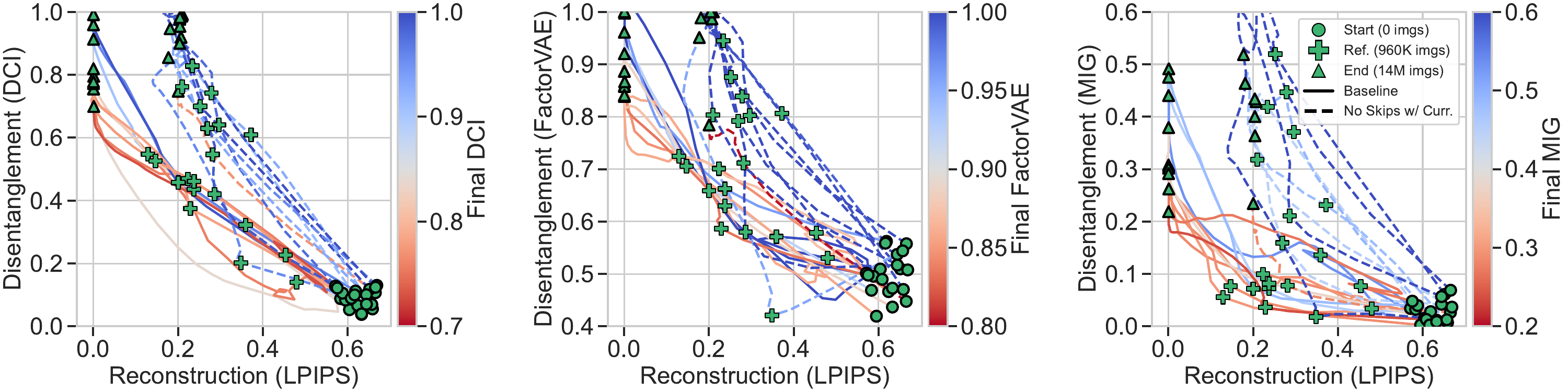}
        \caption{Zeroed skips + $\logsnr$ curriculum from $[0, 10]$ to $[-5, 12]$ over $32$K steps.}
        \label{fig:shapes3d_regimes_zeroed_curriculum}
    \end{subfigure}

    \caption{
    \textbf{Removing skip connections steers trajectories toward disentanglement.}
    We follow the setup in Figure~1 and compare the EncDiff baseline against zeroed-skip variants.
    Zeroing skip connections shifts training trajectories upward toward the disentanglement regime, but also moves them leftward more slowly, reflecting worse reconstruction.
    Adding the $\logsnr$ curriculum on top of zeroed skips steers trajectories to the high-disentanglement region, especially for MIG.
    Together with Table~\ref{tab:zero_skip_results}, these results support H1: skip connections act as reconstruction-oriented shortcuts, and regulating them can make optimisation rely more on the representation pathway.
    }
    \label{fig:zeroed_skip_trajectories}
\end{figure}

We test the role of skip connections in unsupervised diffusion representation learning by modifying the skip pathways of the EncDiff U-Net used in our Shapes3D analysis.
In the \textit{zeroed skips} setting, we replace the encoder skip features passed to the decoder with zeros, removing the shortcut pathway and forcing the decoder to rely more heavily on the U-Net trunk and conditioning representation.
We also compare against Skip Dropout \citep{jun2025disentangling}, which stochastically drops skip-connection features.

Table 12 shows that removing skip connections substantially worsens reconstruction and has mixed effects on disentanglement. 
Under the same sigmoid weighting, zeroing skips increases MIG but lowers mean DCI and FactorVAE.
Adding the log SNR curriculum to the zeroed-skip model improves all three disentanglement metrics.
However, reconstruction degrades sharply, indicating that the skip pathway is useful for image fidelity even when it is harmful for representation learning.
Skip Dropout gives a softer version of this effect, preserving reconstruction quality while improving disentanglement, but the gains are smaller and the $\logsnr$ curriculum does not further improve performance.
This mirrors the behaviour of the standard U-Net with curriculum in \Cref{tab:ablations_shapes3d}, suggesting that when skip connections are present, their reconstruction bias can dominate the curriculum signal.

The trajectory plots in \Cref{fig:zeroed_skip_trajectories} make this behaviour visible.
With zeroed skips, trajectories shift toward the disentanglement regime, achieving higher final representation scores but worse LPIPS.
Adding the $\logsnr$ curriculum preserves this shift and further concentrates trajectories in the high-disentanglement region, particularly for MIG.
Together, these results support \textbf{H1}: skip connections are not merely an implementation detail, but a reconstruction-oriented shortcut that can determine whether optimisation uses the representation pathway or bypasses it.

\noindent\textbf{Zeroed skips + $\logsnr$ curriculum.}
The zeroed-skips + $\logsnr$ curriculum ablation directly probes the interaction between H1 and H2.
Removing skip pathways shifts trajectories toward the disentanglement regime, while adding the $\logsnr$ curriculum shifts them further into the disentanglement region, especially for MIG.
This suggests that noise-level exposure can steer optimisation more effectively once reconstruction shortcuts are restricted.
Together, this supports our main conclusion: the curriculum selects the regime, while architectural bottlenecks and loss-weighting improve stability within it.
Since zeroing skips sharply degrades reconstruction, we instead regulate skip pathways with gated residuals, preserving a weaker reconstruction pathway.

\subsection{Early Noise-Level Exposure}
\label{app:early_noise_level_exposure}

\Cref{tab:fixed_band_subset} shows an early-time slice at $320$K images, using the 150K VQ-VAE checkpoint.
The three bands are chosen to distinguish full noise-level support, removal of high-noise levels, and high-noise only.
The full band $[-5,12]$ gives the strongest early reconstruction among the three settings, reaching LPIPS $0.0348 \pm 0.0138$, while also achieving competitive disentanglement.
Removing high-noise levels with $[0,12]$ slows reconstruction, increasing LPIPS to $0.1721 \pm 0.0413$, but does not collapse the representation: DCI and FactorVAE decrease moderately, while MIG remains comparable to the full band.
Conversely, the high-noise band $[-5,0]$ gives the worst reconstruction, with LPIPS $0.2929 \pm 0.0021$, but the strongest early disentanglement scores.
These results show that early noise-level exposure changes the relative pace of reconstruction and semantic organisation: full support accelerates reconstruction, whereas restricted bands preserve or improve early disentanglement despite worse image fidelity.
This supports \textbf{H2}: noise-level exposure changes the path through reconstruction-disentanglement space, and may therefore influence regime selection.

\begin{table}[!htbp]
\centering
\caption{
\textbf{Representative early fixed-band dynamics used to support \textbf{H2}.}
Metrics are measured at 320K images over 10 seeds using the 150K VQ-VAE checkpoint.
The selected bands compare full support, removing high-noise levels, and using only high-noise levels.
}
\label{tab:fixed_band_subset}
\scriptsize
\begin{tabular}{lcccccc}
\toprule
Description & $\lambda_{\min}$ & $\lambda_{\max}$ & LPIPS$\downarrow$ & DCI$\uparrow$ & FactorVAE$\uparrow$ & MIG$\uparrow$ \\
\midrule
Full band       & $-5$ & $12$ & $0.0348 \pm 0.0138$ & $0.614 \pm 0.053$ & $0.716 \pm 0.122$ & $0.100 \pm 0.049$ \\
No high-noise   & $0$  & $12$ & $0.1721 \pm 0.0413$ & $0.508 \pm 0.073$ & $0.673 \pm 0.078$ & $0.118 \pm 0.073$ \\
Only high-noise & $-5$ & $0$  & $0.2929 \pm 0.0021$ & $0.635 \pm 0.070$ & $0.786 \pm 0.065$ & $0.184 \pm 0.063$ \\
\bottomrule
\end{tabular}
\end{table}

\subsection{Noise-Level Exposure and VQ-VAE Effects}
\label{app:noise_level_ablations}

To isolate the role of noise-level support, we train the toy diffusion autoencoder with fixed $\logsnr$ bands under the $\mathrm{sigmoid}(-\lambda)$ loss-weighting used in our main experiments.
We report results with two VQ-VAE checkpoints: an earlier 30K-step checkpoint used for the trajectory analysis in \Cref{sec:analysis}, and a later 150K-step checkpoint used to more closely follow EncDiff and obtain stronger reconstruction.
The 30K VQ-VAE produces the clearest regime separation, while the 150K VQ-VAE improves reconstruction and makes the regime separation less pronounced.

The fixed-band ablations show that reconstruction is highly sensitive to the high-noise part of the objective, whereas disentanglement is less tied to full noise-level support.
For the 30K VQ-VAE, removing high-noise levels worsens LPIPS from $0.0177$ to $0.4044$, but disentanglement does not collapse: FactorVAE and MIG increase, while DCI decreases.
The strongest disentanglement for this checkpoint comes from restricted bands, with $[-5,5]$ giving the best DCI and FactorVAE, and $[0,12]$ giving the best MIG.
However, using only high-noise levels $[-5,0]$ performs poorly across all disentanglement metrics, suggesting that high-noise exposure alone is insufficient to organise factors in $\z$.
For the 150K VQ-VAE, removing high-noise levels again worsens reconstruction, increasing LPIPS from $0.0017$ to $0.0671$, and using only high-noise levels gives very poor LPIPS ($0.2975$).
Disentanglement varies less sharply for this later checkpoint, but restricted bands remain competitive with the full band, with several bands improving FactorVAE or MIG despite worse reconstruction.
Thus, removing high-noise levels substantially worsens reconstruction, while some restricted bands achieve competitive or improved disentanglement.

\begin{table}[!htbp]
\centering
\caption{Fixed logSNR band ablations (VQ-VAE trained for 150K steps, Diffusion Autoencoder sees 3M images, averaged over 10 seeds).}
\scriptsize
\begin{tabular}{lcccccc}
\toprule
Description & $\lambda_{\min}$ & $\lambda_{\max}$ & LPIPS $\downarrow$ & DCI $\uparrow$ & FactorVAE $\uparrow$ & MIG $\uparrow$ \\
\midrule
Full band & $-5$ & $12$ & $0.0017 \pm 0.0007$ & $0.744 \pm 0.093$ & $0.814 \pm 0.085$ & $0.186 \pm 0.085$ \\
No high-noise & $0$ & $12$ & $0.0671 \pm 0.0167$ & $0.703 \pm 0.117$ & $0.793 \pm 0.120$ & $0.231 \pm 0.123$ \\
No low-noise & $-5$ & $10$ & $0.0015 \pm 0.0002$ & $0.684 \pm 0.041$ & $0.789 \pm 0.045$ & $0.176 \pm 0.045$ \\
No extremes & $0$ & $10$ & $0.0614 \pm 0.0182$ & $0.716 \pm 0.071$ & $0.825 \pm 0.082$ & $0.231 \pm 0.109$ \\
Middle band & $-5$ & $5$  & $0.0021 \pm 0.0001$ & $0.722 \pm 0.064$ & $0.829 \pm 0.107$ & $0.187 \pm 0.069$ \\
Only high-noise & $-5$ & $0$ & $0.2975 \pm 0.0006$ & $0.739 \pm 0.097$ & $0.859 \pm 0.096$ & $0.231 \pm 0.071$ \\
\bottomrule
\end{tabular}
\end{table}

\begin{table}[!htbp]
\centering
\caption{Fixed logSNR band ablations (VQ-VAE trained for 30K steps, Diffusion Autoencoder sees 3M images, averaged over 10 seeds).}
\scriptsize
\begin{tabular}{lcccccc}
\toprule
Description & $\lambda_{\min}$ & $\lambda_{\max}$ & LPIPS $\downarrow$ & DCI $\uparrow$ & FactorVAE $\uparrow$ & MIG $\uparrow$ \\
\midrule
Full band* & $-5$ & $12$ & $0.0177 \pm 0.0399$ & $0.750 \pm 0.102$ & $0.868 \pm 0.088$ & $0.268 \pm 0.121$ \\
No high-noise* & $0$ & $12$ & $0.4044 \pm 0.0725$ & $0.665 \pm 0.202$ & $0.892 \pm 0.153$ & $0.381 \pm 0.157$ \\
No low-noise & $-5$ & $10$ & $0.0054 \pm 0.0012$ & $0.790 \pm 0.105$ & $0.872 \pm 0.122$ & $0.283 \pm 0.139$ \\
No extremes & $0$ & $10$ & $0.3938 \pm 0.0892$ & $0.694 \pm 0.158$ & $0.917 \pm 0.112$ & $0.358 \pm 0.147$ \\
Middle band & $-5$ & $5$ & $0.0063 \pm 0.0011$ & $0.834 \pm 0.118$ & $0.926 \pm 0.085$ & $0.313 \pm 0.134$ \\
Only high-noise & $-5$ & $0$ & $0.2021 \pm 0.0006$ & $0.641 \pm 0.069$ & $0.778 \pm 0.083$ & $0.166 \pm 0.059$ \\
\bottomrule
\end{tabular}
\end{table}

These results motivate our curriculum.
Rather than removing reconstruction-favouring noise levels entirely, we delay exposure to them.
This gives the representation time to organise under a bounded noise range before the model sees the full reconstruction objective.
The VQ-VAE checkpoint also matters: as the VQ-VAE is trained longer, reconstruction improves, but disentanglement becomes less separated across bands.
One possible explanation is that a more mature VQ-VAE encodes finer visual detail directly in the latent space, making reconstruction easier and reducing the pressure to organise information through $\z$.
We therefore use the 30K VQ-VAE to study regime emergence, where the effect is most visible, and the 150K VQ-VAE for experiments that more closely follow EncDiff.

\begin{table}[!htbp]
\centering
\caption{Relative change when training the VQ-VAE longer (150K vs 30K). Values are $(150\text{K} - 30\text{K}) / 30\text{K}$.}
\scriptsize
\begin{tabular}{lcccccc}
\toprule
Description & $\lambda_{\min}$ & $\lambda_{\max}$ & $\Delta$ LPIPS $\downarrow$ & $\Delta$ DCI $\uparrow$ & $\Delta$ FactorVAE $\uparrow$ & $\Delta$ MIG $\uparrow$ \\
\midrule
Full band 
& $-5$ & $12$
& \textcolor{green!60!black}{$-0.904$}
& \textcolor{red}{$-0.008$}
& \textcolor{red}{$-0.062$}
& \textcolor{red}{$-0.306$} \\

No high-noise 
& $0$ & $12$
& \textcolor{green!60!black}{$-0.834$}
& \textcolor{green!60!black}{$+0.057$}
& \textcolor{red}{$-0.111$}
& \textcolor{red}{$-0.394$} \\

No low-noise 
& $-5$ & $10$
& \textcolor{green!60!black}{$-0.722$}
& \textcolor{red}{$-0.134$}
& \textcolor{red}{$-0.095$}
& \textcolor{red}{$-0.378$} \\

No extremes 
& $0$ & $10$
& \textcolor{green!60!black}{$-0.844$}
& \textcolor{green!60!black}{$+0.032$}
& \textcolor{red}{$-0.100$}
& \textcolor{red}{$-0.355$} \\

Middle band 
& $-5$ & $5$
& \textcolor{green!60!black}{$-0.667$}
& \textcolor{red}{$-0.134$}
& \textcolor{red}{$-0.105$}
& \textcolor{red}{$-0.403$} \\

Only high-noise 
& $-5$ & $0$
& \textcolor{red}{$+0.472$}
& \textcolor{green!60!black}{$+0.153$}
& \textcolor{green!60!black}{$+0.104$}
& \textcolor{green!60!black}{$+0.392$} \\

\bottomrule
\end{tabular}
\end{table}

\subsection{$\logsnr$-curriculum}
\label{app:curriculum_ablations}

We ablate the initial $\logsnr$ band and widening duration used by the curricula in \Cref{fig:logsnr_ablations_traj}.
All settings use the same gated residual U-Net and $\mathrm{sigmoid}(-\lambda)$ loss-weighting, and widen to the same final band $[-5,12]$.
Thus, the ablation isolates how the early noise-level support and the speed of widening affect the training trajectory.

\Cref{tab:curriculum_ablations} shows that all curricula reach similar final reconstruction, with LPIPS saturated at approximately $0.0005$.
The main differences are therefore in representation quality.
With a fixed widening duration of $32$K, changing the initial band changes the final disentanglement scores: the high-noise initial band $[-5,0]$ gives strong DCI and FactorVAE, while the moderate band $[0,10]$ gives the best overall combination once the widening duration is tuned.
For the $[0,10]$ initial band, widening over $32$K steps provides the best balance across disentanglement metrics, as seen by the Macro Mean and Macro SD, whereas faster widening ($8$K) and slower widening ($64$K) both reduce at least one disentanglement metric.
This suggests that curriculum design matters beyond simply delaying access to high or low-noise levels; the rate at which the full objective is introduced also affects the representation learned.

\begin{figure}[!htbp]
    \centering
    % \frame{
    \includegraphics[width=\linewidth, trim={0mm 0mm 0mm 0mm}, clip]{figures/DCI_ablations_trajectories_median.pdf}
    \includegraphics[width=\linewidth, trim={0mm 0mm 0mm 0mm}, clip]{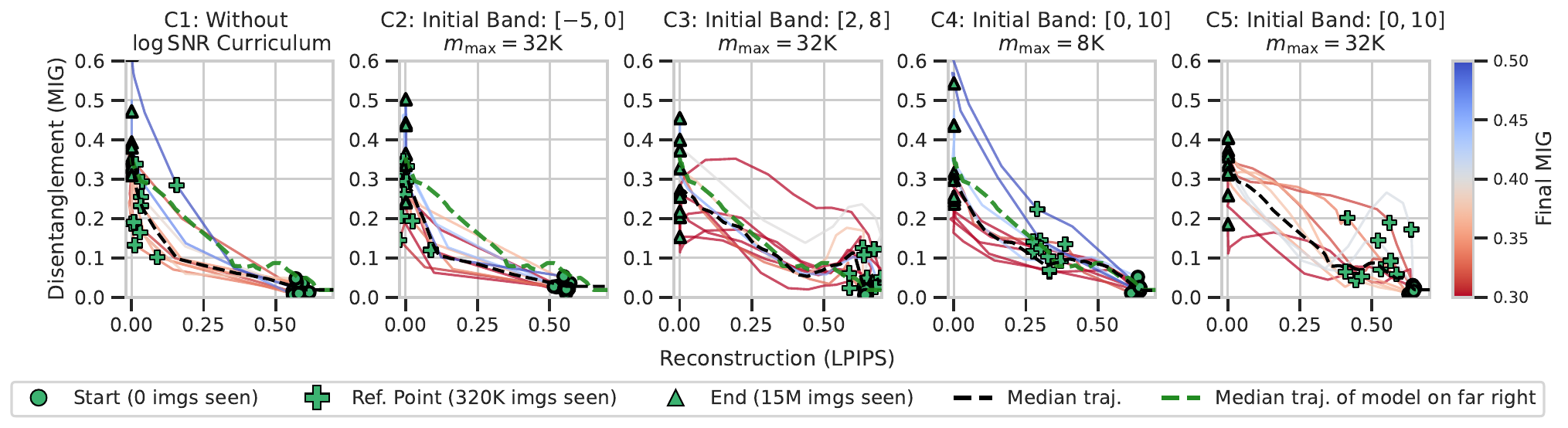}
    \includegraphics[width=\linewidth, trim={0mm 0mm 0mm 0mm}, clip]{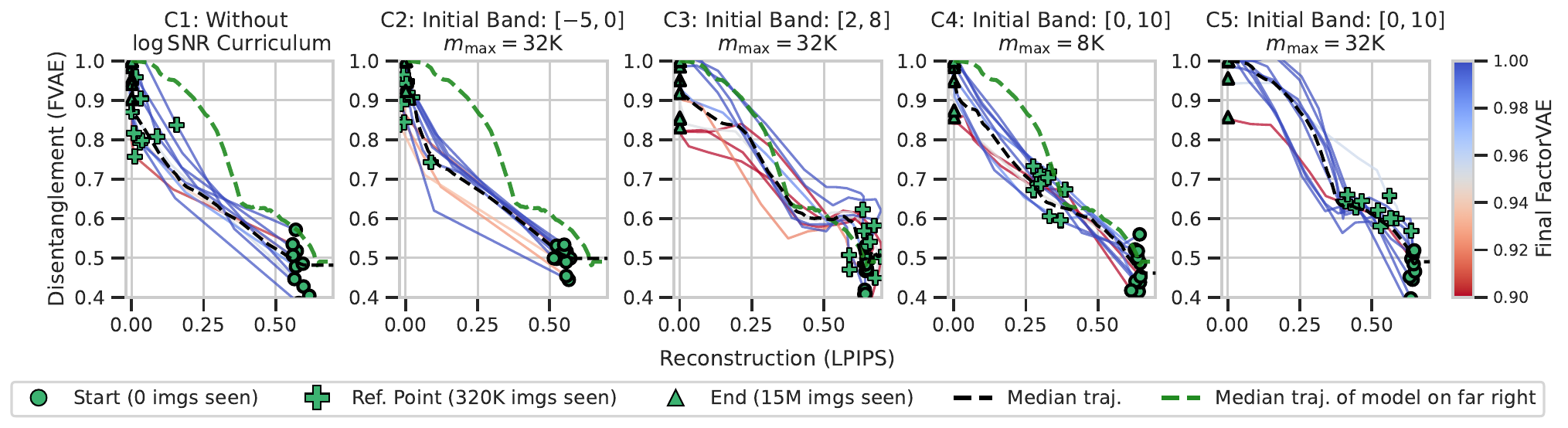}
    \caption{
\textbf{$\logsnr$ curricula steer training trajectories across disentanglement metrics.}
Training trajectories for curriculum ablations on Shapes3D, using DCI, FactorVAE, and MIG as the disentanglement axis.
Panels correspond to different curricula; colours indicate final disentanglement, and dashed curves show median trajectories.
Changing the initial band and widening duration changes the path through reconstruction--disentanglement space, not just the final score.
The moderate curriculum $[0,10]$ with $m_{\max}=32$K gives the strongest overall disentanglement, supporting curriculum design as a mechanism for regime selection.
}
\label{fig:curriculum_trajectories_appendix}
\end{figure}

\Cref{fig:curriculum_trajectories_appendix} shows the corresponding training trajectories for DCI, FactorVAE, and MIG.
The trajectories reveal that curricula change the path through reconstruction-disentanglement space before final convergence.
In particular, the best-performing curriculum follows a different trajectory, delaying early commitment to a reconstruction-dominated path and reaching high final disentanglement with lower variance across seeds.
These results support \textbf{H2}: early noise-level exposure can steer regime selection by changing the relative pace of reconstruction and semantic organisation.

\newpage

% \begin{table}[!htbp]
%     \centering
%     \caption{
%     \textbf{Curriculum ablations on Shapes3D.}
%     We vary the initial $\logsnr$ band $[\lambda_L(0), \lambda_U(0)]$ and widening duration $m_{\max}$, keeping the final band fixed to $[\lambda_{\min}, \lambda_{\max}] = [-5,12]$.
%     Metrics are reported after 15M images over 10 seeds.
%     }
%     \label{tab:curriculum_ablations}
%     \scriptsize
%     \begin{tabular}{ccccccc}
%     \toprule
%     $\lambda_{L}(0)$ & $\lambda_{U}(0)$ & $m_{\max}$ & LPIPS $\downarrow$ & DCI $\uparrow$ & FactorVAE $\uparrow$ & MIG $\uparrow$ \\
%     \midrule
%     $0$  & $12$       & $32$K & $0.0005 \pm 0.0000$ & $0.834 \pm 0.065$ & $0.935 \pm 0.045$ & $0.329 \pm 0.097$ \\
%     $2$  & $8$        & $32$K & $0.0005 \pm 0.0000$ & $0.860 \pm 0.100$ & $0.941 \pm 0.068$ & $0.290 \pm 0.097$ \\
%     $-5$ & $0$        & $32$K & $0.0005 \pm 0.0000$ & $0.914 \pm 0.096$ & $0.976 \pm 0.039$ & $0.348 \pm 0.090$ \\
%     \midrule
%     $0$  & $10$       & $8$K  & $0.0005 \pm 0.0000$ & $0.883 \pm 0.063$ & $0.968 \pm 0.050$ & $\mathbf{0.375 \pm 0.146}$ \\
%     $0$  & $10$       & $32$K & $0.0005 \pm 0.0000$ & $\mathbf{0.926 \pm 0.043}$ & $\mathbf{0.991 \pm 0.017}$ & $0.353 \pm 0.056$ \\
%     $0$  & $10$       & $64$K & $0.0005 \pm 0.0000$ & $0.864 \pm 0.038$ & $0.954 \pm 0.046$ & $0.266 \pm 0.099$ \\
%     \bottomrule
%     \end{tabular}
% \end{table}

\begin{table}[!htbp]
    \centering
    \caption{
    \textbf{Curriculum ablations on Shapes3D.}
    We vary the initial $\logsnr$ band $[\lambda_L(0), \lambda_U(0)]$ and widening duration $m_{\max}$, keeping the final band fixed to $[\lambda_{\min}, \lambda_{\max}] = [-5,12]$.
    Metrics are reported after 15M images over 10 seeds.
    Macro Mean and Macro SD average the means and standard deviations across DCI, FactorVAE, and MIG, respectively.
    }
    \label{tab:curriculum_ablations}
    % \scriptsize
    \resizebox{\linewidth}{!}{
    \begin{tabular}{ccccccccc}
    \toprule
    $\lambda_{L}(0)$ & $\lambda_{U}(0)$ & $m_{\max}$ & LPIPS $\downarrow$ & DCI $\uparrow$ & FactorVAE $\uparrow$ & MIG $\uparrow$ & Macro Mean $\uparrow$ & Macro SD $\downarrow$ \\
    \midrule
    $0$  & $12$ & $32$K & $0.0005 \pm 0.0000$ & $0.834 \pm 0.065$ & $0.935 \pm 0.045$ & $0.329 \pm 0.097$ & $0.699$ & $0.069$ \\
    $2$  & $8$  & $32$K & $0.0005 \pm 0.0000$ & $0.860 \pm 0.100$ & $0.941 \pm 0.068$ & $0.290 \pm 0.097$ & $0.697$ & $0.088$ \\
    $-5$ & $0$  & $32$K & $0.0005 \pm 0.0000$ & $0.914 \pm 0.096$ & $0.976 \pm 0.039$ & $0.348 \pm 0.090$ & $0.746$ & $0.075$ \\
    \midrule
    $0$ & $10$ & $8$K  & $0.0005 \pm 0.0000$ & $0.883 \pm 0.063$ & $0.968 \pm 0.050$ & $\mathbf{0.375 \pm 0.146}$ & $0.742$ & $0.086$ \\
    $0$ & $10$ & $32$K & $0.0005 \pm 0.0000$ & $\mathbf{0.926 \pm 0.043}$ & $\mathbf{0.991 \pm 0.017}$ & $0.353 \pm 0.056$ & $\mathbf{0.757}$ & $\mathbf{0.039}$ \\
    $0$ & $10$ & $64$K & $0.0005 \pm 0.0000$ & $0.864 \pm 0.038$ & $0.954 \pm 0.046$ & $0.266 \pm 0.099$ & $0.695$ & $0.061$ \\
    \bottomrule
    \end{tabular}
    }
\end{table}

\newpage

\subsection{Attribute Disentanglement}

\begin{figure}[!htbp]
    \centering

    % --- Row 1: Cars3D ---
    \begin{subfigure}[t]{0.9\textwidth}
        \centering
        \includegraphics[width=\linewidth]{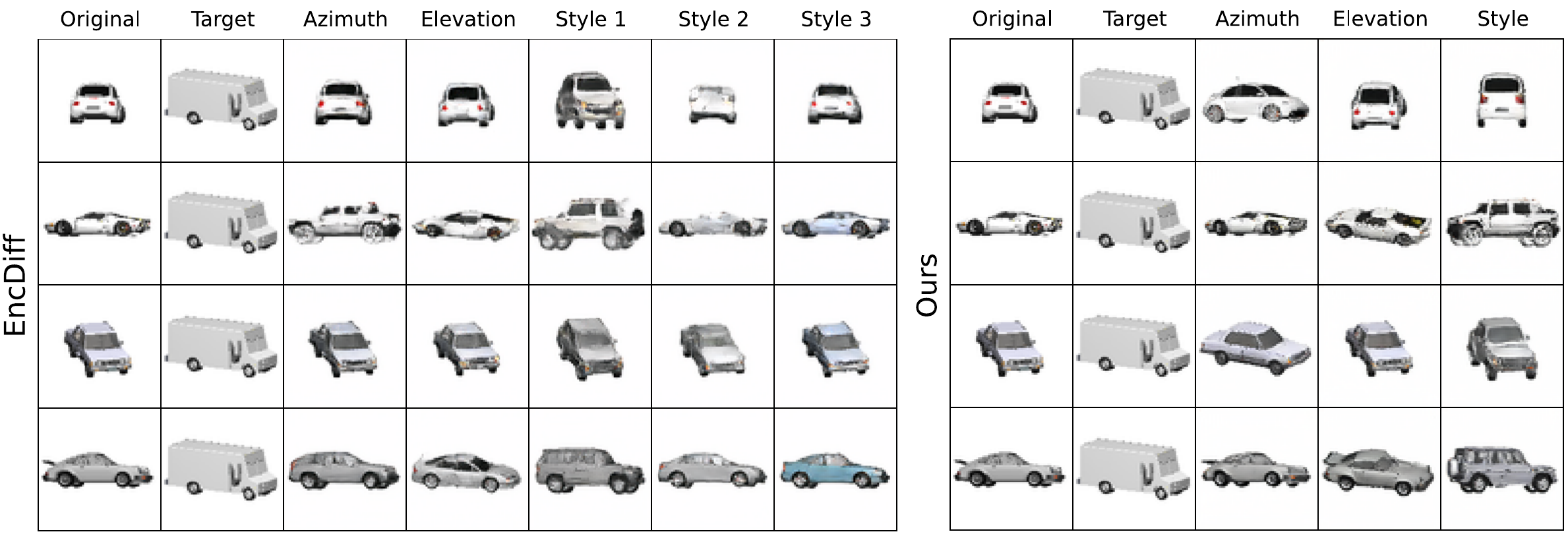}
        \caption{Cars3D example 1}
    \end{subfigure}
    % \hfill
    \begin{subfigure}[t]{0.9\textwidth}
        \centering
        \includegraphics[width=\linewidth]{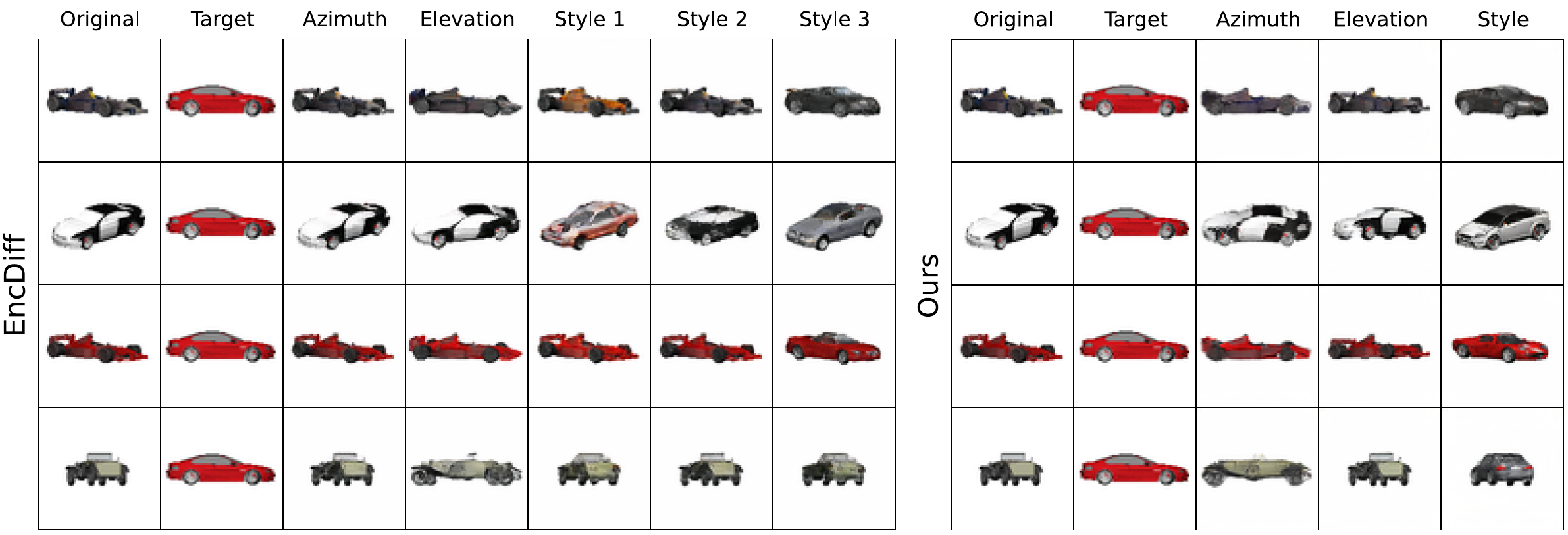}
        \caption{Cars3D example 2}
    \end{subfigure}

    % \vspace{1mm}

    % --- Row 2: Shapes3D ---
    \begin{subfigure}[t]{\textwidth}
        \centering
        \includegraphics[width=\linewidth]{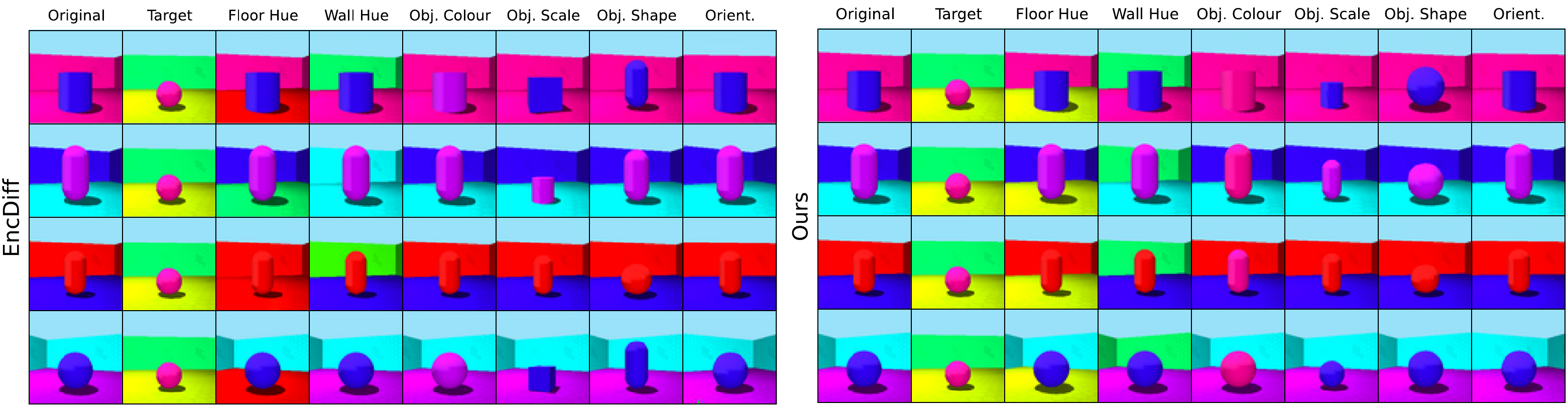}
        \caption{Shapes3D example 1}
    \end{subfigure}
    % \hfill
    \begin{subfigure}[t]{\textwidth}
        \centering
        \includegraphics[width=\linewidth]{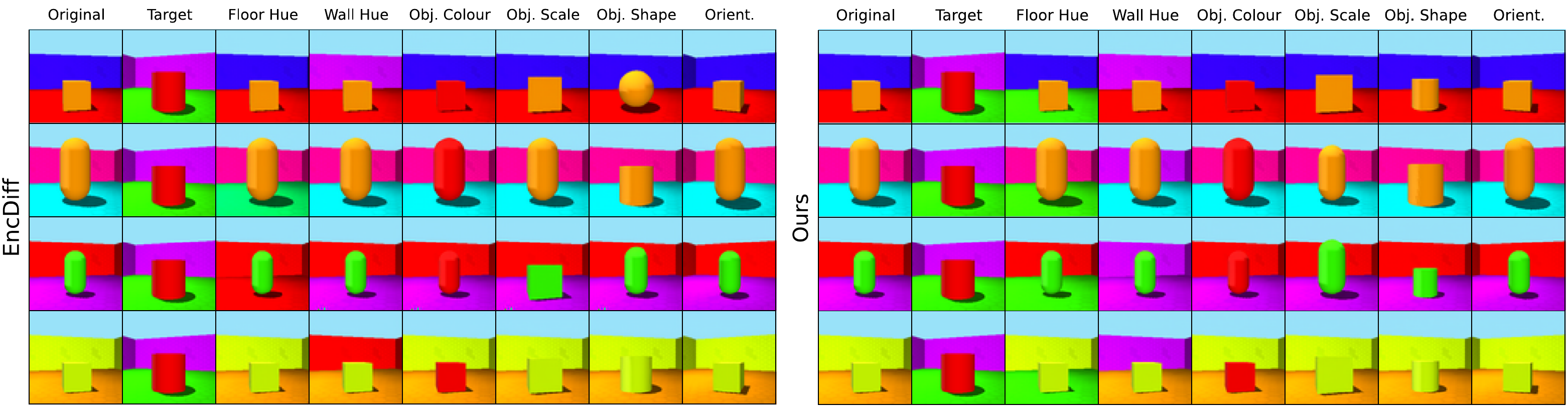}
        \caption{Shapes3D example 2}
    \end{subfigure}

    \caption{
    \textbf{Latent edits reveal interpretable factors across datasets.}
    Each panel compares EncDiff and \textsc{SteeringDRL} by replacing a single source slot with the corresponding target slot.
    On Cars3D, edits capture viewpoint (\texttt{Azimuth}, \texttt{Elevation}) and object identity (\texttt{Object Type / Style}).
    On Shapes3D with $N = 10$, we can more consistently perform all edits, whereas EncDiff often struggles with \texttt{Floor Hue} and \texttt{Object Shape}.
    }
    \label{fig:main_edits}
\end{figure}

\newpage

\subsection{Spatial Disentanglement with Object-Centric Learning}
\label{app:spatial_dis}

\subsubsection{ClevrTex}

\begin{figure}[!htbp]
    \centering

    \begin{minipage}[t]{0.72\textwidth}
        \centering
        \adjustbox{
            max width=\linewidth,
            max height=0.285\textheight
        }{
            \includegraphics{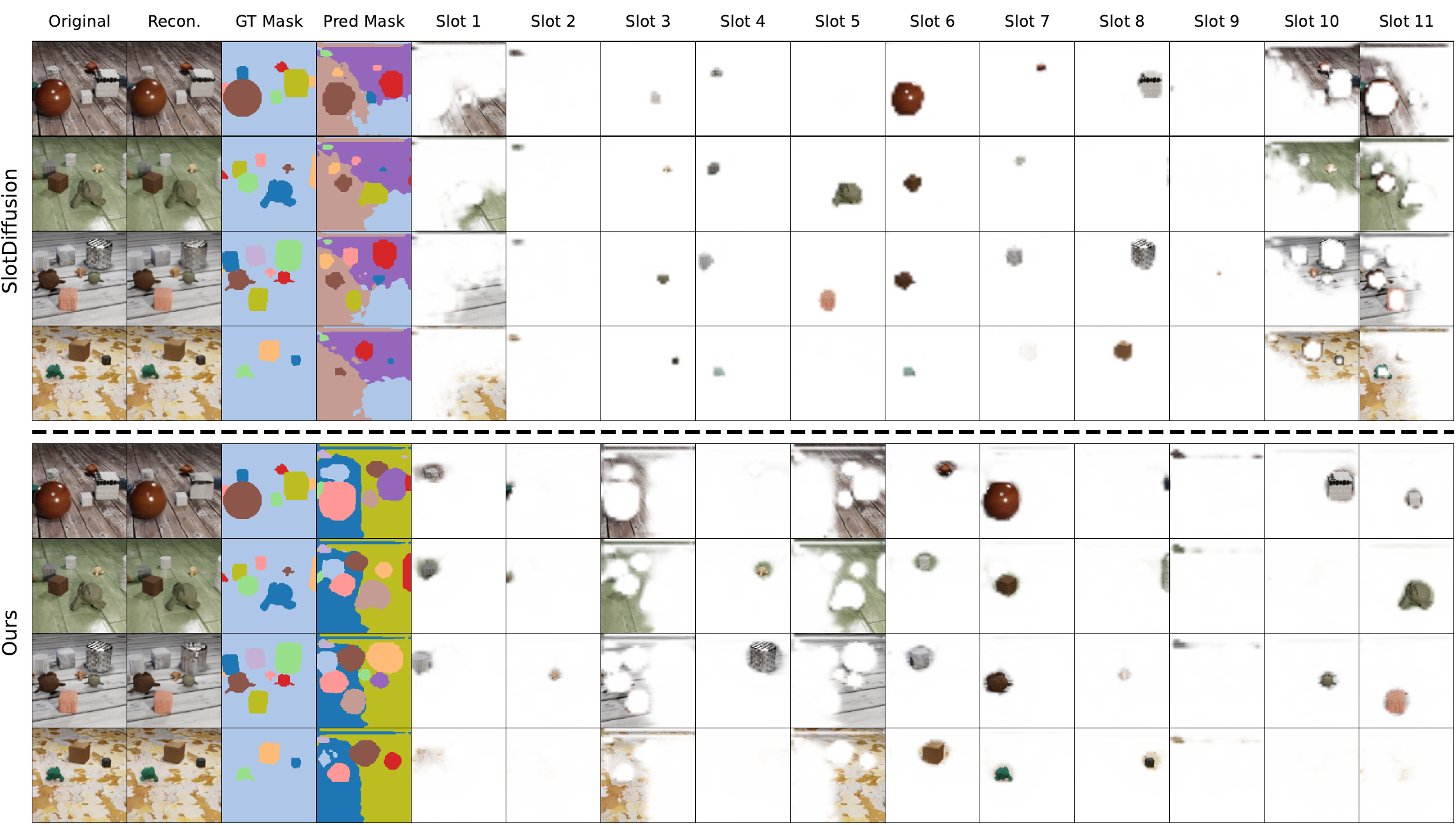}
        }

        % \vspace{2pt}
        {\small (a) Segmentation and slot decomposition}
    \end{minipage}
    \hfill
    \begin{minipage}[t]{0.26\textwidth}
        \centering
        \adjustbox{
            max width=\linewidth,
            max height=0.245\textheight
        }{
            \includegraphics{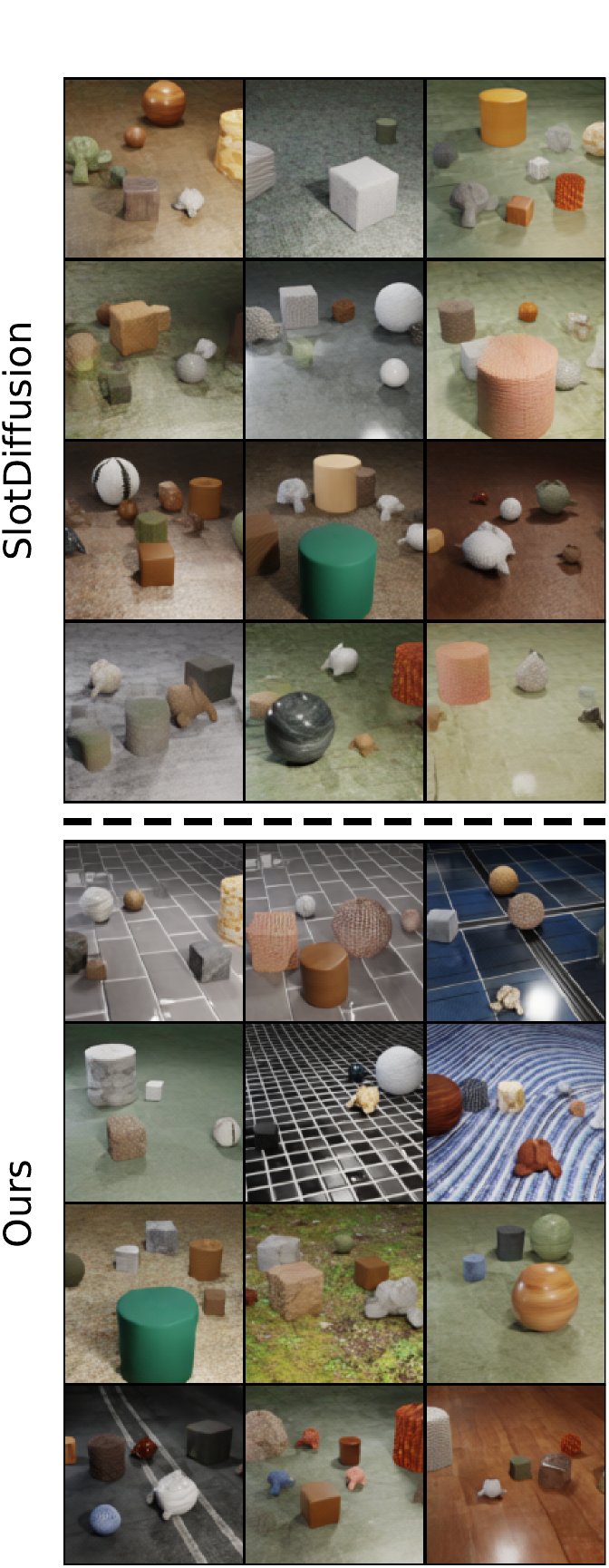}
        }

        % \vspace{2pt}
        {\small (b) Random compositions}
    \end{minipage}
    
\caption{
\textbf{\textsc{SteeringDRL} produces more spatially coherent and compositional slot representations on ClevrTex than SlotDiffusion.}
(a) Reading down each column, a slot captures objects in similar spatial locations across scenes, with fewer background slots compared to SlotDiffusion. 
% This yields cleaner segmentation masks and more structured slot decompositions.
(b) Fewer background slots reduce fragmentation when slots are shuffled for random composition.
}
    \label{fig:clevrtex_seg_and_comp}
\end{figure}

\begin{center}
    \includegraphics[width=0.9\linewidth]{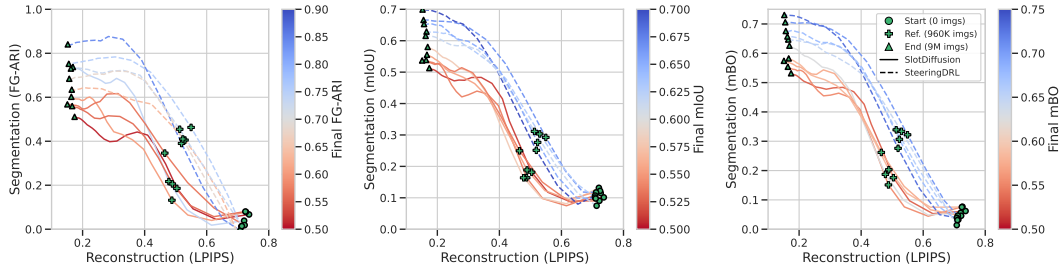}
\captionof{figure}{
\textbf{ClevrTex training trajectories across segmentation metrics.}
We plot reconstruction--segmentation trajectories for FG-ARI, mIoU and mBO using ResNet34 features.
\textsc{SteeringDRL} learns semantic slot structure earlier than SlotDiffusion and converges to higher segmentation scores. % while maintaining strong reconstruction.
}
\end{center}

\begin{table}[!htbp]
\centering
\caption{
\textbf{ClevrTex results with ResNet18 features.}
Here we use ResNet18 for $\Feat(\cdot)$.
Methods are run under the same data split and evaluation protocol over 5 seeds.
}
\label{tab:clevrtex_r18}
\scriptsize
\begin{tabular}{lccccc}
\toprule
& \multicolumn{3}{c}{\textbf{Segmentation}} 
& \multicolumn{1}{c}{\textbf{Recon.}} 
& \multicolumn{1}{c}{\textbf{Comp.}} \\
\cmidrule(lr){2-4} \cmidrule(lr){5-5} \cmidrule(lr){6-6}
\textbf{Method}
& \textbf{FG-ARI} $\uparrow$
& \textbf{mIoU} $\uparrow$
& \textbf{mBO} $\uparrow$
& \textbf{LPIPS} $\downarrow$
& \textbf{FID} $\downarrow$ \\
\midrule
SLATE 
& $62.75 \pm 5.89$
& $56.41 \pm 4.02$
& $60.23 \pm 4.01$
& $0.402 \pm 0.005$
& $88.73 \pm 5.37$ \\
SlotDiffusion
& $66.85 \pm 3.22$
& $56.01 \pm 1.98$
& $60.20 \pm 2.34$
& $0.173 \pm 0.015$
& $47.84 \pm 1.36$ \\
\textbf{\textsc{SteeringDRL}}
& $\mathbf{67.82 \pm 1.40}$
& $\mathbf{59.20 \pm 1.91}$
& $\mathbf{63.29 \pm 1.87}$
& $\mathbf{0.164 \pm 0.004}$
& $\mathbf{44.01 \pm 1.85}$ \\
\bottomrule
\end{tabular}
\end{table}

\begin{center}
    \includegraphics[width=0.9\linewidth]{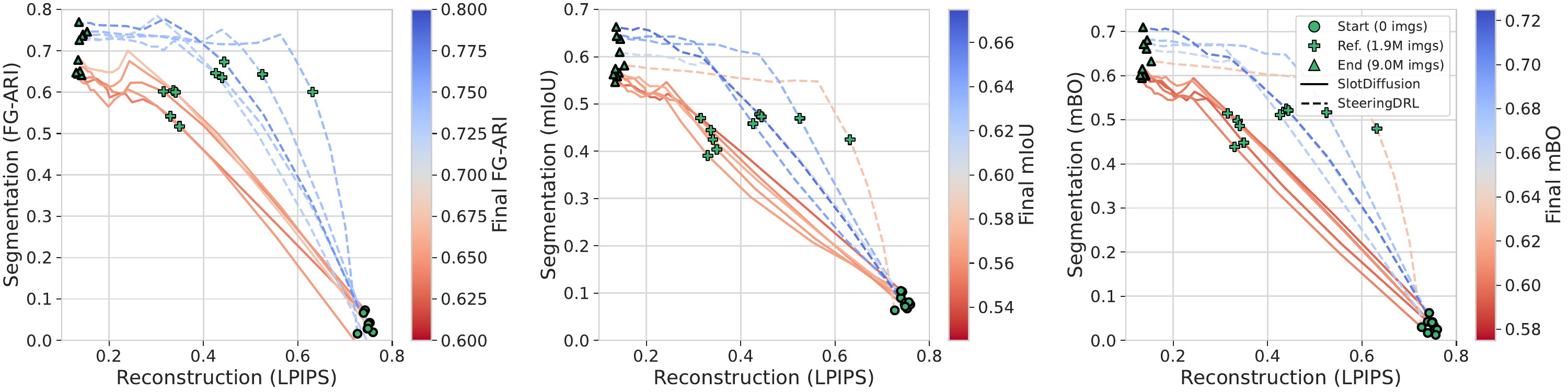}
\captionof{figure}{
\textbf{ClevrTex training trajectories with ResNet18 features.}
\textsc{SteeringDRL} still improves segmentation faster and converges to higher scores than SlotDiffusion.
}
\end{center}

% \newpage

\subsubsection{PascalVOC}
\label{app:pascal}

\begin{center}
    \centering
    \includegraphics[width=\textwidth]{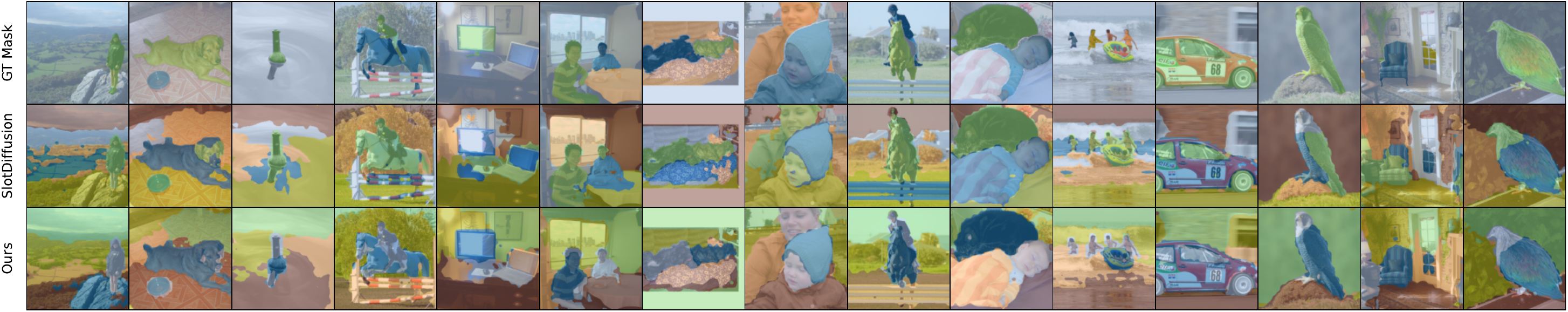}
    \captionof{figure}{
    \textbf{PascalVOC qualitative masks with pretrained DINO features.}
    \textsc{SteeringDRL} better exploits the semantic features in DINO to segment dominant objects, whereas SlotDiffusion often merges objects with the background or nearby regions.
    However, both methods still exhibit object-part segmentation, such as separating the head and body of birds, suggesting that further gains may require explicit semantic alignment as in GLASS \citep{singh2025glass}.
    }
\end{center}

\begin{center}
    \includegraphics[width=\linewidth]{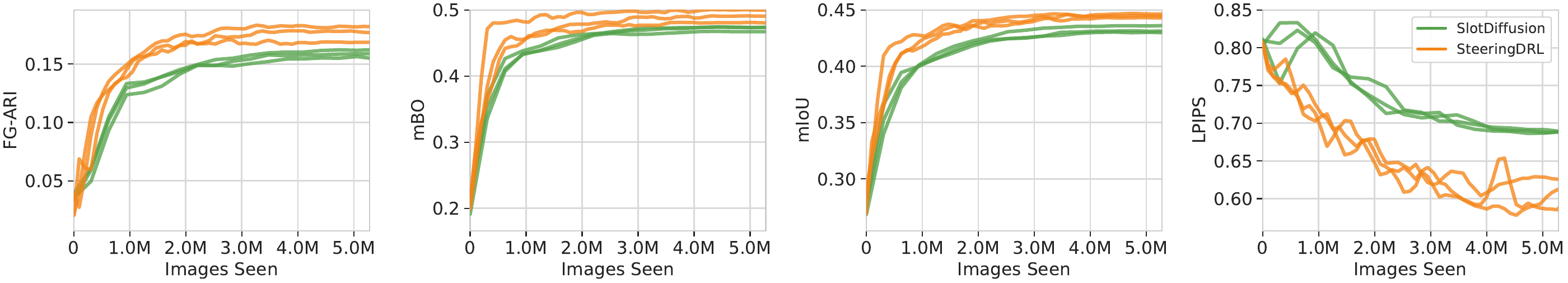}
    \captionof{figure}{
    \textbf{Individual PascalVOC trajectories show faster and stronger convergence.}
    We plot each metric against training progress for SlotDiffusion and \textsc{SteeringDRL}.
    With the same pretrained $\Feat(\cdot)$, \textsc{SteeringDRL} improves FG-ARI, mBO and mIoU earlier in training, while also reducing LPIPS faster.
    This supports the view that, with pretrained semantic features, our inductive biases improve the efficiency of object-centric binding.
    While we do improve reconstruction metrics, the visualisations remain poor, and not comparable to larger model such as StableLSD \citep{jiang2023object} or GLASS \citep{singh2025glass}, which use higher quality VQ-VAEs.
    }
\end{center}

\begin{center}
    \includegraphics[width=\linewidth]{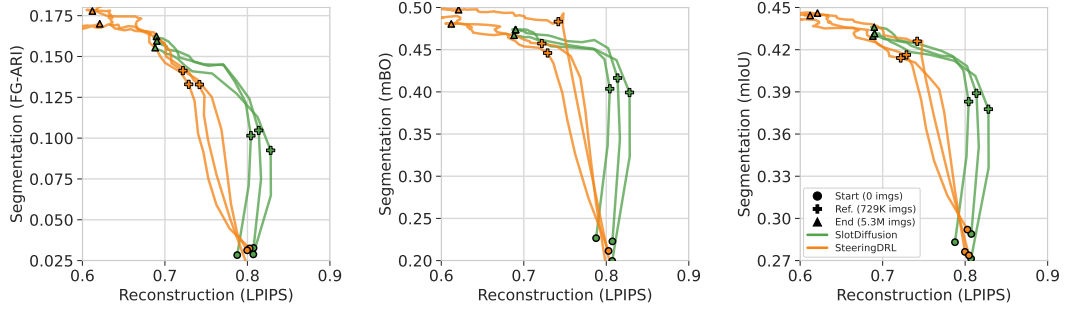}
    \captionof{figure}{
    \textbf{Training trajectories on PascalVOC differ from from-scratch OCL.}
    We plot reconstruction-segmentation trajectories for FG-ARI, mBO and mIoU on PascalVOC.
    Unlike ClevrTex, where representations must emerge from scratch, PascalVOC uses pretrained DINO features for $\Feat(\cdot)$.
    \textsc{SteeringDRL} exploits these pretrained semantics more quickly than SlotDiffusion, reaching higher segmentation scores and better reconstruction earlier in training.
    }
\end{center}

\newpage

\subsubsection{COCO}
\label{app:coco}

\begin{table}[!htbp]
    \centering
    \small
    \caption{
    \textbf{Preliminary COCO results follow the trends observed on Pascal VOC.}
    For spatial disentanglement with object-centric learning, we retain the matched SlotDiffusion baseline's training and evaluation setup and swap in the architectural and curriculum components from \Cref{sec:steering}.
    \textsc{SteeringDRL} accelerates improvements in segmentation (FG-ARI) and reconstruction (LPIPS), while mIoU and mBO show early gains before remaining closely matched to SlotDiffusion.
    Values are reported as mean $\pm$ standard deviation over 3 seeds; the best result at each checkpoint is bold.
    }
    \begin{tabular}{@{}clcccc@{}}
        \toprule
        \textbf{Step}
        & \textbf{Method}
        & \textbf{FG-ARI} $\uparrow$
        & \textbf{mIoU} $\uparrow$
        & \textbf{mBO} $\uparrow$
        & \textbf{LPIPS} $\downarrow$ \\
        \midrule
        $20\mathrm{K}$
        & SlotDiffusion
        & $0.282 \pm 0.013$
        & $0.287 \pm 0.003$
        & $0.293 \pm 0.005$
        & $0.863 \pm 0.010$ \\
        $20\mathrm{K}$
        & \textsc{SteeringDRL}
        & $\mathbf{0.321 \pm 0.012}$
        & $\mathbf{0.305 \pm 0.001}$
        & $\mathbf{0.304 \pm 0.002}$
        & $\mathbf{0.751 \pm 0.010}$ \\
        \midrule
        $40\mathrm{K}$
        & SlotDiffusion
        & $0.323 \pm 0.009$
        & $0.306 \pm 0.001$
        & $0.306 \pm 0.003$
        & $0.824 \pm 0.024$ \\
        $40\mathrm{K}$
        & \textsc{SteeringDRL}
        & $\mathbf{0.352 \pm 0.006}$
        & $\mathbf{0.313 \pm 0.003}$
        & $\mathbf{0.313 \pm 0.003}$
        & $\mathbf{0.730 \pm 0.010}$ \\
        \midrule
        $60\mathrm{K}$
        & SlotDiffusion
        & $0.347 \pm 0.006$
        & $\mathbf{0.314 \pm 0.002}$
        & $\mathbf{0.315 \pm 0.002}$
        & $0.786 \pm 0.010$ \\
        $60\mathrm{K}$
        & \textsc{SteeringDRL}
        & $\mathbf{0.360 \pm 0.012}$
        & $\mathbf{0.314 \pm 0.002}$
        & $0.313 \pm 0.003$
        & $\mathbf{0.706 \pm 0.001}$ \\
        \midrule
        $80\mathrm{K}$
        & SlotDiffusion
        & $0.349 \pm 0.001$
        & $0.313 \pm 0.001$
        & $0.315 \pm 0.004$
        & $0.790 \pm 0.019$ \\
        $80\mathrm{K}$
        & \textsc{SteeringDRL}
        & $\mathbf{0.374 \pm 0.008}$
        & $\mathbf{0.317 \pm 0.001}$
        & $\mathbf{0.317 \pm 0.002}$
        & $\mathbf{0.688 \pm 0.005}$ \\
        \midrule
        $100\mathrm{K}$
        & SlotDiffusion
        & $0.357 \pm 0.004$
        & $\mathbf{0.319 \pm 0.002}$
        & $\mathbf{0.320 \pm 0.002}$
        & $0.772 \pm 0.010$ \\
        $100\mathrm{K}$
        & \textsc{SteeringDRL}
        & $\mathbf{0.381 \pm 0.007}$
        & $0.318 \pm 0.001$
        & $0.318 \pm 0.001$
        & $\mathbf{0.681 \pm 0.008}$ \\
        \bottomrule
    \end{tabular}
    \label{tab:coco_preliminary}
\end{table}

\end{document}